\documentclass[lettersize,journal]{IEEEtran}
\usepackage{amsmath,amsfonts}
\usepackage{algorithmic}
\usepackage{algorithm}
\usepackage{array}
\usepackage[caption=false,font=normalsize,labelfont=sf,textfont=sf]{subfig}
\usepackage{textcomp}
\usepackage{stfloats}
\usepackage{url}
\usepackage{verbatim}
\usepackage{graphicx}
\usepackage{cite}
\usepackage{amssymb}
\usepackage{xcolor}
\usepackage{ctable}
\usepackage{multirow}
\usepackage{listings}

\begin{document}
\newcommand{\notehaibo}[1]{\textcolor{blue}{[haibo: #1]}}
\newcommand{\notesy}[1]{\textcolor{red}{[shuangyao: #1]}}
\newcommand{\notehzy}[1]{\textcolor{orange}{[Zhiyi: #1]}}
\newcommand{\revise}[1]{\textcolor{blue}{#1}}
\newenvironment{algocolor}{%
	\setlength{\parindent}{0pt}
	\itshape
	\color{blue}
}{}

\title{$E^2CoPre$: Energy Efficient and Cooperative Collision Avoidance for UAV Swarms with Trajectory Prediction}

\author{Shuangyao Huang\textsuperscript{\rm 1, 2}, 
	Haibo Zhang\textsuperscript{\rm 1} 
	and~Zhiyi Huang\textsuperscript{\rm 1}
	\thanks{\textsuperscript{\rm 1}School of Computing, 133 Union Street East, University of Otago, Dunedin 9016, New Zealand. E-mails: huangshuangyao@gmail.com, \{haibo.zhang, zhiyi.huang\}@otago.ac.nz}%
	\thanks{\textsuperscript{\rm 2}School of Internet of Things, Xi'an Jiaotong-Liverpool University, Suzhou, China.
	E-mail: shuangyao.huang@xjtlu.edu.cn}%
}

\markboth{Journal of \LaTeX\ Class Files,~Vol.~14, No.~8, August~2021}%
{Shell \MakeLowercase{\textit{et al.}}: A Sample Article Using IEEEtran.cls for IEEE Journals}


\maketitle

\begin{abstract}
	This paper presents a novel solution to address the challenges in achieving energy efficiency and cooperation for collision avoidance in UAV swarms. 
	The proposed method combines Artificial Potential Field (APF) and Particle Swarm Optimization (PSO) techniques. APF provides environmental awareness and implicit coordination to UAVs, while PSO searches for collision-free and energy-efficient trajectories for each UAV in a decentralized manner under the implicit coordination. This decentralized approach is achieved by minimizing a novel cost function that leverages the advantages of the active contour model from image processing. Additionally, future trajectories are predicted by approximating the minima of the novel cost function using calculus of variation, which enables proactive actions and defines the initial conditions for PSO. We propose a two-branch trajectory planning framework that ensures UAVs only change altitudes when necessary for energy considerations. Extensive experiments are conducted to evaluate the effectiveness and efficiency of our method in various situations. 
\end{abstract}


\section{Introduction} \label{intro}
\IEEEPARstart{U}NMANNED aerial vehicles (UAV) are aircraft capable of being controlled remotely or operating autonomously. 
Multi-rotor UAVs are studied in this paper owning to their prevalence in research and industry. {A UAV swarm is a group of UAVs that work together to perform a common task, such as search and rescue \cite{application4}, tracking and monitoring \cite{application4-1}, data collection \cite{application4-4}, and post-disaster communication recovery \cite{application4-3}. However, the blossom of these applications is limited by effective collision avoidance algorithms for UAV swarms. Collision avoidance for UAV swarms is challenging in terms of optimizing energy consumption and achieving cooperation. Firstly, the limited onboard power supplies of off-the-shelf UAVs e.g. DJI Matrice 600 
create constraints on energy consumption, limiting their application in long-distance missions. Secondly, a lack of cooperation may lead to collisions between swarm members, causing safety risks for the entire swarm and unnecessary energy consumption in avoiding each other. 
It is worth noting that the energy consumption of UAVs primarily occurs in two main domains: communication and propulsion. While recent advances in UAV communication strategies \cite{UAV-Comms1, UAV-Comms2} have achieved energy-efficient transmission by minimizing the transmission power based on the channel state information (CSI), propulsion energy remains a prominent contributor due to its significantly higher magnitude compared to communication energy. For instance, communication may consume only a few watts, whereas propulsion can demand hundreds of watts \cite{UAV-Energy}. Hence, this paper focuses on optimizing propulsion energy efficiency in collision avoidance for UAV swarms. }

Existing algorithms for UAV collision avoidance typically suffer from either energy inefficiency or difficulties in achieving cooperation. For instance, velocity-based algorithms 
\cite{vo, cone, reciprocalvo}
and virtual force-based algorithms 
\cite{b11, b12, b15} 
are energy inefficient. 
Model Predictive Control (MPC)-based methods 
\cite{mpc1, mpcapp1, mpcapp2} 
require well-defined control models and parameters for UAVs which are non-trivial to design. Metaheuristic-based algorithms \cite{b14} and Genetic Algorithms (GA) \cite{geneticapplication1} are more appropriate for planning trajectories of individual UAVs, as the dimensionality of the search space increases exponentially with the growth in swarm size. Above all, existing methods are based on receding horizon planning, where UAVs plan trajectories over a finite time window that shifts in the time domain. This receding horizon planning results in short-sightedness and can lead to sub-optimal trajectories from a long-term perspective since only the collisions occurring in the immediate next step are considered. 
Multi-Agent Reinforcement Learning (MARL) 
\cite{MADDPG_oda1, MADDPG_oda2, MADDPG_oda3, vdn_oda1} 
has recently been used to train cooperative policies for collision avoidance considering long-term perspective. 
However, MARL-based methods are subject to time- and  power-intensive training, high variances 
\cite{variance, variance1, variance2, variance3}, 
and low sample efficiency \cite{SampleEfficienty}, which can result in a high failure rate on collision avoidance during  execution. 

In this paper, we propose $E^2CoPre$ - a novel scheme for \textit{E}nergy-\textit{E}fficient and \textit{Co}operative Collision Avoidance with Trajectory \textit{Pre}diction for UAV Swarms. It overcomes the limitations of existing methods with the following contributions:  
\begin{itemize}
	\item \textbf{Coordination on collision avoidance:} $E^2CoPre$ provides implicit coordination on collision avoidance through APF-based environmental awareness and a novel cost function; 
	\item \textbf{Novel cost function:} $E^2CoPre$ introduces a novel cost function for PSO, leveraging the benefits of the active contour model used in image processing to generate energy-efficient and collision-free trajectories; 
	\item \textbf{Trajectory prediction:} $E^2CoPre$ addresses the short-sight limitation of UAVs by predicting their future trajectories at each step. This prediction is achieved by approximating the minima of the cost function; 
	\item \textbf{Energy-efficient inter-UAV collision avoidance:} $E^2CoPre$ schedules conflicting UAVs to different altitudes to resolve the collisions between UAVs with minimum energy consumption using PSO. 
\end{itemize}

{$E^2CoPre$ accomplishes these contributions by effectively addressing the challenges associated with modeling energy consumption and safety, expressing trajectories mathematically, and enhancing cooperative trajectory planning. The energy consumption is modeled based on the findings from our field trials and theoretical analysis in our previous work \cite{huang2021}. The mathematical modeling of safety is transformed into a gradient-based edge detection problem on the environmental field. The trajectory is expressed as a sequence of equally-spaced discrete waypoints. Cooperation of trajectory planning is addressed by APF-based environmental awareness and a novel cost function inspired by the active contour model \cite{snakes}.} 
We extensively evaluate $E^2CoPre$ against state-of-the-art collision avoidance schemes. Additionally, the effectiveness and robustness of $E^2CoPre$ are thoroughly validated through parameter analysis and ablation tests. 
The results show that $E^2CoPre$ outperforms the state-of-the-art schemes. 



The rest of this paper is organised as follows. Section \ref{Relatedwork} discusses related works on collision avoidance for UAV swarms. The system model and framework of $E^2CoPre$ are introduced in Sections \ref{System} and \ref{Framework}. Section \ref{APF} presents the environment representation using APF. Section \ref{Cost} details the cost functions used in PSO search. Trajectory prediction is introduced in Section \ref{Predict}, and trajectory planning using PSO is presented in Section \ref{Plan}. Experiments are presented in Section \ref{Simul}. Finally, conclusions are drawn in Section \ref{Concl}.

\section{Related Work} \label{Relatedwork} 
Existing studies on UAV collision avoidance can be divided into five categories: velocity-based, virtual force-based, metaheuristic-based, model predictive control-based, and machine learning-based. 
\subsection{Velocity-Based Methods}
Velocity-based methods aim to avoid collisions by adjusting the velocities of UAVs. A representative algorithm of this category is Velocity Obstacle (VO) \cite{vo, cone}. In this approach, each UAV maintains a pool of velocities known as velocity obstacle, which consists of velocities that would cause collisions with other UAVs or obstacles. UAVs avoid collisions by selecting velocities outside their velocity obstacles. While straightforward, velocity-based methods often result in frequent and abrupt velocity adjustments, leading to energy-inefficient trajectories.

\subsection{Virtual Force-Based Methods}
Virtual force-based methods utilize the concept of Artificial Potential Fields (APF) \cite{b15} to model the environment as a potential field, where differentiable functions represent field intensities with maxima at obstacles and minima at target points. The UAVs are guided by virtual forces derived from the gradients on the potential fields, which attract them to their destinations and push them away from obstacles \cite{b11}. 
However, local optima may appear on the potential field with zero gradients, resulting in UAVs getting trapped and unable to escape. Additionally, without coordination, UAVs may push each other back and forth like an oscillation, leading to energy-inefficient zig-zag trajectories.

\subsection{Metaheuristic-Based Methods} 
Metaheuristic-based methods, such as swarm intelligence, genetic algorithms, and graph path-finding algorithms, offer alternative solutions for collision avoidance in UAVs. Swarm intelligence methods, such as Particle Swarm Optimization (PSO) \cite{b14}, mimic the behavior of fish swarms searching for food by minimizing a cost function. Particles exchange information on personal and global best positions, and evolve together to the target point. The information exchange encourages a balance between exploring unknown areas and exploiting promising areas. 
In \cite{hyb5}, a term measuring the proximity to obstacles is introduced to the velocity update of particles in PSO. The new term acts like a virtual force repelling the particles away from the obstacles. In \cite{hyb5}, PSO search is performed on a potential field that has a smoothing effect. The costs are simply the field intensities. However, these methods are limited by the curse of dimensionality, which hinders the cooperation of swarm members as the search space's dimensionality increases with the swarm size. 

Genetic algorithms \cite{genetic}, on the other hand, mimic the evolutionary process of chromosomes by minimizing a cost function. 
In \cite{geneticapplication1}, collision-free trajectories of UAVs are planned using genetic algorithms. A set of trajectories are first initialized and keep evolving through crossover and mutation. A collision penalty is introduced in the cost function. Similar to swarm intelligence methods, these methods are limited by the curse of dimensionality. Hence, achieving cooperation among swarm members is challenging as the search space's dimensionality increases exponentially with swarm size.

Graph path-finding algorithms such as 
A* \cite{pathfinding1}, D* \cite{pathfinding2}, and D* Lite \cite{pathfinding3} 
incorporate heuristics into graph search algorithms such as Dijkstra's algorithm. The environment is modeled as a weighted graph whose nodes represent waypoints, and edge weights represent travel costs. In \cite{graphpathfinding1}, collision-free trajectories are searched using A* on weighted graph. The edges within proximity to the obstacles are removed from the graph. However, acquiring weighted graphs in complex and dynamic environments is non-trivial. Moreover, these methods are subject to the curse of dimensionality, as the sizes of weighted graphs grow exponentially with the number of UAVs. 

Overall, PSO generally outperforms other methods in various optimization problems due to its fast convergence, sufficient exploration, and efficient exploitation capabilities \cite{GAPSO}. 

\subsection{Model Predictive Control-Based Methods}
Model Predictive Control (MPC) generates optimal control signals based on the predictions of a system's future behavior over a finite time window. The control signals are determined by minimizing a cost function while satisfying system constraints at each step, and the time window is shifted in time horizon for the following steps. Linear and nonlinear MPC are two classes of MPC. Linear MPC is only valid over a small section of the trajectory as it simplifies the system dynamics by linearizing the UAV model. On the other hand, nonlinear MPC provides a more accurate model of UAV dynamics by directly modeling the nonlinearities in the system. However, designing the nonlinear model for UAVs is complex and challenging.  
In \cite{mpcapp1}, future trajectories of UAVs are predicted by linear MPC satisfying kinetic constraints over a finite time horizon. Collisions are identified by proximity between future trajectories, and are resolved by altitude change. 
In \cite{mpcapp2}, nonlinear MPC is used to find trajectories for UAVs minimizing a cost function that considers both obstacles and kinetic constraints of UAVs. 
However, it requires well-defined control models and parameters that are challenging to acquire. 

Above all, velocity-based, virtual force-based, metaheuristic-based, and model predictive control-based methods are all based on receding horizon planning, which results in short-sightedness of UAVs and can lead to sub-optimal trajectories from a long-term perspective. 

\subsection{MARL-Based Methods} 
Multi-Agent Reinforcement Learning (MARL) is a promising approach for multi-UAV collision avoidance 
\cite{MADDPG_oda1, MADDPG_oda2, MADDPG_oda3, vdn_oda1}. 
MARL trains cooperative policies for UAVs in a centralized manner, considering global information that is not accessible during execution. 
However, the success of MARL-based methods heavily relies on the availability of well-defined simulation environments and reward signals, as well as extensive training time and computational power. 
Additionally, MARL suffers from high variances
\cite{variance, variance1, variance2, variance3}
and low sample efficiency \cite{SampleEfficienty}, leading to sub-optimal policies and high failure rates during online execution. Thus, it is challenging to implement MARL-based methods in pratice. 

Aiming to address the limitations of existing methods, $E^2CoPre$ achieves cooperation among UAVs through the implicit coordination of APF, while energy efficiency is ensured by minimizing a novel cost function using trajectory-based PSO. To reduce the complexity of PSO search space, each trajectory is represented with only two variables, and the short-sightedness of UAVs is addressed by trajectory prediction. 

\section{System Model} \label{System}



This paper considers a scenario in which UAVs follow pre-planned static paths defined by sequences of waypoints. During the flight, when potential collisions are detected, the UAVs take necessary actions to avoid the collisions. Once the collisions are avoided, the UAVs fly toward the next waypoint on their path to resume their course. These upcoming waypoints, which the UAVs aim to reach after avoiding collisions, are referred to as the target points of the collision avoidance process. 
In the paper we assume that static obstacles such as buildings and towers are avoided in static path planning. Our main focus is avoiding collisions with dynamic obstacles such as birds or adversarial UAVs. 

{During collision avoidance, UAVs can change altitude but must return to their original altitudes after the maneuver. Hence, flying upward or downward during collision avoidance makes no difference in energy consumption. We assume the speed of each UAV remains constant during collision avoidance, and each UAV adjusts its direction of movement to avoid collisions for energy efficiency. This assumption stems from the recognition that dynamic obstacle avoidance usually occurs within a short timeframe, making sudden speed changes impractical due to the large accelerations involved, resulting in a substantial increase in energy consumption. } 

{It is also assumed that each UAV knows its position, velocity, and that of other swarm members and obstacles within its sensing range, which can be achieved using GPS and LiDAR sensors. Typical LiDAR sensors, e.g. SICK MRS 1000,  
are capable to detect obstacles in three-dimensional space with a horizontal aperture angle of 270$^{\circ}$ and a vertical aperture angle of 7.5$^{\circ}$. SICK MRS1000 also has a high scanning frequency of 50 $Hz$ and a small angular resolution of 0.25$^{\circ}$. 
Thus a UAV mounted with such a LiDAR sensor can scan the environment every 20 $ms$ using a scanning window of $270^{\circ}\times7.5^{\circ}$ with an angular resolution of 0.25$^{\circ}$. Also onboard microcomputer, e.g. Raspberry Pi 
is used for decision-making. }

\section{System Framework} \label{Framework}
\begin{figure}[t]
	\centering
	\includegraphics[width=0.45\textwidth]{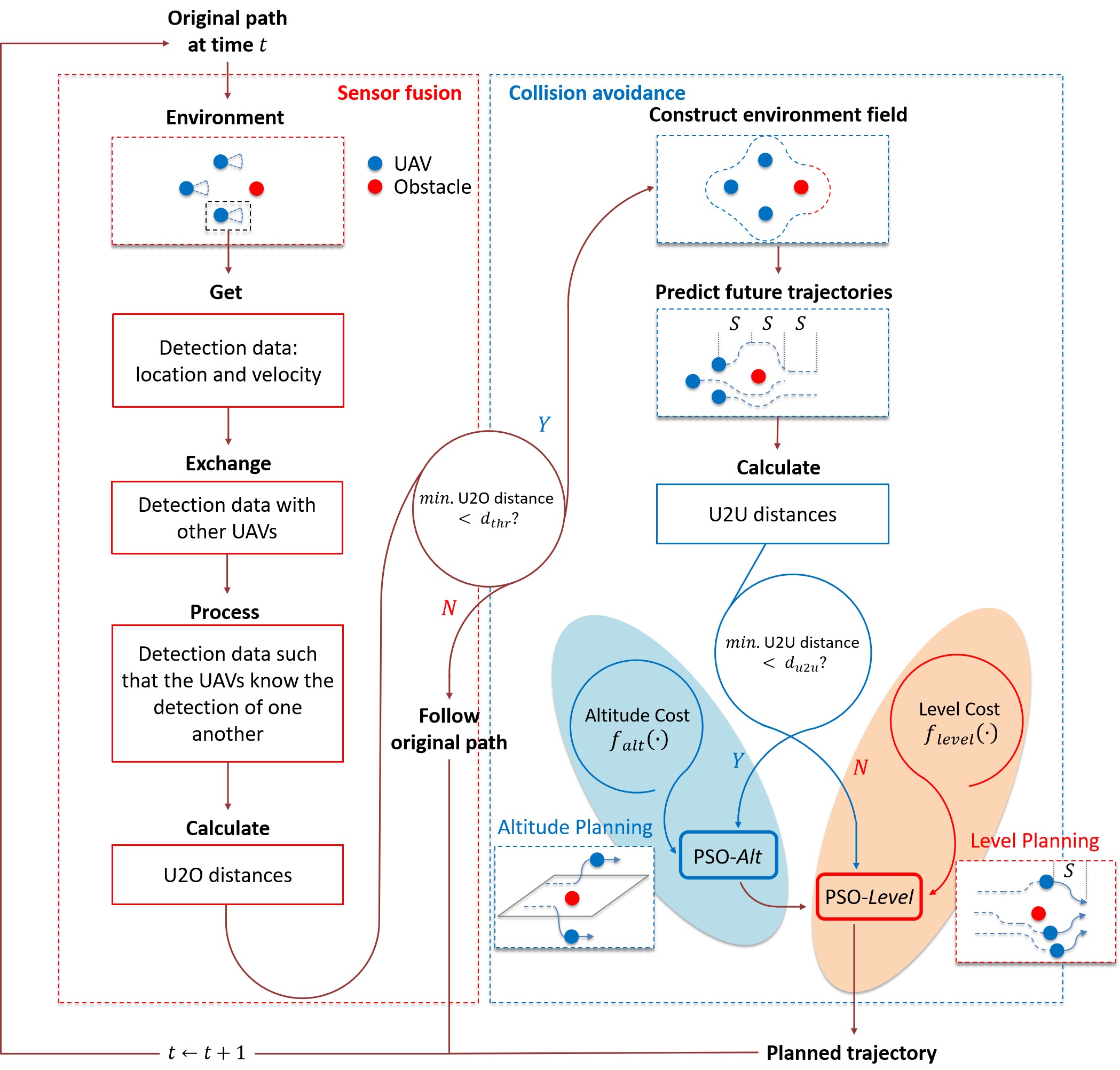}
	\caption{The system framework of $E^2CoPre$.}
	\label{fig1}
\end{figure}

{The system flow for one planning step is detailed in Fig.~\ref{fig1}. The entire process is repeated at each planning step, even if no obstacles are detected, to ensure that the UAVs consistently maintain awareness of each other. The UAVs initially fly along their pre-planned trajectories. 
Each UAV first gets its position and velocity, as well as obstacles at each step. These information is then exchanged within the swarm and processed by each UAV such that the UAVs are aware of the status of one another and are informed of the obstacles detected by their teammates. Collision avoidance initiates when the minimum UAV-to-Obstacle (U2O) distance falls below a predefined safety threshold $d_{thr}$. This approach ensures collision avoidance is initiated only when it becomes imperative, as governed by the safety threshold $d_{thr}$. When obstacles are initially detected, they often remain at a considerable distance from the UAVs, and the risk of collisions is low. Therefore, initiating the collision avoidance process at this point is unnecessary. All the UAVs initiate the collision avoidance process together to cooperatively avoid collisions. 
During collision avoidance, the environment field is first constructed and shared among the UAVs to establish environment awareness and provides implicit coordination. Based on this environment field, the trajectory of each UAV for future planning steps is predicted. The predicted trajectories detect potential UAV-to-UAV (U2U) collisions and initialize the PSO search in trajectory planning. The trajectory planning is divided into two branches: level planning and altitude planning. If no potential U2U collision is detected by trajectory prediction, level planning plans the trajectory by minimizing a level cost function $f_{level}(\cdot)$ using PSO. 
On the other hand, if a potential U2U collision is detected, altitude planning steps in, scheduling conflicting UAVs to collision-free level planes at different altitudes by minimizing an altitude cost function $f_{alt}(\cdot)$ using PSO. Subsequently, level planning proceeds to plan the trajectory for each UAV on their respective level planes. The UAVs return to their pre-planned trajectories after collision avoidance. 
This two-branch approach optimizes the energy consumption of UAVs by enabling them to maintain flight on level planes whenever possible, and only adjusting altitudes when necessary. }

\section{Environment Field Construction} \label{APF} 

\begin{figure}[b]
	\centering
	\subfloat[]{\includegraphics[width=0.5\linewidth]{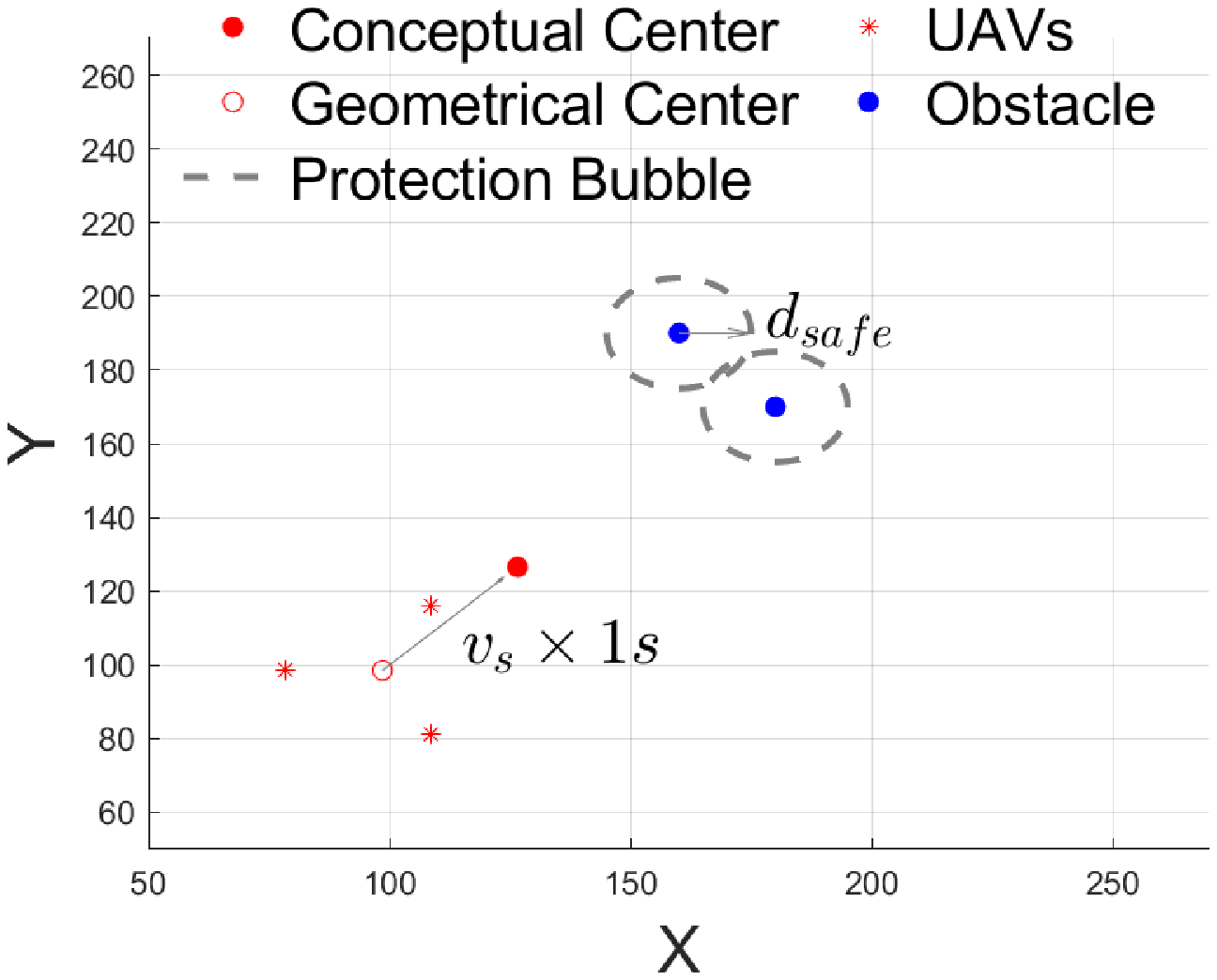}%
		\label{2a}}
	\hfil
	\subfloat[]{\includegraphics[width=0.5\linewidth]{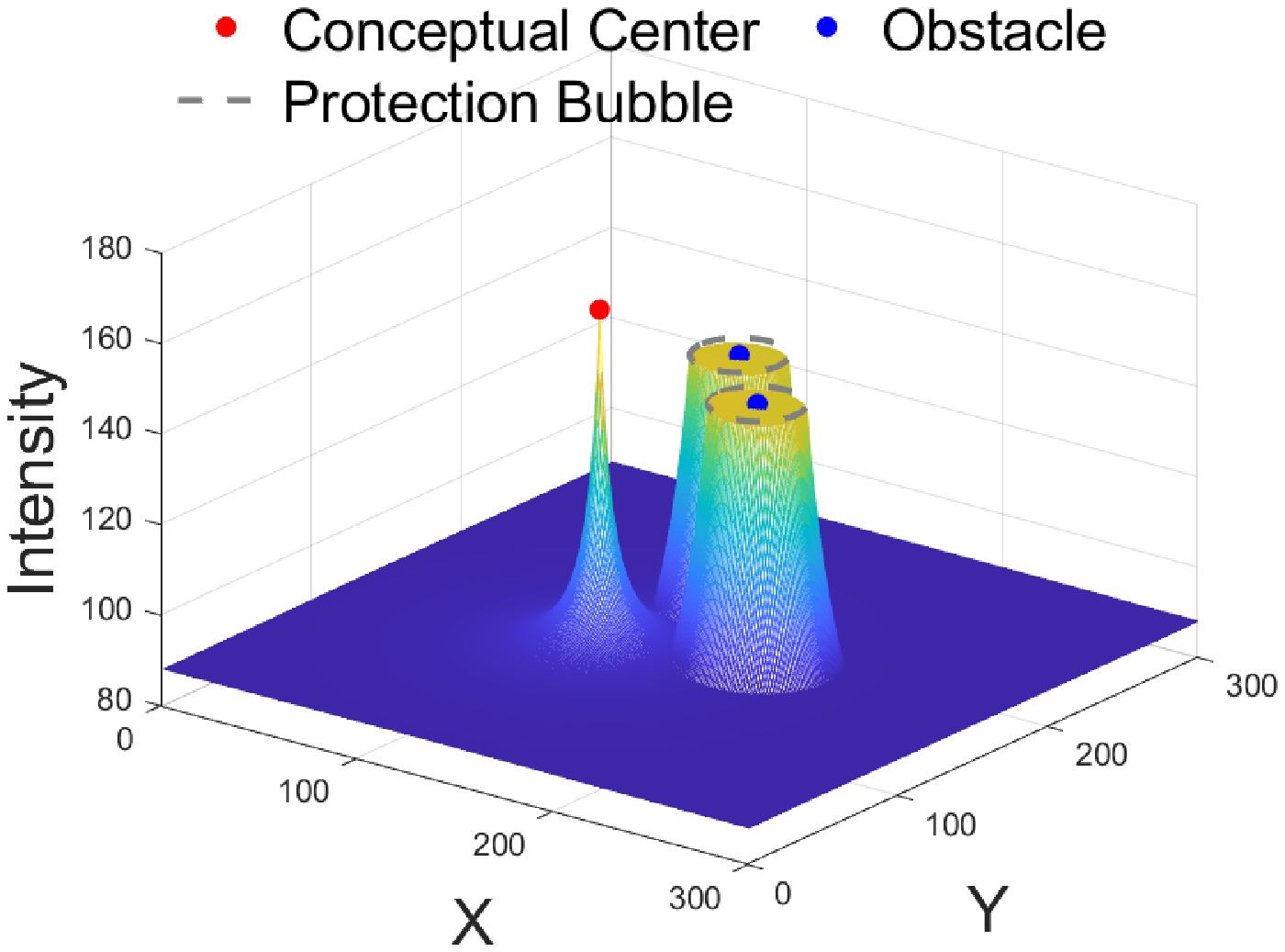}%
		\label{2b}}
	\caption{Example of environment field. (a) The two-dimensional layout of the environment. (b) The environment field in three-dimensional view. }
	\label{fig2}
\end{figure}

We define the environment field as the addition of multiple repulsive fields, where each detected obstacle is modeled as a repulsive field, and the UAV swarm as one entity is modeled as a repulsive field. The environment field is constructed when a UAV detects an obstacle within its sensing range using LiDAR sensors, and it is shared among all the swarm members. If the obstacles are at different altitudes from the swarm, their repulsive fields are added to the swarm's field. Details on the obstacles' and swarm's repulsive fields are available in~\cite{huang2021}. 
Fig.~\ref{fig2} gives an example of the environment field containing a swarm of three UAVs and two obstacles. 

The repulsive field of the swarm $\varPhi_s(q)$ is constructed as: 
\begin{equation} \label{eq:swarmfield} 
	\begin{aligned} 
		\varPhi_s(q)=\left\{\begin{array}{l c}  
			\dfrac{v_s}{|{q, p^*}|^2}, & |{q, p^*}| \leq R^s \\ 
			~~~0, & |{q, p^*}| > R^s
		\end{array} 
		\right. 
	\end{aligned} 
\end{equation} 
where $q$ represents a UAV's position, and $\varPhi_s(q)$ is the field intensity at position $q$. $|{\cdot, \cdot}|$ is the distance between two points, and $R^s$ is the influential range of the field. $p^*$ represents the conceptual center of the swarm. The conceptual center is defined by shifting the swarm's geometric center towards its target point by one planning step. Constructing the swarm's repulsive field on its conceptual center reflects the swarm's velocity in the environment field and avoids attractive fields. 

Instead of modeling each UAV individually, the swarm is represented as a repulsive field to prevent UAVs from being placed at peaks on the environment field. 
The swarm-based repulsive field enables implicit coordination. As introduced in Section \ref{Cost}, the implicit coordination is achieved by planning trajectories of UAVs along contours of the environment field. There will be no contours for UAVs if they are placed at peaks of the environment field. 


On the other hand, the repulsive field for the $j^{th}$ obstacle detected by the swarm is defined as follows. 
\begin{equation} \label{eq:obstaclefield} 
	\begin{aligned} 
		\varPhi^o_j(q)=\left\{\begin{array}{l c}  
			\dfrac{\max\{v^o_j, v_s\}}{d^2_{safe}}, & |{q, p^o_j}|\leq d_{safe} \\ 
			\dfrac{\max\{v^o_j, v_s\}}{|{q, p^o_j}|^2}, & d_{safe} < |{q, p^o_j}| \leq R_j^o \\ 
			~~~~~~~0, & |{q, p^o_j}|> R_j^o 
		\end{array} 
		\right. 
	\end{aligned} 
\end{equation} 
where $p^o_j$ is the obstacle's position, and $R_j^o$ is the influential range of the field. The $\max$ operator ensures the swarm field and the obstacle's field have comparable intensities when $v^o_j < v_s$, especially when $v^o_j=0$ for static obstacles. $d_{safe}$ represents the minimum distance that a UAV should maintain from obstacles to ensure safety. However, it is important to note that being closer to obstacles than $d_{safe}$ does not necessarily mean an imminent collision. Instead, UAVs are alerted, and are required to fly away. The purpose of setting $d_{safe}$ is to provide a safety buffer that allows UAVs to detect and respond to potential risks before causing imminent collisions. 
The area within $d_{safe}$ around the obstacle is called protection bubble, and has the maximum field intensity. Protection bubble prevents UAVs from getting closer to the obstacle than $d_{safe}$. Details on protection bubble will be given in the following section as it's achieved using the level cost function. 


Based on the potential fields for the swarm and obstacles, the environment field $\varPhi(q)$ is defined by adding all these fields together as follows:
\begin{equation} \label{eq:environmentfield} 
	\begin{aligned} 
		\varPhi(q)=&\varPhi_s(q)+ \sum_{j=1}^{M}\varPhi^o_j(q) \\ 
	\end{aligned} 
\end{equation} 
where $M$ is the number of obstacles. 


%

\section{Cost Functions} \label{Cost}


\subsection{Level Cost $f_{level}(\cdot)$} \label{Cost1} 
$f_{level}(\cdot)$ consists of two parts: an energy efficiency part $f_{eng}(\cdot)$ and a safety part $f_{saf}(\cdot)$, as follows. 
\begin{equation} \label{eq:fitness} 
	\begin{aligned} 
		f_{level}(S^k)=\lambda_1f_{eng}(S^k)+\lambda_2f_{saf}(S^k), \\ 
	\end{aligned} 
\end{equation} 
where $S^k$ is the trajectory to be optimized, and $k$ is the number of planning steps in the trajectory. $k>1$ for trajectory prediction, as it predicts future trajectories. On the other hand, $k=1$ for trajectory planning, as it optimizes the trajectory for the immediate next step. 
$f_{eng}(S^k)$ is the cost function to ensure energy efficiency whereas $f_{saf}(S^k)$ is the cost function to ensure safety. $ \lambda_1$ and $\lambda_2$ are coefficients where $\lambda_1+\lambda_2=1$. 

\subsubsection{Energy Efficiency $f_{eng}(\cdot)$} \label{Energy} 


The energy consumption of a quad-copter flying along a trajectory 
$S^k$ can be modeled as follows. Details on derivation can be found in \cite{huang2021}. 
\begin{equation} \label{eq:energy} 
	\begin{gathered} 
		E = E_n + E_{len} + E_{comms} \\
		E_{n} = \int_{\eta_s}^{\eta_e} F_n(S^k(p)) (v \sin\alpha +v_i)\cdotp \sin\beta dp \\ 
		E_{len} = (F_{drag} \sin\alpha \cos\beta + mg\cos\alpha\cos\beta) \\ 
		\cdot(v \sin\alpha +v_i) \cdot(\eta_e-\eta_s) \\ 
		E_{comms} = \int_{\eta_s}^{\eta_e} P_{comms}(S^k(p)) dp, 
	\end{gathered} 
\end{equation} 
where $E_{n}$ is the energy consumed on performing turnings, $E_{len}$ is the energy depending on trajectory length, and $E_{comms}$ is the energy consumed in communication. 
$p$ is an arc length parameter and $\eta_e = \eta_s + k\cdot|S|$ where $|S|$ is the length of one planning step. 
$[\alpha, \beta, \gamma]$ are the UAV body's pitch, roll, and yaw angles, $F_n$ is the centripetal force required to perform turnings, $F_{drag}$ is the air drag force, $v$ is the UAV's ground speed, $v_i$ is the induced velocity of the propellers, $m$ is the total mass of the UAV, and $g$ is the gravitational acceleration. $P_{comms}(S^k(p))$ is the energy consumed by the communication module, and it is a function of $S^k(p)$ because the amount of data sent and received depends on the length of the trajectory. 

In one planning step, it is reasonable to assume that the pitch angle $\alpha$,  velocity $v$, air drag force $F_{drag}$, and induced velocity $v_i$ do not change when the UAV is flying along the trajectory. Moreover, the trajectory length in one planning step is fixed. Therefore, $E_{len}$ and $E_{comms}$ are all constant in one planning step. 
Hence, minimizing $E$ is equivalent to minimizing $E_n$. 
Field experiments have been conducted to validate the above assumptions. Comprehensive details regarding these experiments can be found in our earlier publication \cite{huang2021}.



For mathematical modeling, let $r(S^k(p))$ be the turning radius of the UAV, we have $F_n(S^k(p))=\frac{mv^2}{r(S^k(p))}$. Hence, 
\begin{equation} \label{eq:energy1} 
	\begin{aligned} 
		E_n = & mv^2(v \sin\alpha +v_i)\sin\beta\int_{\eta_s}^{\eta_e} \frac{1}{r(S^k(p))} dp \\ 
		= & mv^2(v \sin\alpha +v_i)\sin\beta \int_{\eta_s}^{\eta_e} | \kappa(S^k(p)) | dp \\ 
		= & e_v \int_{\eta_s}^{\eta_e} |[S^{k}(p)]''| dp, 
	\end{aligned} 
\end{equation} 
where $\kappa(S^k(p))$ and $[S^{k}(p)]''$ are the curvature and the second order derivative of trajectory $S^k(p)$ with regard to $p$. $e_v = mv^2(v \sin\alpha +v_i)\sin\beta$ is a velocity-dependent coefficient and is constant in a small step. 

For the sake of math modeling, the energy efficiency term of $f_{level}(S^k(p))$ is designed as follows.  
\begin{equation} \label{eq:fitness1} 
	\begin{aligned} 
		f_{eng}(S^k(p))= \int_{\eta_s}^{\eta_e} \dfrac{1}{2}|[S^{k}(p)]''|^2 dp,  
	\end{aligned} 
\end{equation} 
where the square helps the calculation in trajectory prediction. 

\subsubsection{Safety $f_{saf}(\cdot)$} \label{Safety} 

\begin{figure}[t]
	\centering
	\subfloat[]{\includegraphics[width=0.5\linewidth]{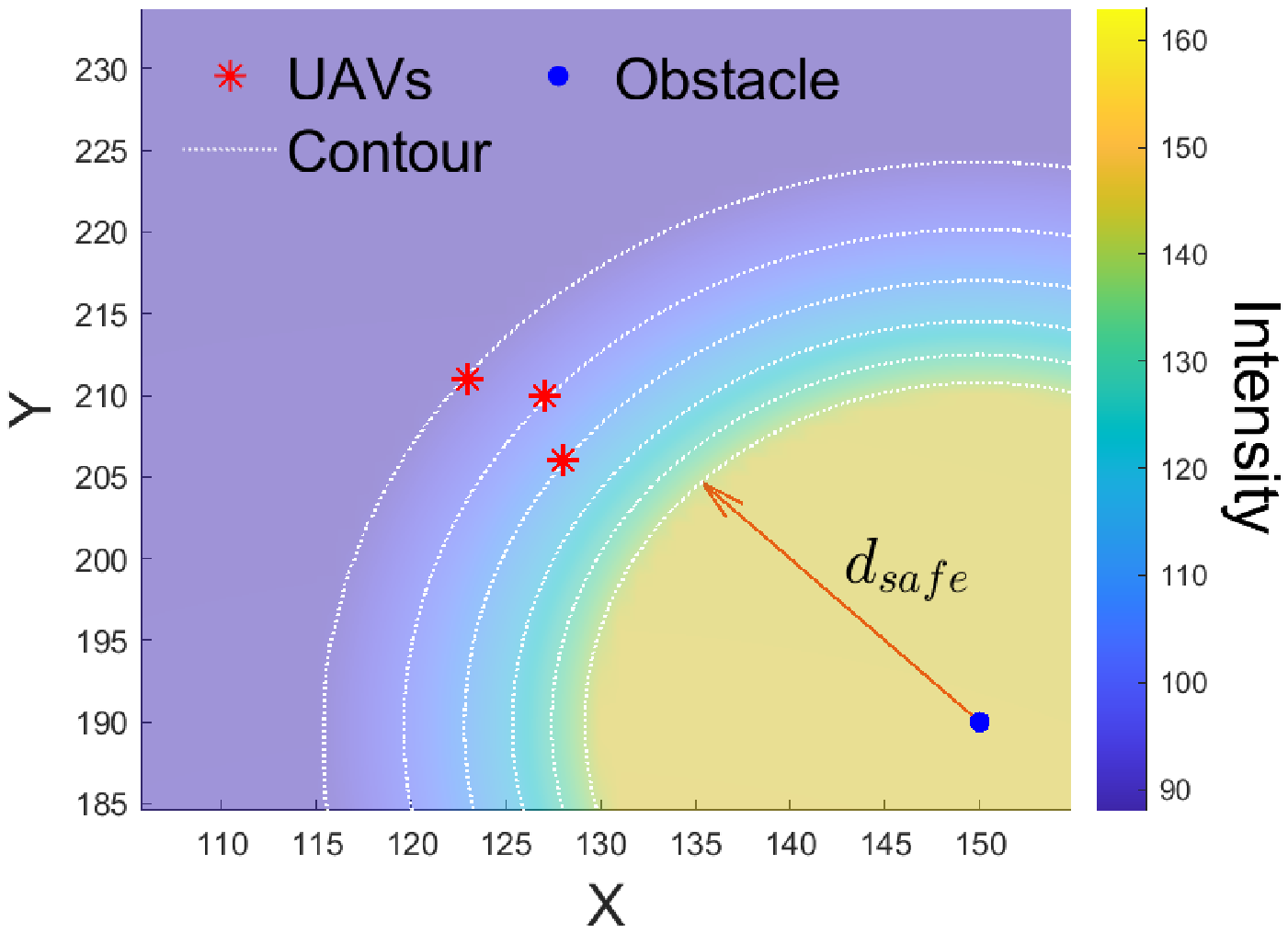}%
			\label{contours1}}
	\subfloat[]{\includegraphics[width=0.5\linewidth]{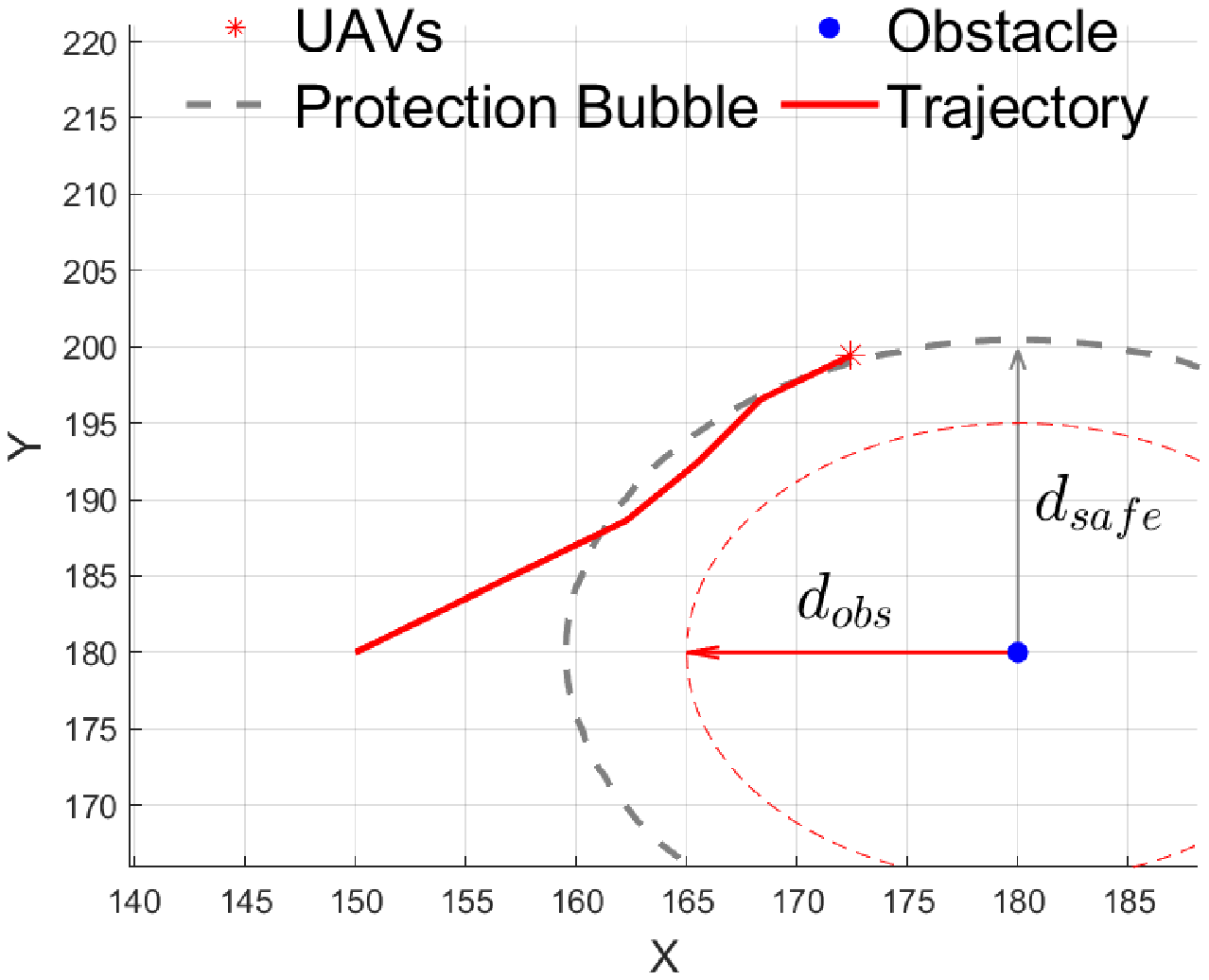}%
			\label{contours2}}
	\caption{Contours and protection bubble. (a) The contours of the environment field. (b) The protection bubble around an obstacle. }
	\label{contours}
\end{figure}

The safety term $f_{saf}(S^k(p))$ ensures collision avoidance by planning trajectories for UAVs along different contours, which represent curves at various intensity levels of the environment field. 
As shown in Fig. \ref{contours1}, the contours are smooth, ensuring energy efficiency, and they never pass through peaks, which correspond to obstacles. Moreover, the contours never intersect, effectively separating the UAVs and avoiding collisions between them. 

A trajectory along a contour corresponds to the curve that has the minimum variation in intensity. To facilitate this, we transform the environment field into a binary field using Eq. \eqref{eq:binfield}, such that any points with intensities lower than the UAV's current field intensity is assigned an intensity of 1, while those with higher intensities are assigned an intensity of -1. 
\begin{equation} \label{eq:binfield}
	\varPhi_{b}(q) = \begin{cases} ~~1, & \varPhi(q)\geq \varPhi(p_0),\\
		-1,  &  \varPhi(q)< \varPhi(p_0),
	\end{cases}
\end{equation}
where $p_0$ and $\varPhi(p_0)$ are the UAV's current position and field intensity, respectively. 
Eq. \eqref{eq:binfield} ensures that the curve with the minimum variation in intensity becomes an edge on the binary field, corresponding to the regions with the maximum gradient magnitudes.
To identify such an edge on the binary field, the safety term of $f_{level}(S^k(p))$ is defined as follows: 
\begin{equation} \label{eq:fitness2} 
	f_{saf}(S^k(p)) = -\int_{\eta_s}^{\eta_e} \dfrac{1}{2}|\triangledown\varPhi_{b}(S^k(p))|^2 dp, 
\end{equation} 
where $\triangledown\varPhi_{b}$ represents the gradients of binary field $\varPhi_{b}$. $f_{saf}(S^k(p))$ is minimized when $S^k(p)$ overlaps with an edge of $\varPhi_{b}$, which is a contour on the environment field. 
Eq. \eqref{eq:binfield} and \eqref{eq:fitness2} play an essential role in preventing UAVs from entering the protection bubble. As shown in Fig. \ref{contours2}, if a UAV enters the protection bubble of an obstacle, it will be attracted back to the edge of the protection bubble at the next step. This is because that the edge of the protection bubble has the maximum gradients and, therefore, the minimum cost. 

Substituting Eq. \eqref{eq:fitness1} and Eq. \eqref{eq:fitness2} into Eq. \eqref{eq:fitness}, the level cost $f_{level}(\cdot)$ has the following form: 
\begin{equation} \label{eq:fitnessnew} 
	\begin{aligned} 
		f_{level}(S^k(p))=\int_{\eta_s}^{\eta_e} \dfrac{1}{2}\lambda_1 |[S^{k}(p)]''|^2 - \dfrac{1}{2}\lambda_2 | \triangledown\varPhi_{b}(S^k(p)) |^2 dp. \\ 
	\end{aligned} 
\end{equation} 
Eq. \eqref{eq:fitnessnew} resembles the active contour model \cite{snakes}, a common approach for contour extraction in image processing. However, our method differs in that it lacks a term for minimizing trajectory length, as the trajectory length is fixed to $k\cdot|S|$. 


Although planning trajectories on contours avoid passing through peaks or intersect with each other, hard constraints on U2O and U2U distances are necessary to ensure safety. Specifically, the distance between any UAV and an obstacle should be no smaller than a threshold distance $d_{obs}$, and the distance between any two UAVs should be no smaller than a threshold distance $d_{u2u}$ at all times. Thanks to the protection bubble, the hard constraint on U2O distance can be easily ensured by setting $d_{safe}\geq d_{obs}+|S|$. The hard constraint on U2U distance will be ensured by altitude planning, which will be introduced in the following subsections.

\subsection{Altitude Cost $f_{alt}(\cdot)$} \label{Cost2} 

The altitude cost function $f_{alt}(\cdot)$ is used to adjust the conflicting UAVs to different altitudes to avoid collisions. $f_{alt}(\cdot)$ measures the safety and energy consumption of UAVs in altitude change, and is designed as follows. 
\begin{equation} \label{eq:fitnessz} 
	\begin{gathered} 
		f_{alt}(\boldsymbol{S}) = \sum_{i\in[1,W]}|\triangle Alt(S_i)|, \\ 
		\text{subject\ to} \min_{i, j\in[1, W], i\neq j}|S_i, S_j|\geq d_{u2u}, 
	\end{gathered} 
\end{equation} 
where $W$ is the number of UAVs involved in the collisions, $d_{u2u}$ is the threshold distance for U2U collisions, $\boldsymbol{S}=[S_1, S_2, \cdots, S_W]$ are the trajectories of the UAVs, and $|\triangle Alt(S_i)|$ denotes the altitude change of the $i$-th UAV. The modulus operator stems from the system model where UAVs change altitude to avoid collisions but must return to their original altitudes after the maneuver. Hence, a positive or negative $\triangle Alt(S_i)$ makes no difference in energy consumption. 

Minimizing $f_{alt}$ ensures the UAVs are at least $d_{u2u}$ from each other. At the same time, the collective altitude change, and hence the total energy consumption of the swarm, is also minimized. 
$f_{alt}$ is only used for altitude scheduling, hence is independent of the lengths of trajectories. Therefore, the notations of $p$ and $k$ are ignored. 

\section{Trajectory Prediction} \label{Predict} 

The objective of trajectory prediction is to predict the optimal trajectory $[S^{k}]^*$ for $k$ future steps by minimizing $f_{level}(\cdot)$. 
Trajectory prediction should satisfy two requirements: 
\begin{itemize}
	\item \textbf{Prompt Response:} The trajectory prediction method must be computational light for prompt response; 
	\item \textbf{Long View:} $k$ must be larger than 1. 
\end{itemize}

These requirements prevent the usage of metaheuristic-based algorithms such as PSO due to high time complexity, which increases with $k$. Therefore, a faster optimization method is needed. 
$f_{level}({S}^{k})$ is an integral over a trajectory ${S}^{k}$. The method commonly used to find the minimum of an integral is known as calculus of variation in algebra. 

Let $y(x)$ be a function of $x$, and $F(y, y', y''|x)$ be a function of $y$ and its derivatives $y'$ and $y''$, calculus of variation minimizes the integral \[J(y) = \int_{x_0}^{x_1}F(y, y', y'', \cdots|x)dx\] by solving the Euler-Lagrange (EL) equation 
\begin{equation*} \label{EL1}
	\begin{aligned} 
			\dfrac{\partial F}{\partial y} - \dfrac{\partial}{\partial x}\dfrac{\partial F}{\partial y'} + \dfrac{\partial^2}{\partial x^2}\dfrac{\partial F}{\partial y''} = 0. 
		\end{aligned} 
\end{equation*} 
The EL equation for $F(S(p), S''(p)|p)=f_{level}({S}^{k})$ is expressed as
\begin{equation} \label{eq:EL0} 
	\begin{aligned} 
		\lambda_1 [S^{k}(p)]'''' - \lambda_2 \triangledown | \triangledown\varPhi_{b}(S^{k}(p))| = 0, 
	\end{aligned} 
\end{equation} 
where $[S^{k}(p)]''''$ is the forth order derivative of $S^{k}(p)$ with regard to $p$. 
Let $E_{s}=-| \triangledown\varPhi_{b}(S^{k}(p))|$, Eq. \eqref{eq:EL0} becomes 
\begin{equation} \label{eq:EL} 
	\begin{aligned} 
		\lambda_1 [S^{k}(p)]'''' + \lambda_2 \triangledown E_{s} = 0 
	\end{aligned} 
\end{equation} 

The solution for Eq. \eqref{eq:EL} can be found through iterative optimization of $S^{k}(p)$. 
We first give the right-hand side of Eq. \eqref{eq:EL} an initial term $\dfrac{\partial S_{n}^{k}(p)}{\partial {n}}$ where $n$ denotes the number of iterations. Eq. \eqref{eq:EL} becomes: 
\begin{equation} \label{eq:EL1} 
	\begin{aligned} 
		\lambda_1 [S^{k}(p)]'''' + \lambda_2 \triangledown E_{s} = \dfrac{\partial S_n^{k}(p)}{\partial n}. \\ 
	\end{aligned} 
\end{equation} 
The right-hand side of Eq. \eqref{eq:EL1} becomes 0 when the level cost function Eq. \eqref{eq:fitnessnew} stabilizes at the minimum. Eq. \eqref{eq:EL1} can be viewed as a gradient descend algorithm searching for the minimum of Eq. \eqref{eq:fitnessnew}. 

{
The difficulty in directly solving Eq. \eqref{eq:EL1} lies in that it is difficult to give a mathematical expression for $S_n^{k}$ as it represents a random-shaped trajectory. Instead, we express $S^k_n$ as a sequence of equally-spaced discrete points $S^k_n=[\boldsymbol{s}_1, \boldsymbol{s}_2, \cdots, \boldsymbol{s}_L]_n$, $S^k_n(i)=\boldsymbol{s}_i=[x_i, y_i]=[x(ih), y(ih)]$, where $h\in\mathbb{R}^+$. $L$ is the length of $S^k_n$.
Therefore, we have 
\begin{equation} \label{vectorize}
	\begin{gathered} 
		[S_n^{k}]''''(i) = \dfrac{\boldsymbol{s}_{i+2}-4\boldsymbol{s}_{i+1}+6\boldsymbol{s}_i-4\boldsymbol{s}_{i-1}+\boldsymbol{s}_{i-2}}{h^4}. 
	\end{gathered} 
\end{equation} 
For simplicity, we assume $h=1$. 
Substituting Eq. \eqref{vectorize} into Eq. \eqref{eq:EL1}, we have matrix form of the EL equation 
\begin{equation} \label{eq:EL2} 
	\begin{aligned} 
		{S}^k_{n+1} = {M}^{-1} ({S}^k_{n}+\triangledown E_{s}), 
	\end{aligned} 
\end{equation} 
where 
\begin{equation*} 
	\begin{gathered} 
		{S}^k_n = \left[\begin{matrix}
				\boldsymbol{s}_1,  
				\boldsymbol{s}_2, 
				\boldsymbol{s}_3, 
				\cdots, 
				\boldsymbol{s}_{L}
			\end{matrix}\right]^T_n, \\ 
		M = \left[\begin{matrix}
				u_1 & u_3 & u_2 & \cdots & u_2 & u_3 \\
				u_3 & u_1 & u_3 & u_2 & \cdots & u_2 \\
				u_2 & u_3 & u_1 & u_3 & u_2 & \cdots \\
				& & & \cdots \\ 
				u_3 & u_2 & \cdots & u_2 & u_3 & u_1 \\
			\end{matrix}\right], \\ 
		u_1=-6\lambda_1+1, u_2=-\lambda_1, u_3=4\lambda_1
	\end{gathered} 
\end{equation*} 
Based on Eq. \eqref{eq:EL2}, UAVs in a swarm can predict their trajectories independently at the same time. Since the trajectory prediction using Eq. \eqref{eq:EL2} is just a linear mapping of matrices, the computational overhead of this module is neglectable. 
}

Let $S^*$ be the prediction result. The collisions between UAV $i$ and $j$ are detected by the minimum distance between the predicted trajectories 
\begin{equation}  
	\begin{aligned} 
		d_{i,j} = \min_{1\leq l\leq L}\mid S_i^*(l), S_j^*(l)\mid
	\end{aligned} 
\end{equation} 
where $S_i^*$ and $S_j^*$ are the prediction results for UAV $i$ and $j$, respectively. $S_i^*(l)$ and $S_j^*(l)$ represent the $l^{th}$ waypoint in trajectory $S_i^*$ and $S_j^*$, respectively. Collisions occur between UAV $i$ and $j$ if $d_{i,j}$ is smaller than a preset collision threshold. On the other hand, the parameters of $S^*$ are used to initialize PSO-\textit{Level} in level planning and are detailed in Section \ref{Plan}. 


\section{Trajectory Planning} \label{Plan}

\subsection{Level Planning} \label{Level}
If the predicted trajectories show no collision, the system enters level planning. 
Level planning plans the optimal trajectories for actual flight in the immediate next step by minimizing $f_{level}(\cdot)$ using PSO-\textit{Level}. 

\begin{figure}[bt]
	\centering
	\includegraphics[width=0.35\textwidth]{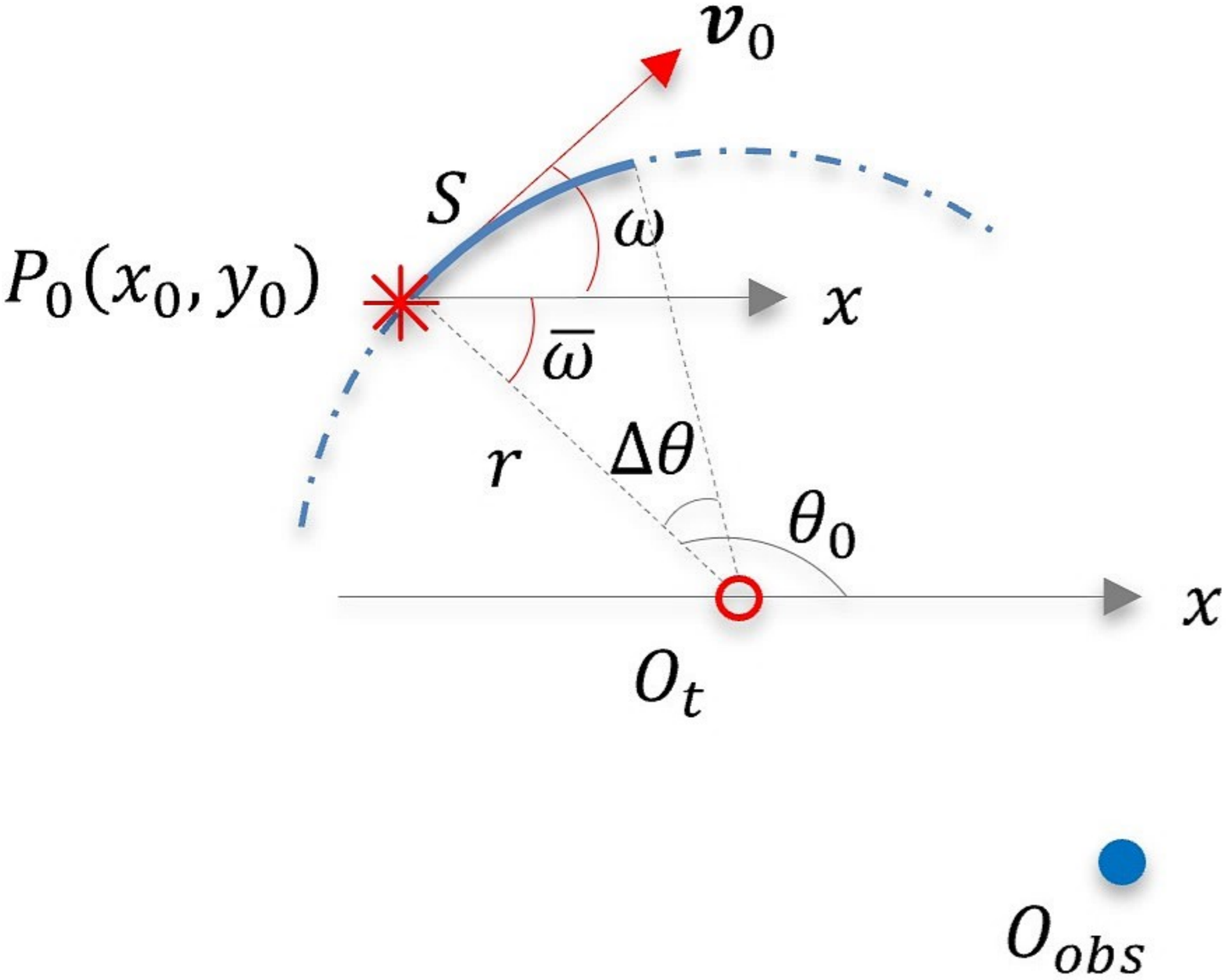} 
	\caption{Parameterization of a trajectory. }
	\label{figPara}
\end{figure}

PSO-\textit{Level} is trajectory-based PSO. Each particle represents a trajectory. Since one planning step is small, we use an arc to represent a trajectory as it is natural for turnings of multi-rotor copters and matches their aerodynamics. As the trajectory length in one planning step is fixed, a trajectory can be expressed using two variables, curvature $\kappa$ and slope $\omega$, as illustrated in Fig. \ref{figPara}. 
Therefore, the search space of PSO-\textit{Level} is reduced to two-dimensional to avoid the curse of dimensionality. 
In Fig. \ref{figPara}, the UAV and obstacle are depicted with red asterisks and blue dots, respectively. The dashed blue curve is a contour of the environment field. The solid blue curve $S$ is the trajectory being planned. The position and velocity of the UAV are $P_0$ and $\boldsymbol{v}_0$, respectively. $O_t$ and $r$ are the center and radius of the arc $S$. $\delta\theta$ and $\omega$ are the center angle and slope of the arc $S$. 
As illustrated in Fig. \ref{figPara}, an arc $S$ is uniquely defined by its variables $[\omega, \kappa]$ and initial point $[x_0,y_0]$. Hence, a particle can be expressed as $\boldsymbol{\xi}=[\omega, \kappa]_{|[x_0,y_0]}$. When $\kappa=0$, it is a straight line along the UAV's current velocity. Any point $P_i=[x_i(\omega, \kappa), y_i(\omega, \kappa)]$ on the arc can be expressed by 
\begin{equation} \label{eq16}
	\begin{aligned}
		&x_i(\omega, \kappa)=\dfrac{\cos(\theta_i)}{\kappa}+\dfrac{\cos(\bar{\omega})}{\kappa}+x_0, \\ 
		&y_i(\omega, \kappa)=\dfrac{\sin(\theta_i)}{\kappa}+\dfrac{\sin(\bar{\omega})}{\kappa}+y_0, \\ 
		&\theta_i\in[\theta_0-\triangle\theta, \theta_0].
	\end{aligned}
\end{equation}
where $\theta_i$ is the slope of the vector from the turning center $O_t$ to a point on the arc, and $\Delta\theta$ is the range of $\theta_i$. 

The particles are initialized using the parameters of the predicted trajectories. The predicted trajectory $[S^{k}]^*$ can already approximate the minima of $f_{level}(\cdot)$, although $[S^{k}]^*$ is for long-term view and may be sub-optimal locally. Therefore, it is a good idea to use $[S^{k}]^*$ as the initial positions of the particles in PSO-\textit{Level}, given the position of a particle $\boldsymbol{\xi}_i$ represents a trajectory. With this prediction-based initialization, the particles converge fast and avoid local optima traps as they are initialized in close proximity to the global optimum. 
Overall, the initial positions of particles are set as follows. 
\begin{equation} \label{eq7} 
	\begin{aligned} 
		& \boldsymbol{\xi}_i^{0}=[f_{\omega}([S^{1}]^*), f_{\kappa}([S^{1}]^*)]_{|[x_0,y_0]} + \mathcal{N}^2(0, 1), 
	\end{aligned} 
\end{equation}
where $[S^{1}]^*$ is the first planning step on the predicted trajectory $[S^{k}]^*$, $f_{\omega}$ and $f_{\kappa}$ are functions that get the slope and curvature of a trajectory. $\mathcal{N}^2(0, 1)$ is two-dimensional Gaussian noise to encourage exploration. 
{
After initialization, the particles evolve following standard PSO update Eq.~(\ref{eq6}). 
\begin{equation} \label{eq6} 
	\begin{aligned} 
		& \boldsymbol{v}_i^{t+1}=\mu_0\boldsymbol{v}_i^{t}+\mu_1\cdot c_1 (\boldsymbol{p}_i^{t}-\boldsymbol{\xi}_i^{t})+\mu_2\cdot c_2 (\boldsymbol{p}_g^{t}-\boldsymbol{\xi}_i^{t}) \\ 
		& \boldsymbol{\xi}_i^{t+1}=\boldsymbol{\xi}_i^{t}+\boldsymbol{v}_i^{t+1} 
	\end{aligned} 
\end{equation}
where $\boldsymbol{v}_i^t$ and $\boldsymbol{\xi}_i^t$ are the velocity and position of particle $i$ at time $t$, $\boldsymbol{p}_i^{t}$ is the personal best experience of particle $i$ at time $t$ and $\boldsymbol{p}_g^{t}$ is the global best experience of the swarm at time $t$. $\mu_0$ is an inertia weight, $\mu_1$ and $\mu_2$ are random numbers uniformly distributed in $[0, 1]$ independently, and $c_1$ and $c_2$ are acceleration coefficients for cognitive and social components. 
} 

\subsection{Altitude Planning} \label{Alt} 
If the results of trajectory prediction show collisions between the trajectories of UAVs, the involved UAVs will resolve their collisions in altitude planning. 
Let $W$ be the number of UAVs involved in the potential collisions. The search space for PSO-\textit{Alt} is $W$ dimensional. Each dimension corresponds to the altitude adjustment of a UAV. A particle in PSO-\textit{Alt} corresponds to the altitude adjustments of all UAVs $\boldsymbol{\bar{\xi}}^{\ t}=[\triangle Alt_1, \triangle Alt_2, \cdots, \triangle Alt_W]^t$. If $\triangle Alt_i > 0$, UAV-$i$ flies upward for a distance $|\triangle Alt_i|$. If $\triangle Alt_i < 0$, UAV-$i$ flies downward for a distance $|\triangle Alt_i|$. Otherwise, UAV-$i$ does not change altitude. The update of PSO-\textit{Alt} follows the same rules as PSO-\textit{Level} defined by Eq. \eqref{eq6}. 
{
	It is worth noting that there may be scenarios in which multiple solutions yield the same minimal cost. 
	In such instances, the actual solution adopted depends on the randomness in PSO search. 
}

For decentralized altitude planning, we propose an approach in which each UAV independently performs PSO-\textit{Alt} simultaneously with the same search space. The UAVs then exchange their results and corresponding cost values, allowing them to share the same results. The UAVs can ensure they all reach the same solution by adopting the result with the smallest cost. 
The decentralized PSO-\textit{Alt} significantly enhances the robustness of altitude planning for two reasons. Firstly, it ensures that each UAV can access results from others in case of computation failure, provided that the online communication channel is valid. Secondly, it allows for the synchronization of planning results among UAVs. If one UAV finds the optimal solution while others find sub-optimal ones. The optimal solution is broadcast to and adopted by all UAVs. On the other hand, if all UAVs fail to find the optimal solution, the decentralized PSO-\textit{Alt} guarantees that they will unanimously adopt the same result. These properties of PSO-\textit{Alt} avoid potential conflicts that could arise from decentralized planning. 




\section{Performance Evaluation} \label{Simul} 

\subsection{Simulation Setup} 
The trajectories planned by $E^2CoPre$ are solely determined by the relative positions of the UAVs and obstacles, making $E^2CoPre$ orientation invariant. 
For simplicity, we design the simulation environment as a square with dimensions $300\times300$, where the UAV swarm always spawns at the left side and flies towards the right wide. 
Without loss of generality, we evaluate $E^2CoPre$ under two specific scenarios: \textit{Obstacle in Front} and \textit{Obstacle on Side}. 
In the \textit{Obstacle in Front} scenario, the obstacles approach the UAV swarm directly, heading towards the UAVs. On the other hand, the \textit{Obstacle on Side} scenario involves obstacles that approach the UAV swarm from either the left or right side, creating lateral challenges for the swarm. 

In our simulations, we set the initial distance between the swarm and the obstacle to 200 $m$. The swarm's speed is set to $v_s=10\ m/s$, which reflects the typical velocity range of off-the-shelf UAVs such as DJI Matrice and Mavic. 
We treat the obstacles as adversarial UAVs, having the same velocity range as the swarm. Each UAV in the swarm, including its payload, has a weight of $1\ Kg$. 
The UAV's sensing range is set to $100\ m$. The collision avoidance process is activated once the minimum U2O distance falls below $50\ m$. We set the length of one planning step to the distance UAVs flying in $1\ second$, which corresponds to the swarm's speed $v_s$. We predict ten steps in trajectory prediction, considering the limited sensing range of the UAVs. To maintain safety, the protection bubble radius is set to $d_{safe} = 20\ m$. Additionally, we set the hard constraints for U2O and U2U distances to $d_{obs}=10\ m$ and $d_{u2u}=5\ m$, respectively. 
The UAV swarm is organized in a circular formation, with the swarm members evenly spaced at a radius of $\tau$ ($\tau\in[1, 20]\ m$). The size of the swarm $N$ varies from 2 to 10, allowing us to evaluate $E^2CoPre$ under different swarm configurations. The cognitive and social components in PSO are both set to 0.5 for all algorithms. 

{
On the other hand, two types of obstacles: mass point obstacles and shaped obstacles are considered, without losing generality. Mass point obstacles are relatively small compared to the swarm and can thus be simplified as mass points. In contrast, shaped obstacles are comparable in size to the swarm, making it necessary to consider their shapes. For simplicity, a shaped obstacle is represented as a series of mass points that are randomly distributed in its vicinity. The environment field is constructed on the series of mass points, and it's required that the UAVs should not collide with any of these mass points. Fig. \ref{figShape} illustrates how a shaped obstacle is modeled based on a mass point obstacle, and the difference between their potential fields, where the obstacles and their potenfiel fields are depicted using red dots and blue circles, respectively. 
\begin{figure}[b]
	\centering
	\includegraphics[width=0.30\textwidth]{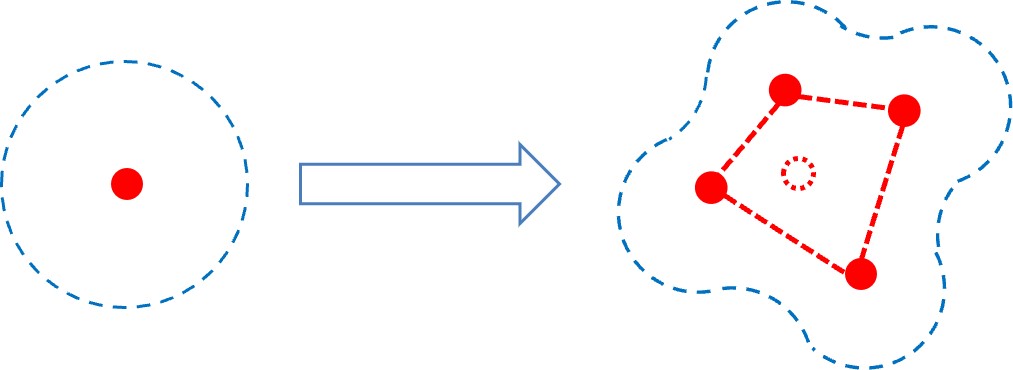} 
	\caption{Mass point obstacle and shaped obstacle. }
	\label{figShape}
\end{figure}
}

We conduct comparative analysis of $E^2CoPre$ with conventional metaheuristic-based and MARL-based algorithms. For metaheuristic-base methods, we select three different schemes that also utilize the combination of APF and PSO for collision avoidance as follows: 
\begin{itemize}
	\item $FFPSO$ \cite{hyb5}: A virtual force term that repels UAVs from each other is added to the velocity update equation in waypoint-based PSO. 
	The cost function is simply the negative of the distance between the particle and its destination;  
	\item $PPSO$ \cite{hyb6}: A smoothing field is introduced to smooth UAVs' trajectories. Then, a waypoint-based PSO is adopted to find the trajectories on the smoothing field. The cost of particles is just the field intensity; 
	\item $E^2Coop$ \cite{huang2021}: A shared environment field provides implicit coordination. UAVs plan for their optimal trajectories using trajectory-based PSO in a two-dimensional space. U2U collisions are avoided using virtual forces. 
\end{itemize} 
These schemes differ from $E^2CoPre$ by either adopting waypoint-based PSO, using simple cost functions that overlook UAVs' energy consumption, or neglecting swarm members' cooperation. Additionally, none of these schemes incorporate trajectory prediction in collision avoidance or PSO's initialization. Comparing with these schemes helps reveal the distinct advantages of $E^2CoPre$ and highlight the benefits of its key contributions. All simulations are performed on 11$th$ Gen Intel Core i5-1135G7 processor with 16.0 $GB$ RAM. 

{
On the other hand, we compare with \textit{CODE} \cite{huang2022}, the state-of-art MARL algorithm for collision avoidance for UAV swarms, to validate the advantage of $E^2CoPre$ over MARL-based algorithms in terms of safety and collision rates. Different from other MARL-based methods such as independent Q learning (IQL) \cite{iql}, multi-agent deep deterministic policy gradient (MADDPG) \cite{MADDPG}, value decomposition networks (VDN) \cite{VDN}, QMIX \cite{Qmix}, counterfactual multi-agent policy gradients (COMA) \cite{COMA}, and Shapley Q value \cite{shapley_oda1}, \textit{CODE} considers global information in training, applies to continuous action space such as UAV control commands, and has low computational complexity. 
The training environments and network structures follow the original settings in \textit{CODE} \cite{huang2022}. The training is performed on NVIDIA Quadro P6000 GPU. 
}

\subsection{Simulation Results} 
The results on energy consumption and safety of $E^2CoPre$ are reported in this section. One hundred trajectories are collected in each experiment. The computational complexity of metaheuristics-based algorithms in trajectory planning, such as $E^2CoPre$, is difficult to quantify as it involves two cost functions, environment modeling and trajectory prediction based on calculus of variation. We use computation time as an alternative measure to quantify the computational complexity of $E^2CoPre$. The average computation time of $E^2CoPre$, $FFPSO$, $PPSO$, and $E^2Coop$ are summarized in Table \ref{table_time}. 

\begin{table}[h]
	\centering
     \caption{Computation time}
	\begin{tabular}{||c c c c||} 
		\specialrule{.2em}{.1em}{.1em}
		$E^2CoPre$ & $FFPSO$ & $PPSO$ & $E^2Coop$ \\ 
		\hline 
		$0.48\pm0.05$   & $0.05 \pm 0.005$   & $0.025 \pm 0.0024$   & $0.92\pm0.07$ \\ [0.5ex] 
		\specialrule{.2em}{.1em}{.1em}
	\end{tabular}
	
	\label{table_time}
\end{table}

\subsubsection{Energy Consumption} \label{exp-energy}
The energy consumption of UAVs is calculated following Eq. \eqref{eq:energy} and can be simplified to 
\begin{equation*} 
	\begin{aligned} 
		E &= E_n + E_{len} + E_{comms} \\
		&= \oint \bar{P}_{n}\cdot m\cdot|S''(p)| + \bar{P}_{len}\cdot mg\cdot[|S(p)| + \triangle Alt(S(p))] \\
		&+ \bar{P}_{comms}\cdot|S(p)|dp, \\ 
	\end{aligned} 
\end{equation*} 
where $\oint dp$ denotes the integration over the entire trajectory. Hence the notation $k$ is emitted. $E_n$, $E_{len}$ and $E_{comms}$ are turning-dependent, length-dependent and communication-dependent energy terms. $m$ and $g$ are the UAV's inclusive weight and gravitational acceleration, respectively. 
$\triangle Alt(p)$ is the altitude change of the trajectory. 
Only $E_n$ and $E_{len}$ are relevant to $m$. 
$\bar{P}_{n}$, $\bar{P}_{len}$ and $\bar{P}_{comms}$ are the average power per unit curvature and length, respectively. Because the energy consumed in wireless communication is usually much smaller than that generating thrusts, we have $\bar{P}_{comms}\ll \bar{P}_{n}$, $\bar{P}_{comms}\ll \bar{P}_{len}$. Therefore, the three coefficients are set to $\bar{P}_{len}=\bar{P}_{len}=1$ and $\bar{P}_{comms}=0.01$. 


\begin{figure}[bt]
	\centering
	\subfloat[]{\includegraphics[width=0.5\linewidth]{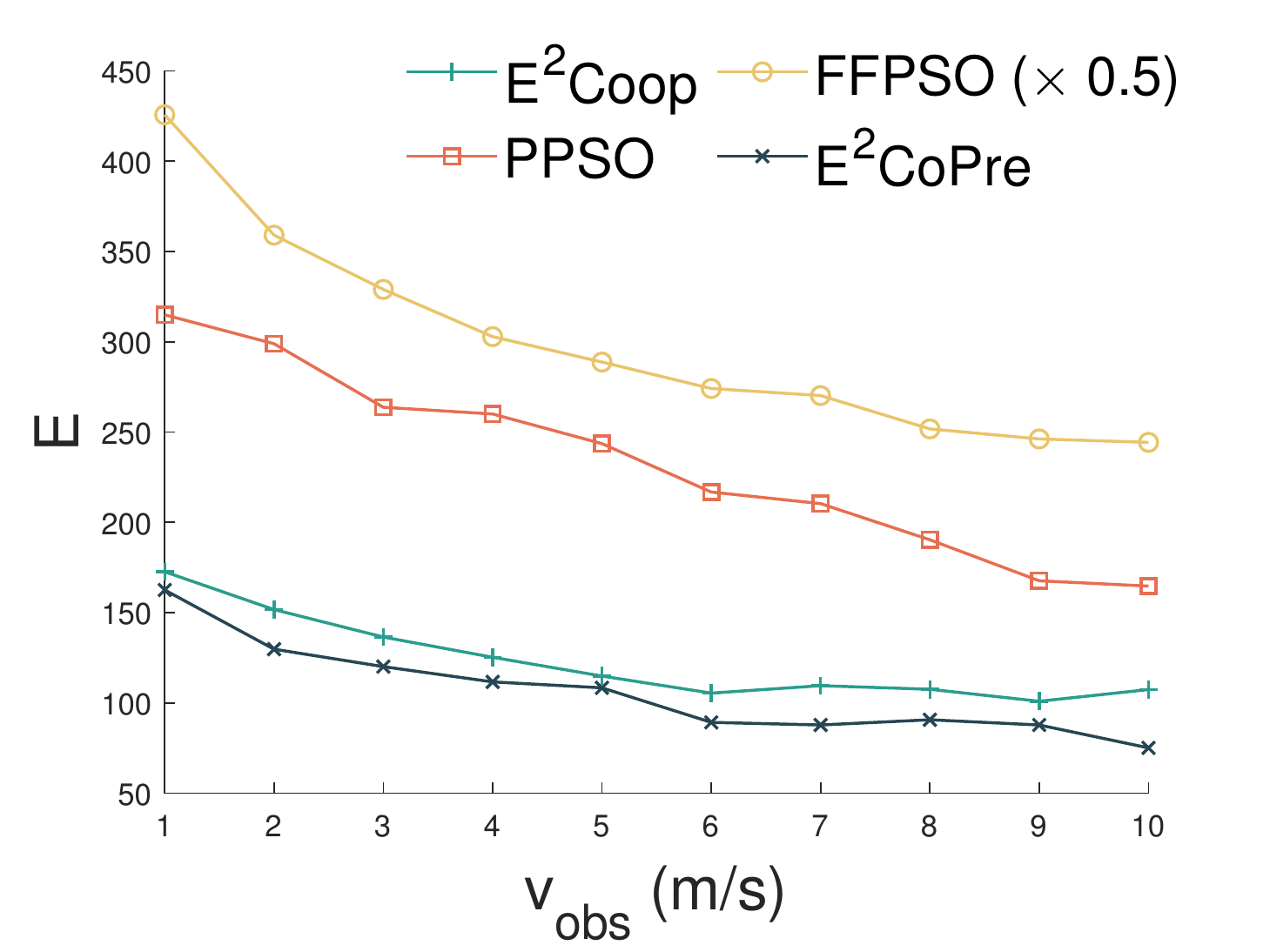}%
		\label{fig16a}}
	\hfil
	\subfloat[]{\includegraphics[width=0.5\linewidth]{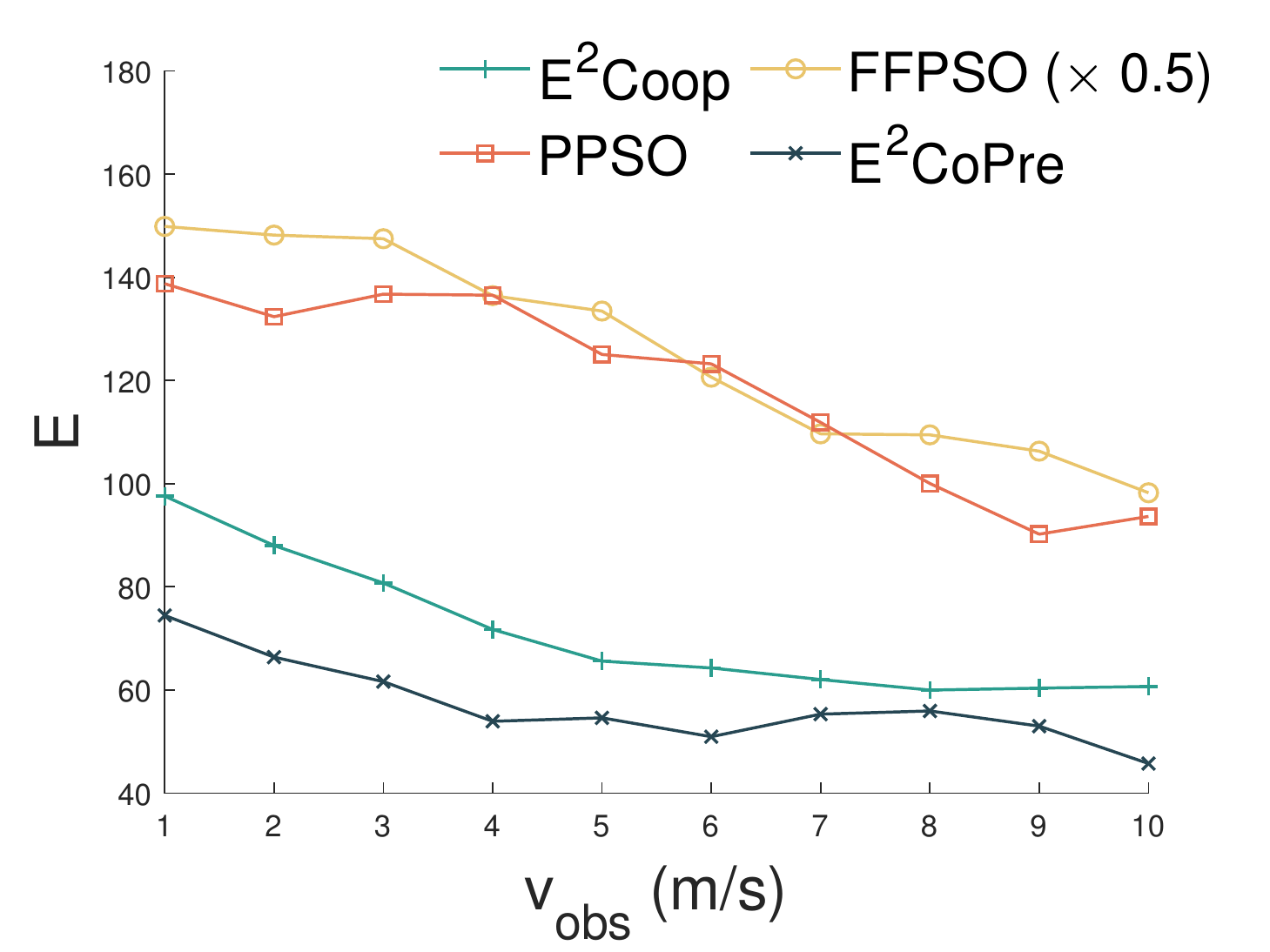}%
		\label{fig16b}}
	\hfil
	\subfloat[]{\includegraphics[width=0.5\linewidth]{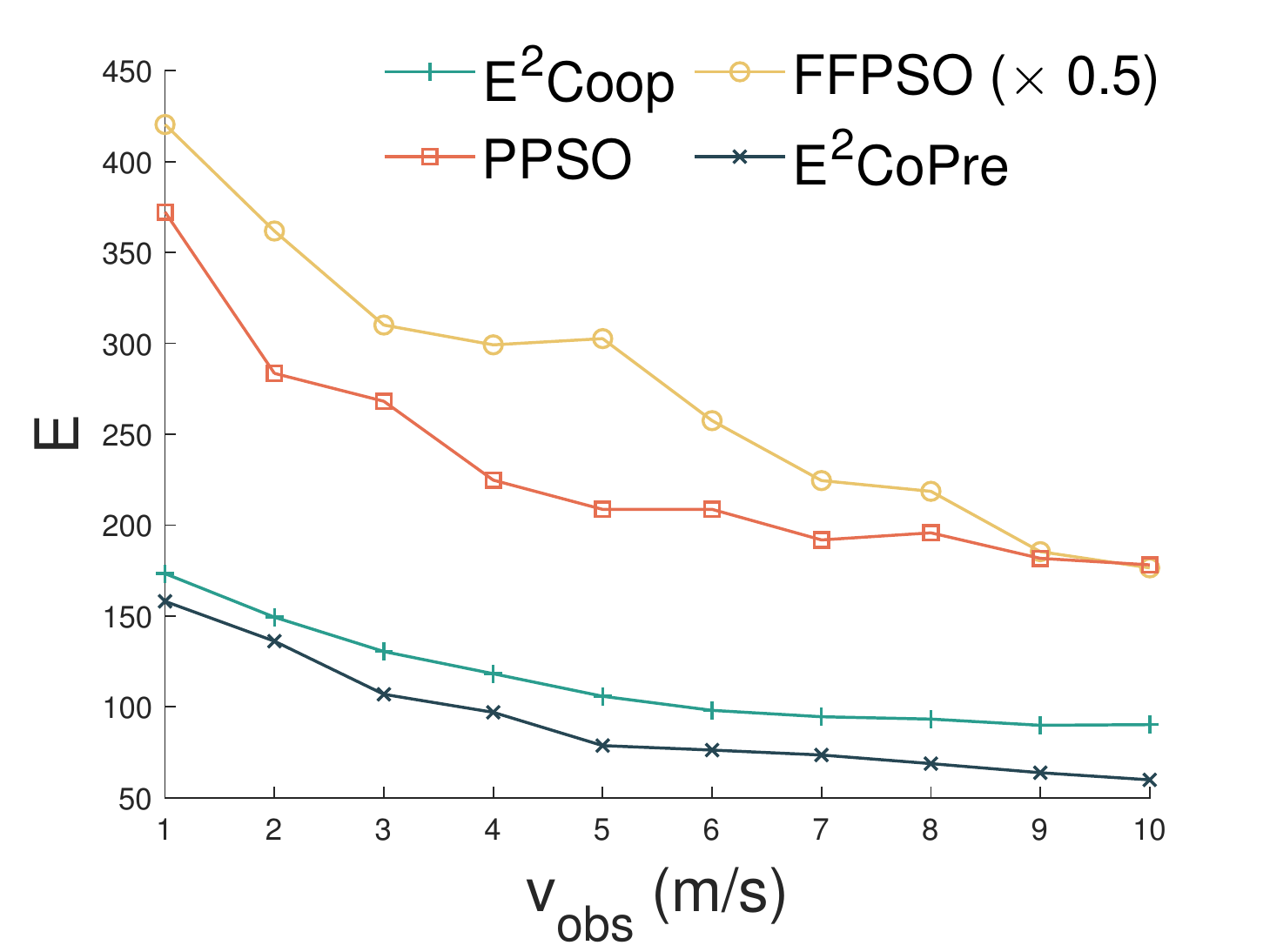}%
		\label{fig16c}}
	\hfil
	\subfloat[]{\includegraphics[width=0.5\linewidth]{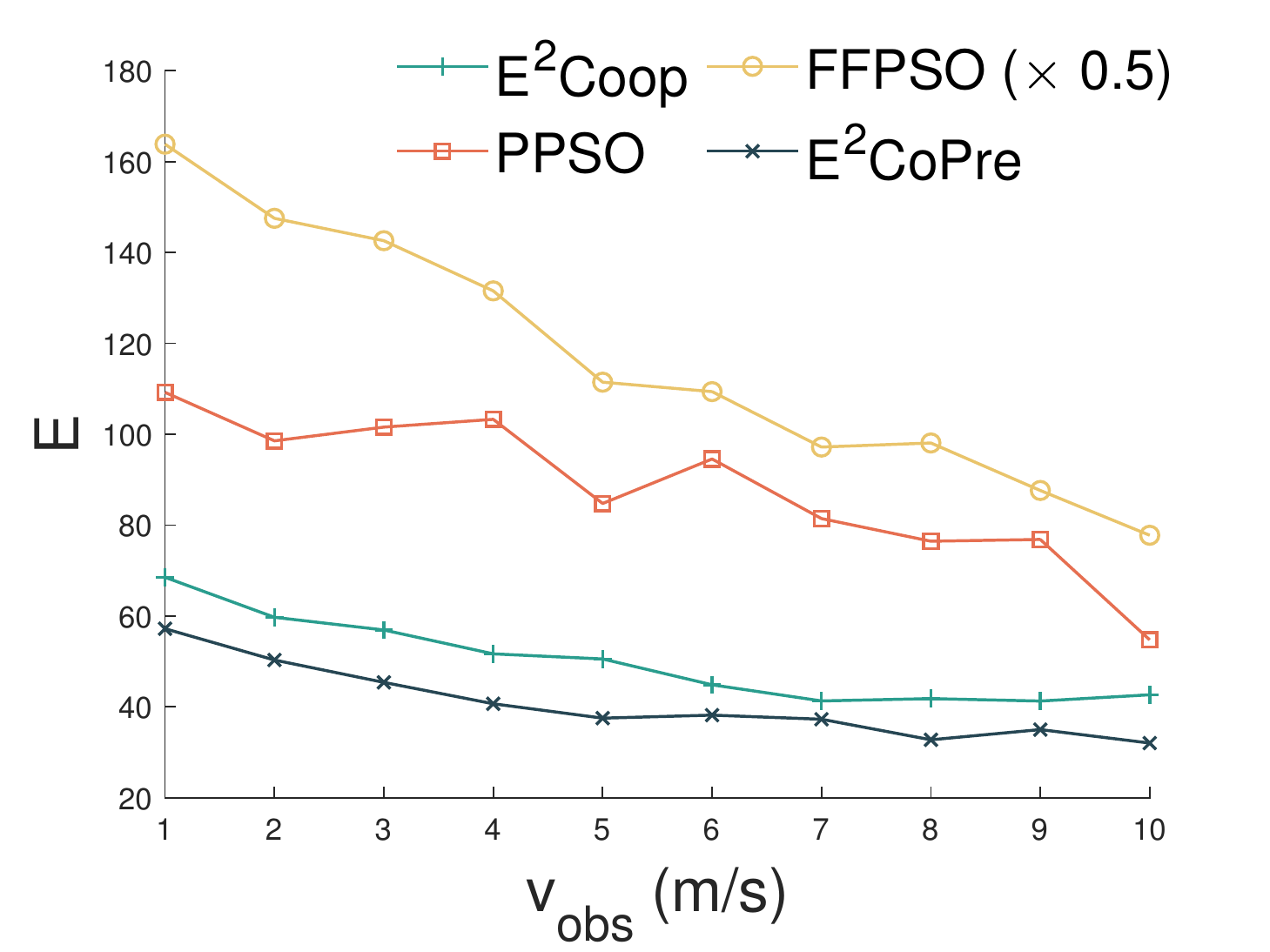}%
		\label{fig16d}}
	\hfil
	\caption{The energy consumption with different obstacle velocities. (a) \textit{Obstacle in Front} with mass point obstacles. (b) \textit{Obstacle on Side} with mass point obstacles. (c) \textit{Obstacle in Front} with shaped obstacles. (d) \textit{Obstacle on Side} with shaped obstacles. } 
	\label{fig16}
\end{figure}

The results on energy consumption are shown in Fig. \ref{fig16}, where the data of $FFPSO$ is re-scaled by 0.5 for easy comparison. 
Generally, the energy consumption of UAVs in scenario \textit{Obstacle on Side} is lower than \textit{Obstacle in Front} as their trajectories have fewer intersections with the obstacles and hence require fewer maneuvers from fewer UAVs to avoid collisions. $FFPSO$ has the highest energy consumption as it avoids collisions purely based on virtual forces and generates zig-zag trajectories that require significant turning-dependent energy. On the other hand, $PPSO$ is more energy-efficient than $FFPSO$ as its trajectories have fewer turnings thanks to the smoothing field in $PPSO$. Furthermore, $E^2Coop$ demonstrates even higher energy efficiency due to its implicit coordination provided by the shared APF and the smoothing effect achieved through trajectory-based PSO.
However, $E^2CoPre$ has the lowest energy consumption among all four schemes in both scenarios for the following reasons: 
\begin{itemize}
	\item U2U collisions are resolved effectively in three-dimensional space based on altitude adjustment. By considering altitude as an additional dimension for collision avoidance, $E^2CoPre$ achieves considerable improvement on energy consumption compared to $E^2Coop$ which uses virtual forces to address U2U collisions; 
	\item Collisions are identified and resolved proactively by trajectory prediction, which effectively minimizes unnecessary maneuvers that would consume additional energy; 
	\item PSO is initialized with trajectory prediction in $E^2CoPre$, which leads to rapid and accurate convergence. This prediction-based initialization enables the particles to start close to the global optimum, avoiding getting trapped in local optima. 
\end{itemize}
%

{
In scenario \textit{Obstacle in Front}, $E^2CoPre$ can save 83\%, 54\% and 15\% energy compared to $FFPSO$, $PPSO$ and $E^2Coop$ in average, respectively, when avoiding mass point obstacles, and 67\%, 61\% and 21\%, respectively, when avoiding shaped obstacles. In scenario \textit{Obstacle on Side}, $E^2CoPre$ can save 78\%, 51\% and 19\% energy compared to $FFPSO$, $PPSO$ and $E^2Coop$ in average, respectively, when avoiding mass point obstacles, and 65\%, 53\% and 19\%, respectively, when avoiding shaped obstacles. 
}

\begin{figure}[bt]
	\centering
	\subfloat[]{\includegraphics[width=0.5\linewidth]{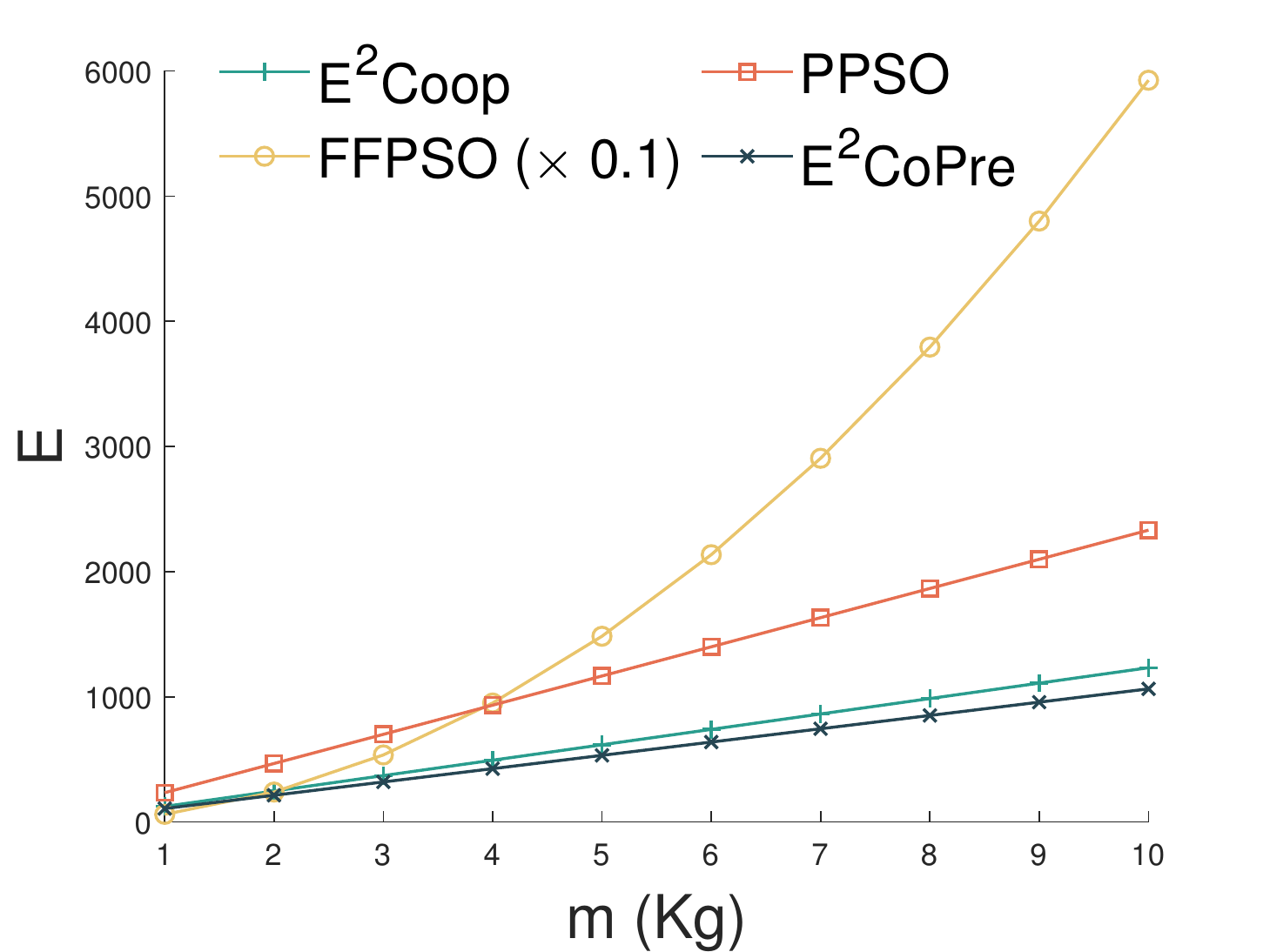}%
		\label{fig19a}}
	\hfil
	\subfloat[]{\includegraphics[width=0.5\linewidth]{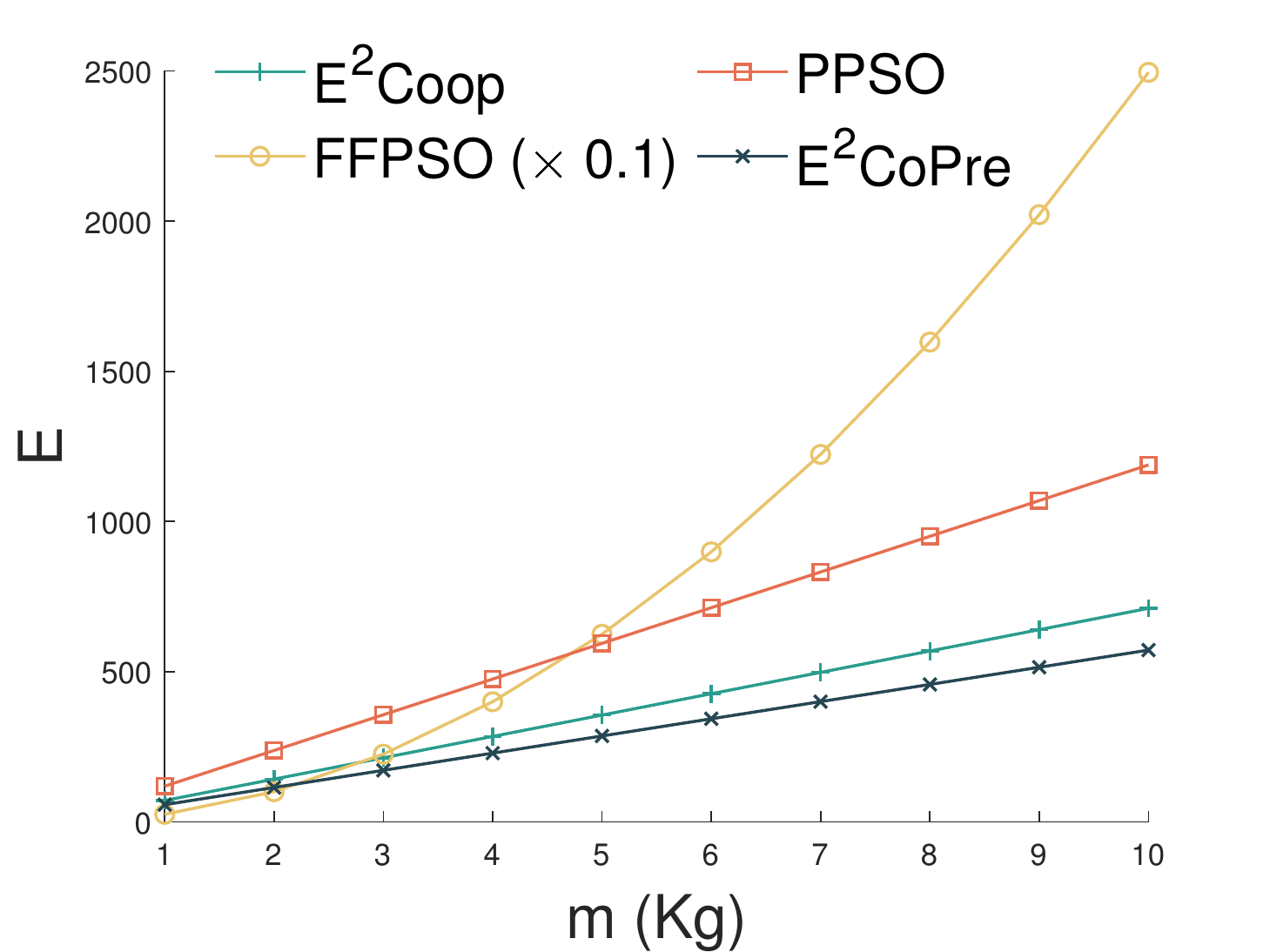}%
		\label{fig19b}}
	\caption{The energy consumption with different UAV weights. (a) \textit{Obstacle in Front}. (b) \textit{Obstacle on Side}. } 
	\label{fig19}
\end{figure}

Additionally, we conduct tests to evaluate the average energy consumption across various UAV weights $(m)$. In our simulations, we vary $m$ within the range of $1\ Kg$ to $10\ Kg$, with a step of $1\ Kg$, considering industry-level UAVs, such as DJI MATRICE 600, are capable of carrying a maximum takeoff weight of 9 $Kg$. 
For simplicity, only mass point obstacles are considered. The results are presented in Fig. \ref{fig19}, where the energy consumption of $FFPSO$ has been re-scaled by 0.1 for better visualization. $FFPSO$ is the most sensitive to UAV weights as it generates the most zig-zag trajectories. 
The average energy consumption of $E^2CoPre$ outperforms the other baseline schemes across all UAV weights as it addresses U2U collisions in three-dimensional space, generating the most smooth trajectories. Even though $E^2CoPre$ requires altitude changes, PSO-\textit{Alt} ensures the energy consumption is minimized. 
For the rest of this section, we let $m=1\ Kg$ in all simulations.

\subsubsection{Safety} \label{exp-safety}

Safety is measured by the minimum U2O distances and the minimum U2U distances. For shaped obstacles, the distances are calculated from the UAVs to their surface. 
\begin{figure}[bt]
	\centering
	\subfloat[]{\includegraphics[width=0.5\linewidth]{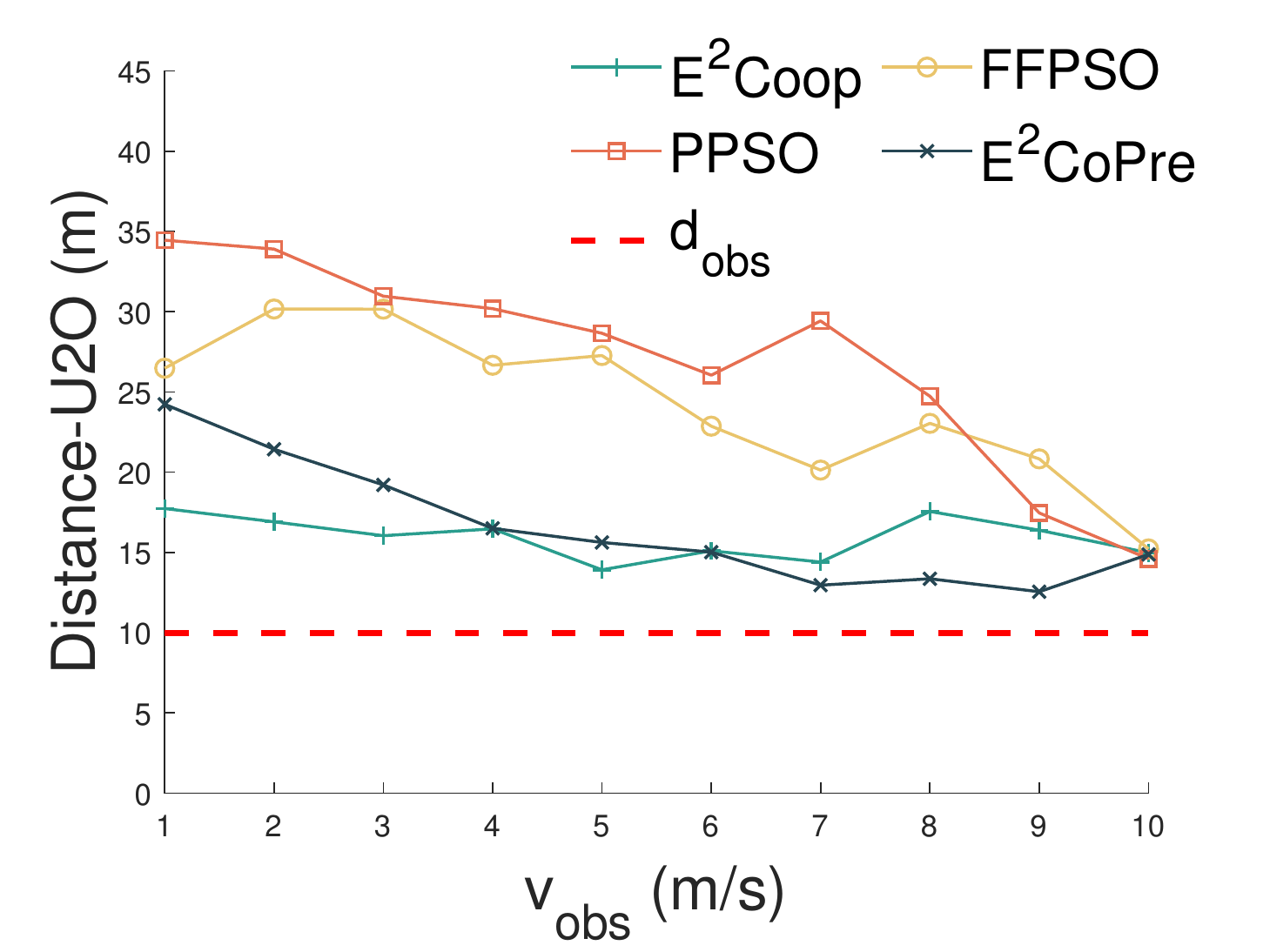}%
		\label{fig15a}}
	\hfil
	\subfloat[]{\includegraphics[width=0.5\linewidth]{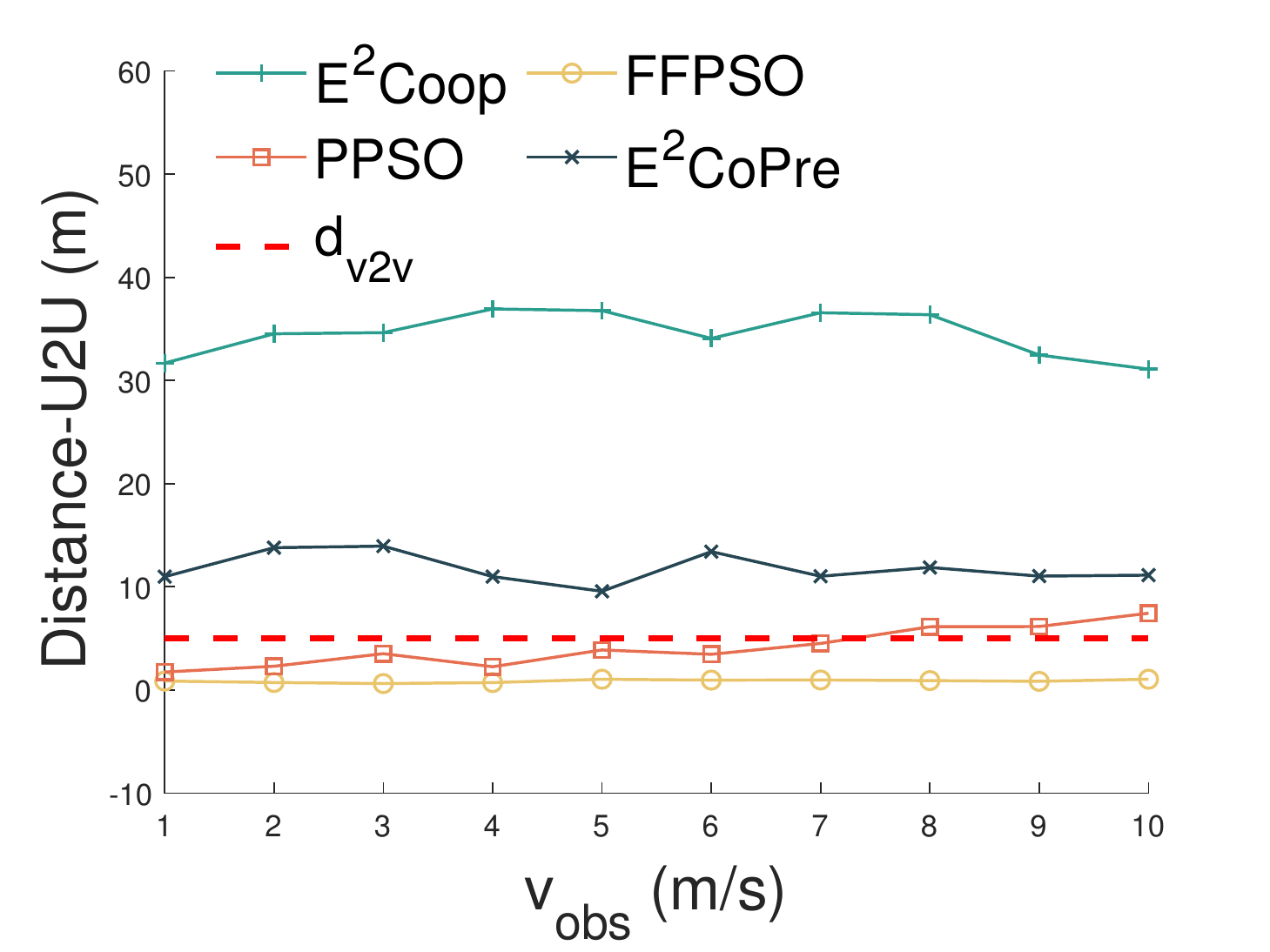}%
		\label{fig15b}}
	\hfil
	\subfloat[]{\includegraphics[width=0.5\linewidth]{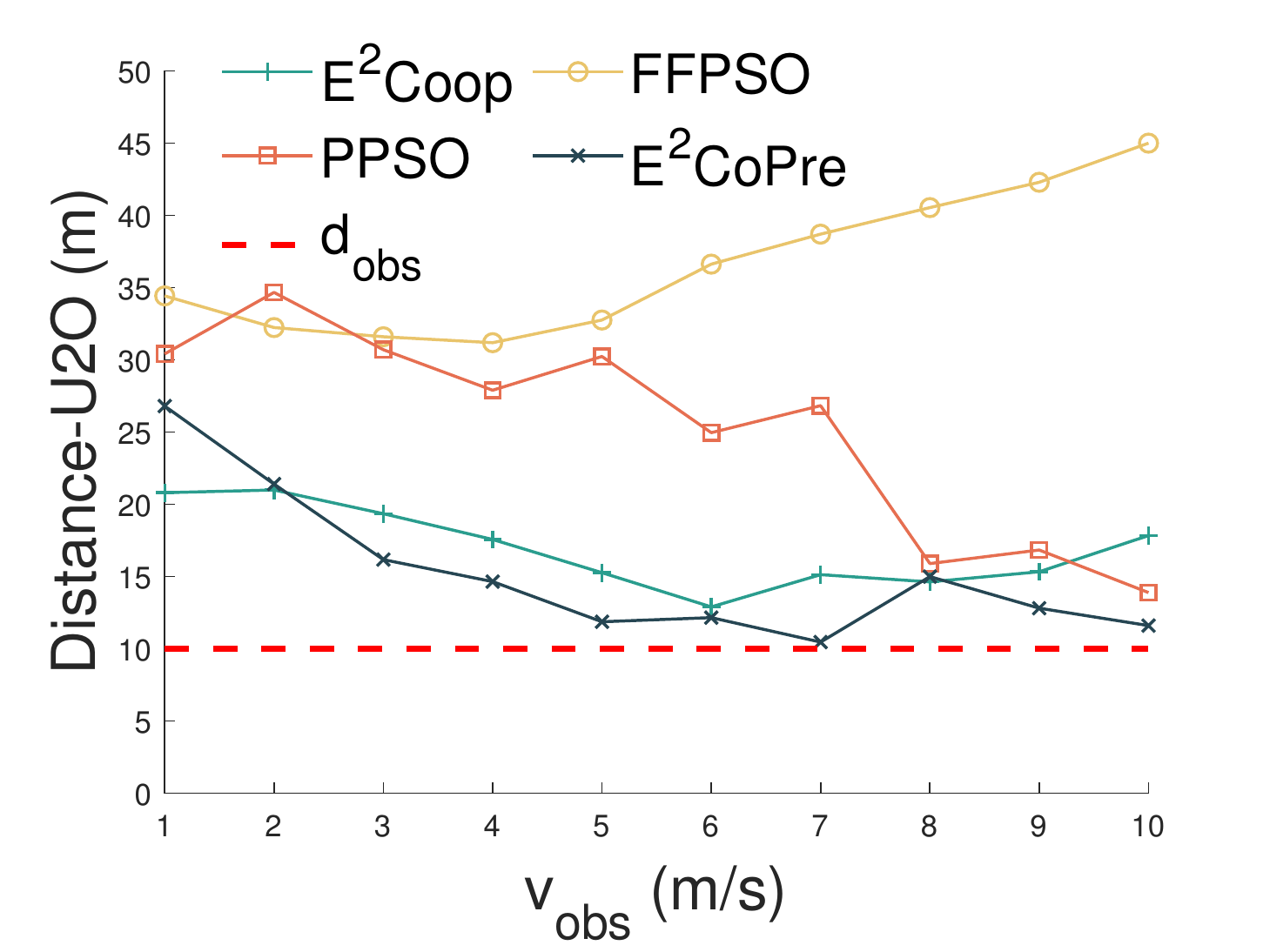}%
		\label{fig15c}}
	\hfil
	\subfloat[]{\includegraphics[width=0.5\linewidth]{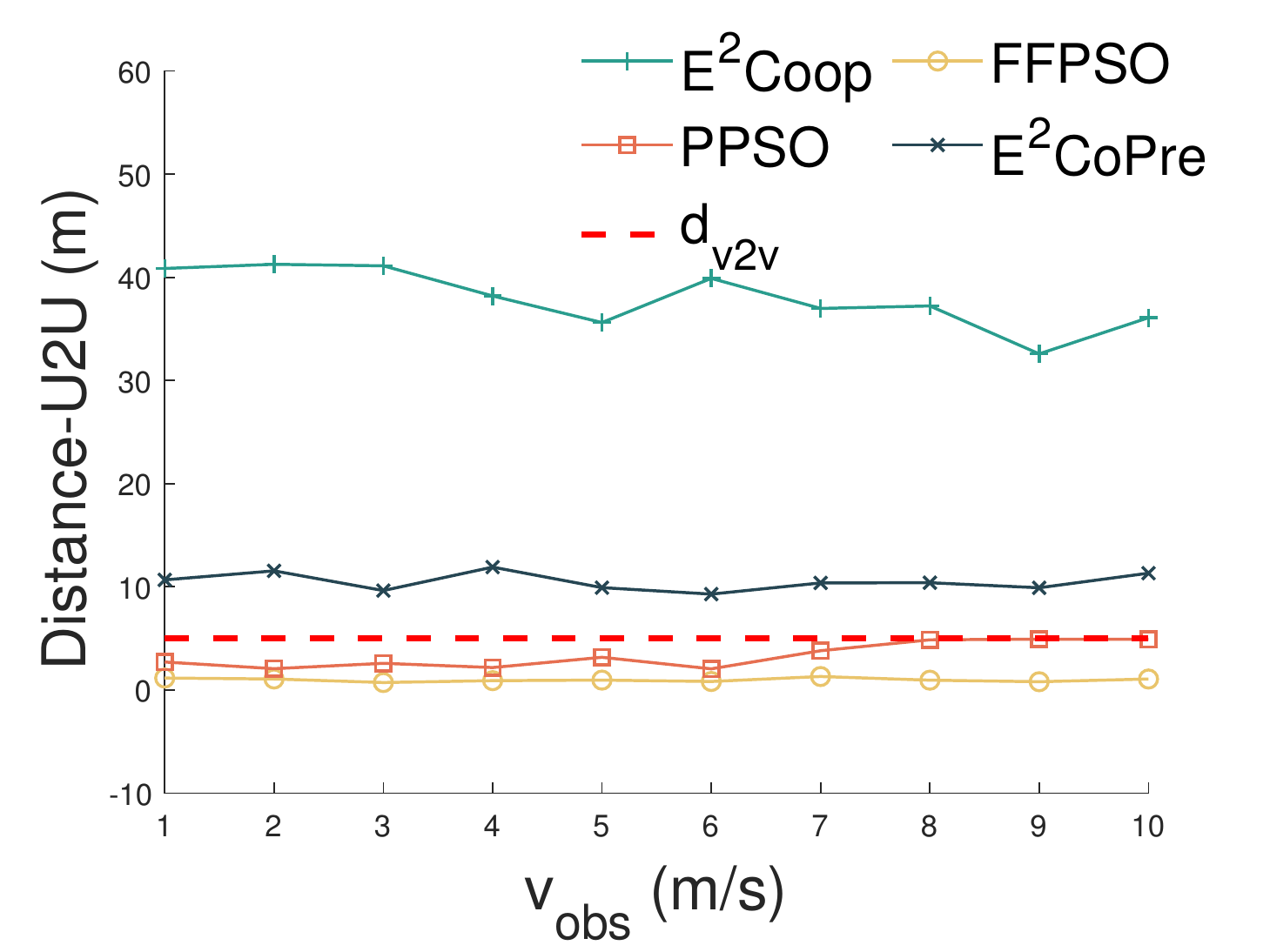}%
		\label{fig15d}}
	\hfil
	\subfloat[]{\includegraphics[width=0.5\linewidth]{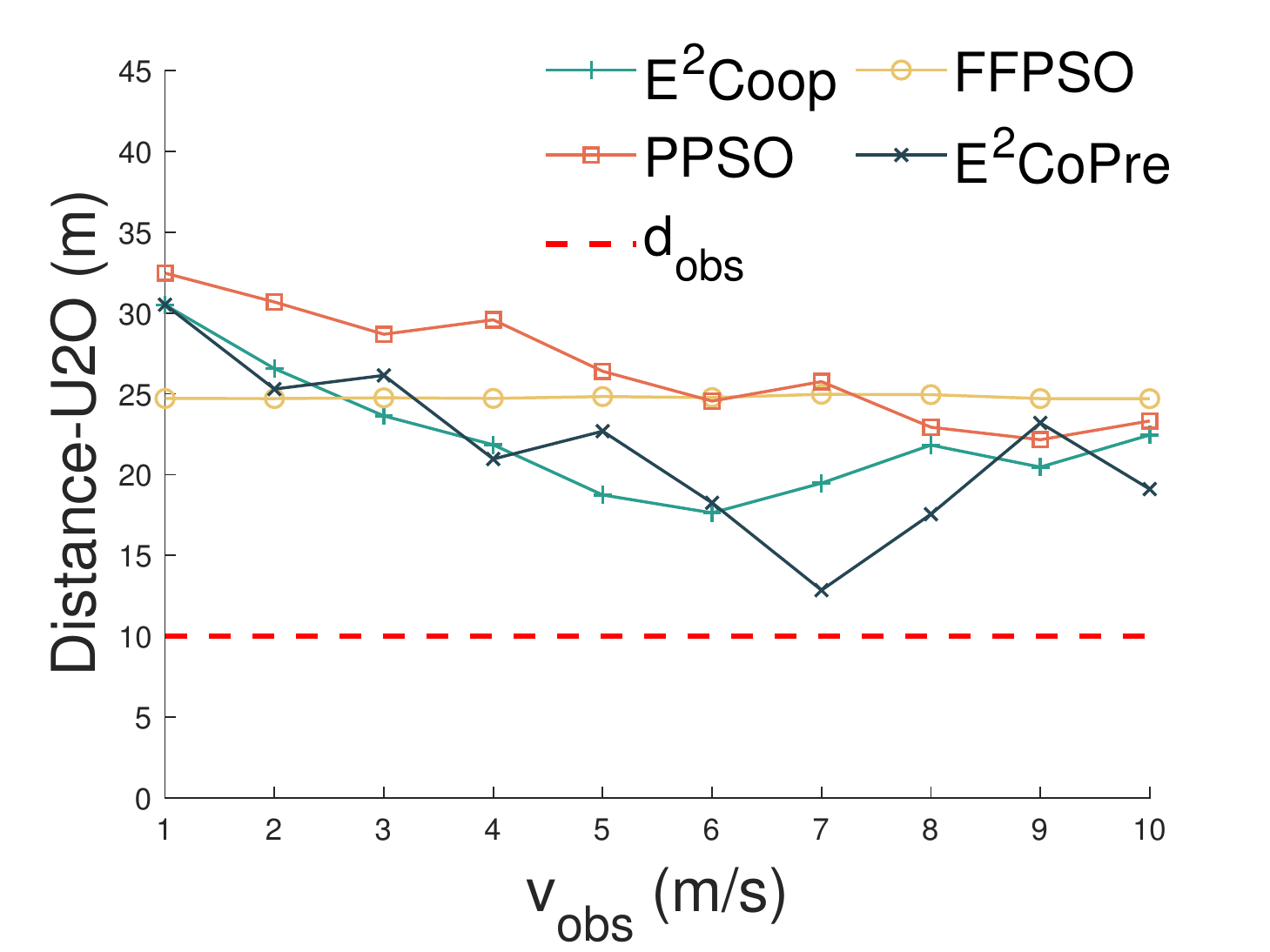}%
		\label{fig15e}}
	\hfil
	\subfloat[]{\includegraphics[width=0.5\linewidth]{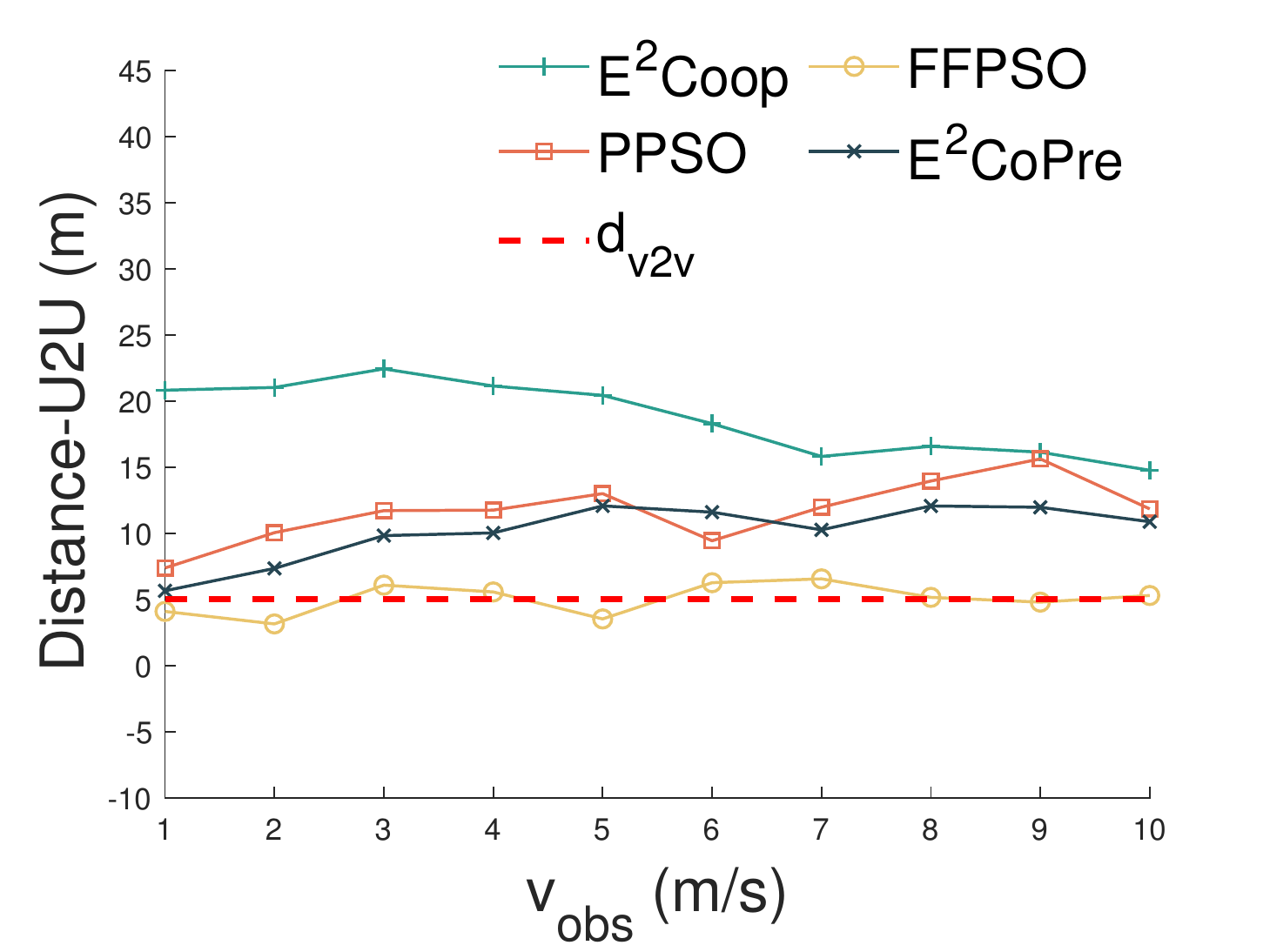}%
		\label{fig15f}}
	\hfil
	\subfloat[]{\includegraphics[width=0.5\linewidth]{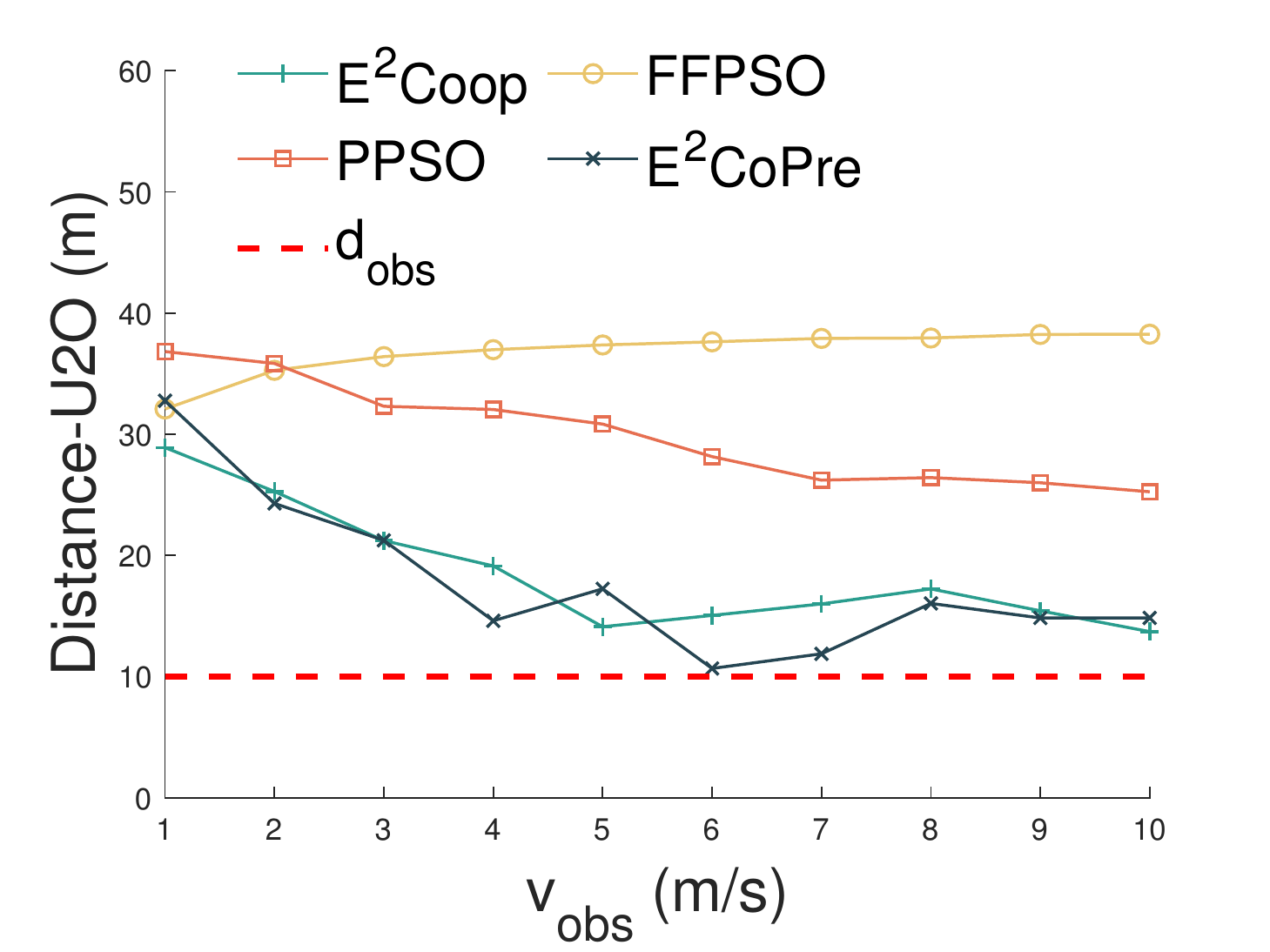}%
		\label{fig15g}}
	\hfil
	\subfloat[]{\includegraphics[width=0.5\linewidth]{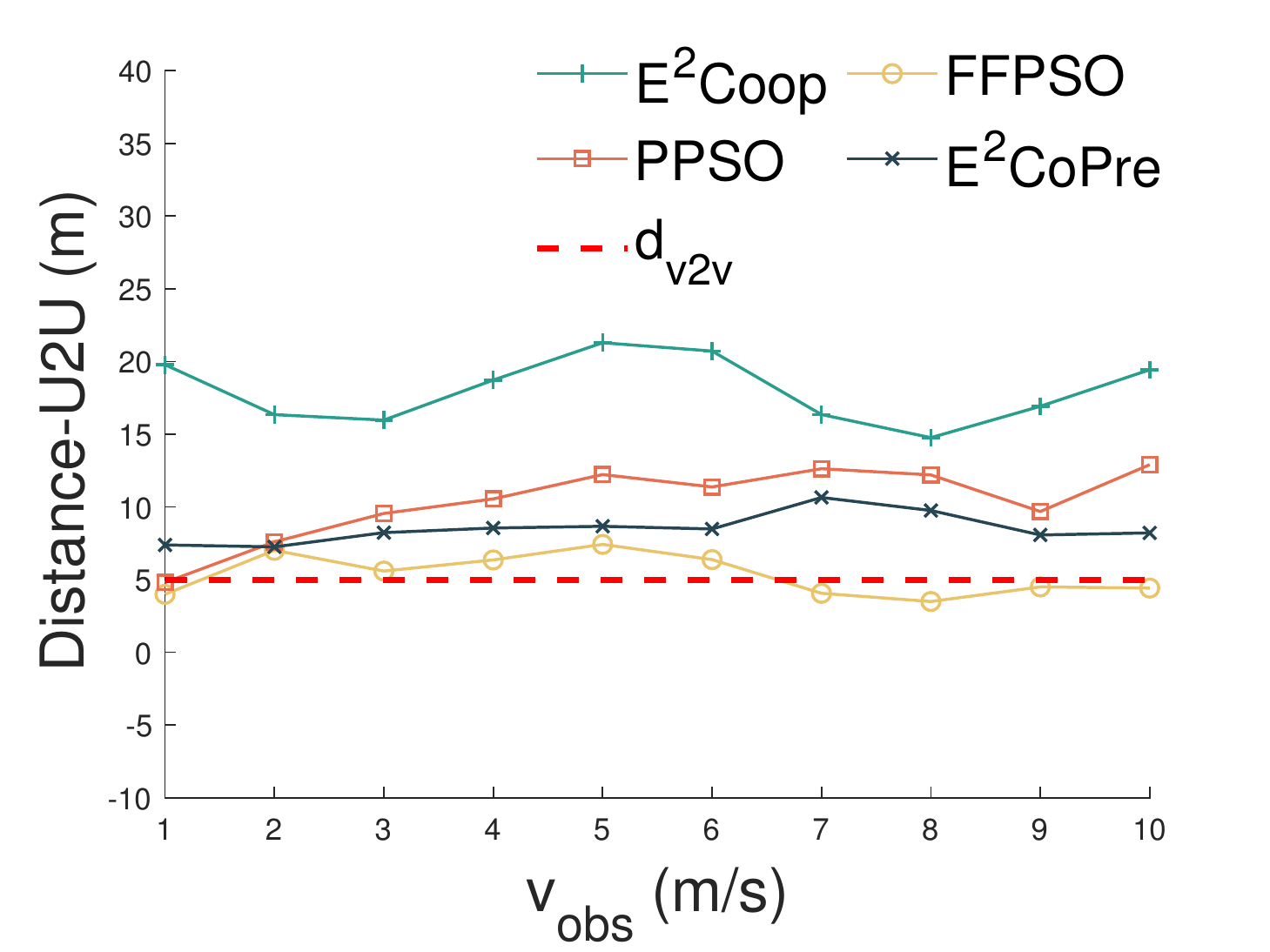}%
		\label{fig15h}}
	\caption{The minimum U2O and U2U distances with different obstacle velocities. (a) \textit{Obstacle in Front} U2O with mass point obstacles. (b) \textit{Obstacle in Front} U2U with mass point obstacles. (c) \textit{Obstacle on Side} U2O with mass point obstacles. (d) \textit{Obstacle on Side} U2U with mass point obstacles. (e) \textit{Obstacle in Front} U2O with shaped obstacles. (f) \textit{Obstacle in Front} U2U with shaped obstacles. (g) \textit{Obstacle on Side} U2O with shaped obstacles. (h) \textit{Obstacle on Side} U2U with shaped obstacles. }
	\label{fig15}
\end{figure}

When avoiding mass point obstacles (Fig. \ref{fig15a}, \ref{fig15b}, \ref{fig15c} and \ref{fig15d}), the minimum U2O distances decrease with the obstacle's velocity in scenario \textit{Obstacle in Front}, due to the reduced reaction time for UAVs as obstacles fly fast. 
In contrast, the minimum U2O distances in all algorithms show a decreasing trend with the obstacle's velocity in scenario \textit{Obstacle on Side}, except in the case of \textit{FFPSO}, where the trend initially decreases, followed by an increase with the obstacle's velocity. This special trend can be attributed to the collision avoidance mechanism of \textit{FFPSO}. \textit{FFPSO} avoids collisions by virtual forces generated by the obstacles pushing the UAVs away from them. As a result, the obstacles only impact the UAVs' trajectories when they are within a certain distance from the UAVs. Therefore, when the obstacles fly at high velocities, they quickly bypass the UAVs' trajectories and move in different directions. As a result, the obstacles' virtual forces have little impact on the UAVs, and the UAVs can maintain larger distances from the obstacles. 

On the other hand, \textit{FFPSO} and \textit{PPSO} both have the minimum U2U distances fall below the threshold $d_{u2u}$ in both scenarios, indicating the presence of U2U collisions. 
Moreover, the minimum U2U distance in $E^2CoPre$ is notably smaller than that in $E^2Coop$, showcasing the superiority of PSO-\textit{Alt} over the virtual forces-based approach in U2U collision avoidance. 
The analysis for mass point obstacles holds for shaped obstacles (Fig. \ref{fig15e}, \ref{fig15f}, \ref{fig15g} and \ref{fig15h}).

\subsubsection{Comparison With MARL-Based Methods} \label{exp-MARL}

{
One hundred trajectories of three UAVs avoiding two obstacles are collected for each algorithm in scenario \textit{Obstacle in Front}, with the swarm and obstacles flying at a speed of $10\ m/s$ and $5\ m/s$, respectively. 
The numerical results on safety and energy efficiency are shown in Table \ref{table_RL}. Table \ref{table_RL} shows the minimum distances of non-collision trajectories, as the path generation in $CoDe$ terminates when collisions occur. 
The numerical results show that although the trajectories of \textit{CoDe} are 30\% more energy efficient than $E^2CoPre$, it suffers a collision rate of 3\% in execution, while $E^2CoPre$ ensures 0 collision rate. 
}

\begin{table}[h]
	\centering
 	\caption{Numerical results of $E^2CoPre$ and $CoDe$. }
	\begin{tabular}{||c |c c c c||} 
		\specialrule{.2em}{.1em}{.1em}
		& Energy & U2O 	   & U2U 	  & Collision \\ 
		& Cost & Distance & Distance & Rate \\ [0.5ex] 
		\hline 
		$E^2CoPre$ 	   & $108.34\pm2.1$   & 15.62   & 10.98   & 0 \\ [0.5ex] 
		\textit{CoDe}  & $75.93\pm31.8$ & {19.83} & {28.26} & {3\%} \\ [0.5ex] 
		\specialrule{.2em}{.1em}{.1em}
	\end{tabular}

	\label{table_RL}
\end{table}

\subsubsection{Parameter Analysis} 
Impacts of important parameters, $d_{safe}$, $d_{u2u}$, and $\lambda_1$ in $E^2CoPre$ are shown in Fig.~\ref{fig17}, and are summarized as follows: 

\begin{figure}
	\centering
	\subfloat[]{\includegraphics[width=0.5\linewidth]{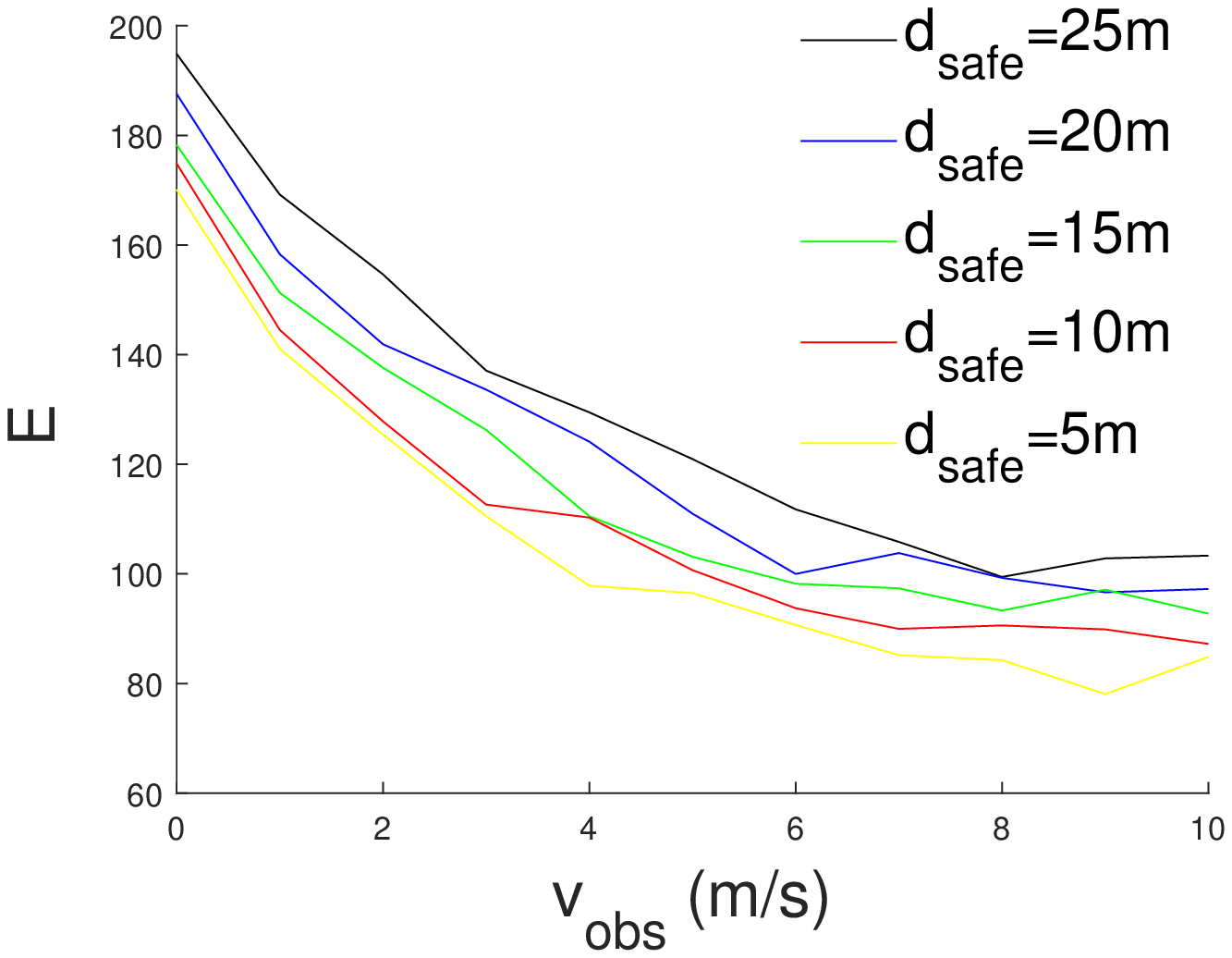}%
		\label{17a}}
	\hfil
	\subfloat[]{\includegraphics[width=0.5\linewidth]{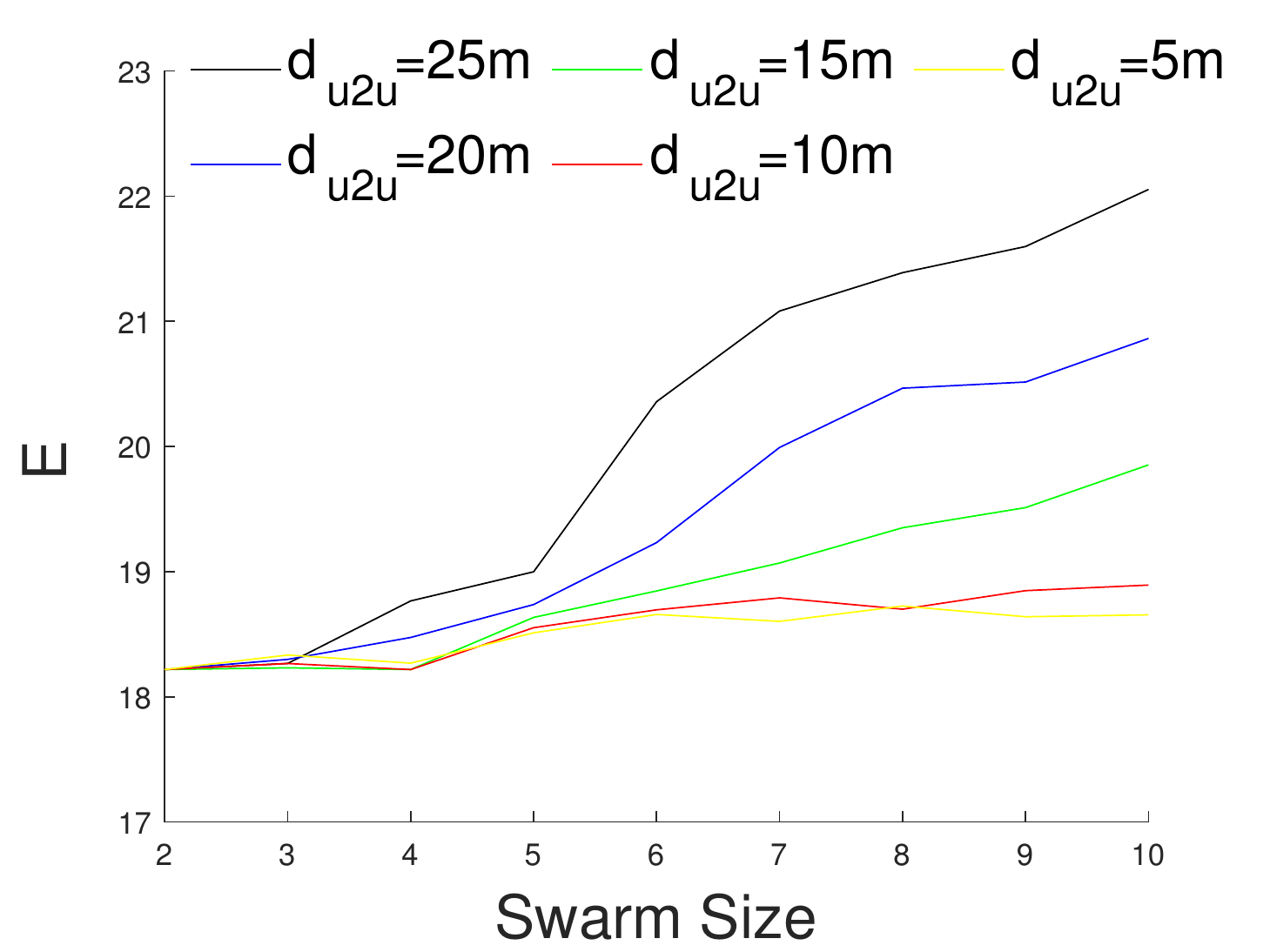}%
		\label{17b}}
	\hfil
	\subfloat[]{\includegraphics[width=0.5\linewidth]{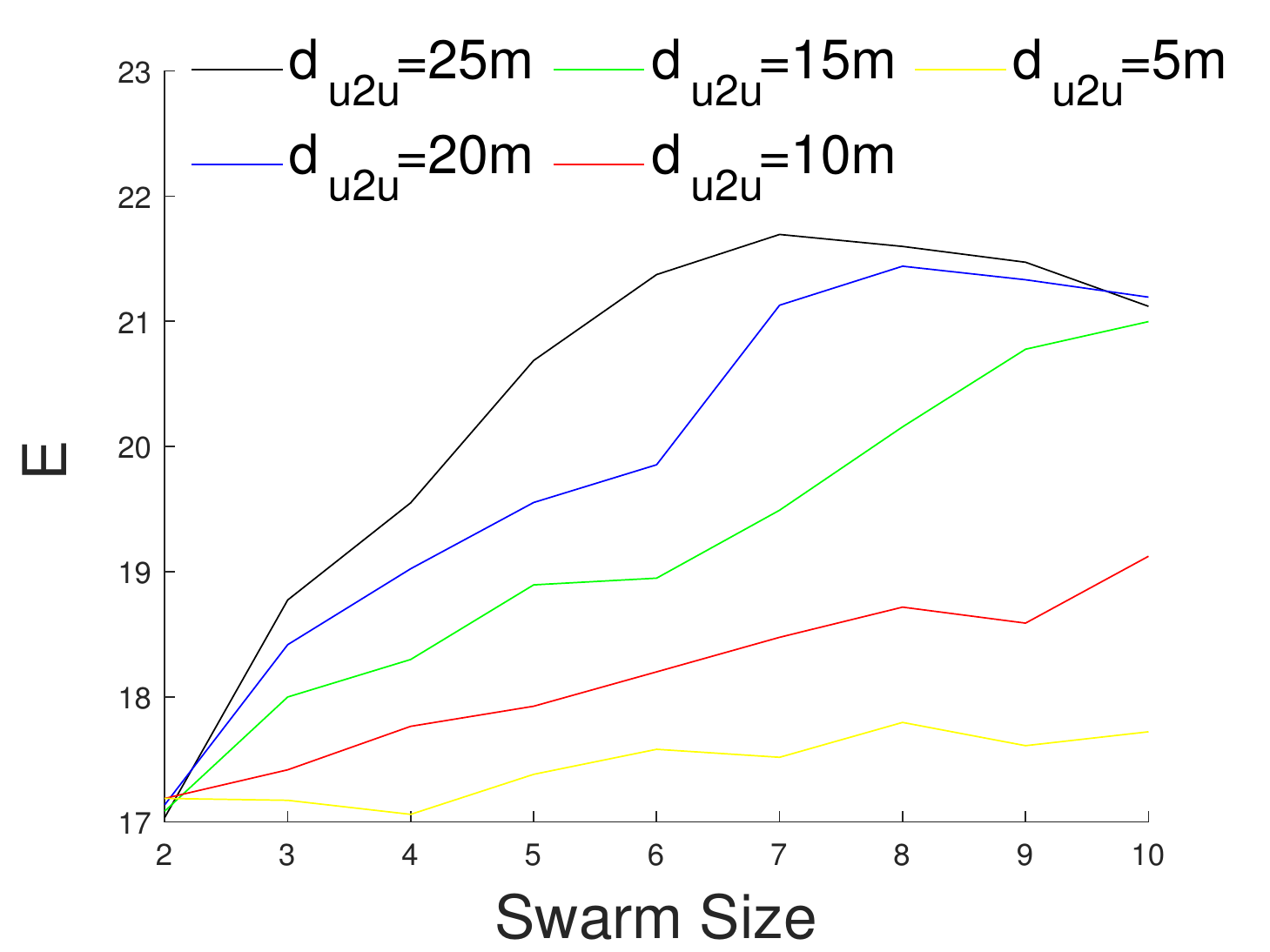}%
		\label{17c}}
    \hfil
	\subfloat[]{\includegraphics[width=0.5\linewidth]{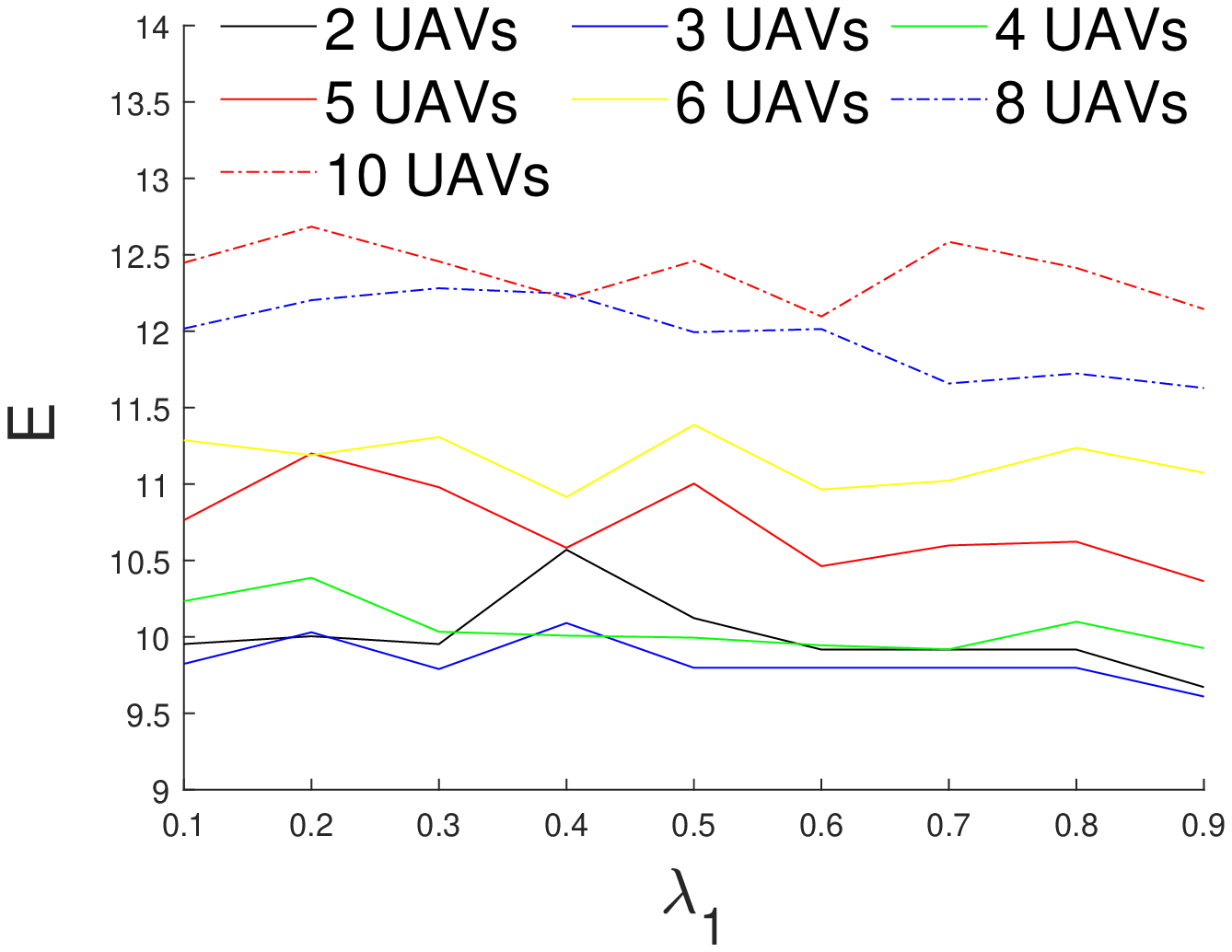}%
		\label{17d}}
	\hfil
	\subfloat[]{\includegraphics[width=0.5\linewidth]{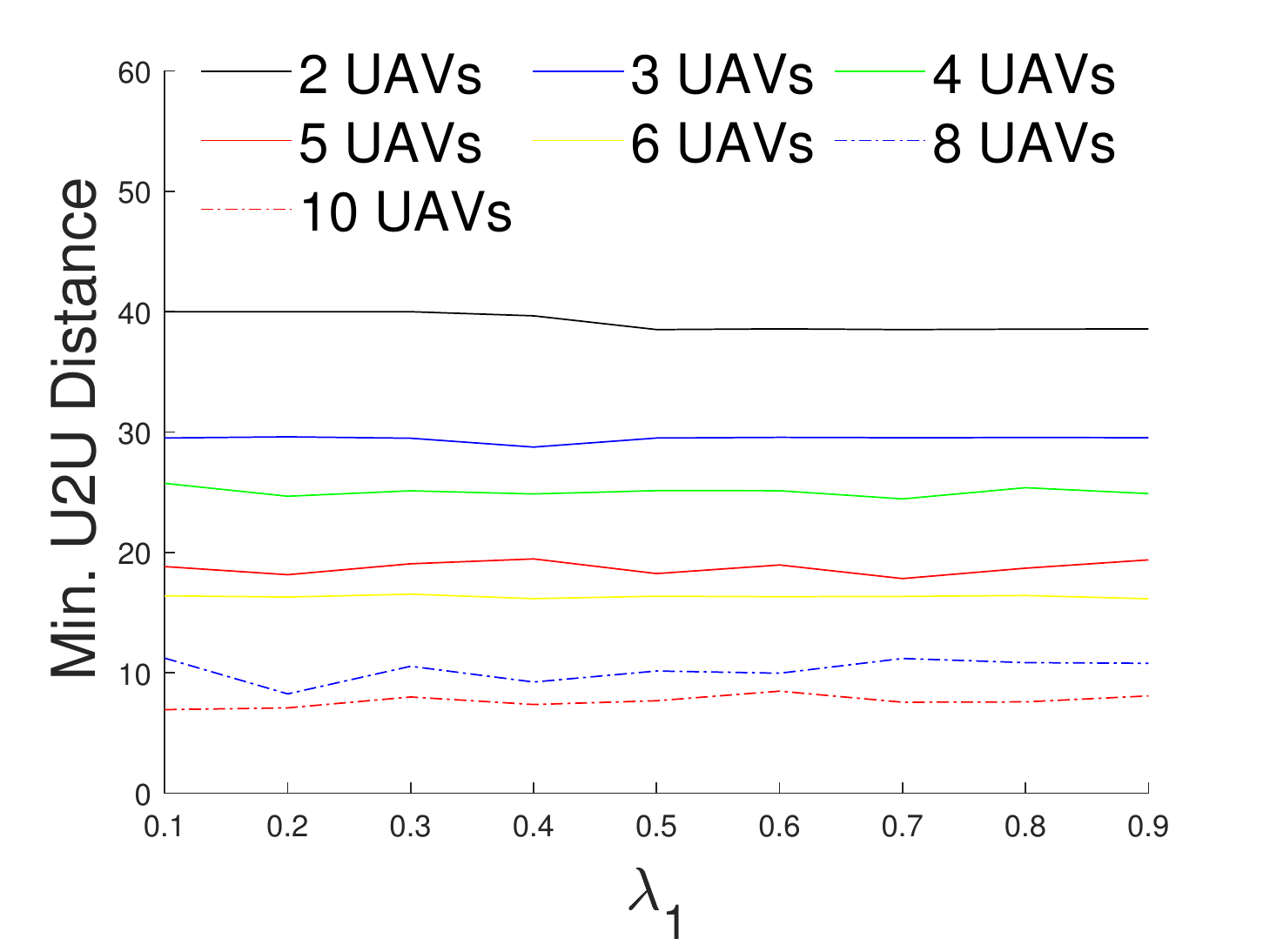}%
		\label{17e}}
	\caption{Impacts of parameters. (a) $d_{safe}$. (b) $d_{u2u}$ ($\tau=20m$). (c) $d_{u2u}$ ($\tau=10m$). (d) and (e) $\lambda_1$. }
	\label{fig17}
\end{figure}



\begin{itemize}
	\item \textbf{Impact of $d_{safe}$. } 
	From Fig.~\ref{17a}, the energy consumption of $E^2CoPre$ increases with $d_{safe}$ for all obstacle speeds. Larger $d_{safe}$ forces the trajectories to deviate further from the obstacles, leading to more curved and longer paths, which in turn, increases the energy consumption; 
	\item \textbf{Impact of $d_{u2u}$. } From Fig.~\ref{17b} and \ref{17c}, the energy consumption of $E^2CoPre$ increases with the swarm size. This is because, with more UAVs, the trajectories of the UAVs need to be longer and more curved to maintain the required $d_{u2u}$ distance between them; 
	In general, the energy consumption under $\tau=10\ m$ is larger than $\tau=20\ m$ due to the increased crowdedness of the swarm. 
	\item \textbf{Impact of $\lambda_1$. } From Fig.~\ref{17d} and \ref{17e}, the energy consumption and minimum U2U distance are not significantly affected by the selection of $\lambda_1$. 
\end{itemize}

\subsubsection{Ablations} 

\begin{figure}[bt]
	\centering
	\includegraphics[width=0.3\textwidth]{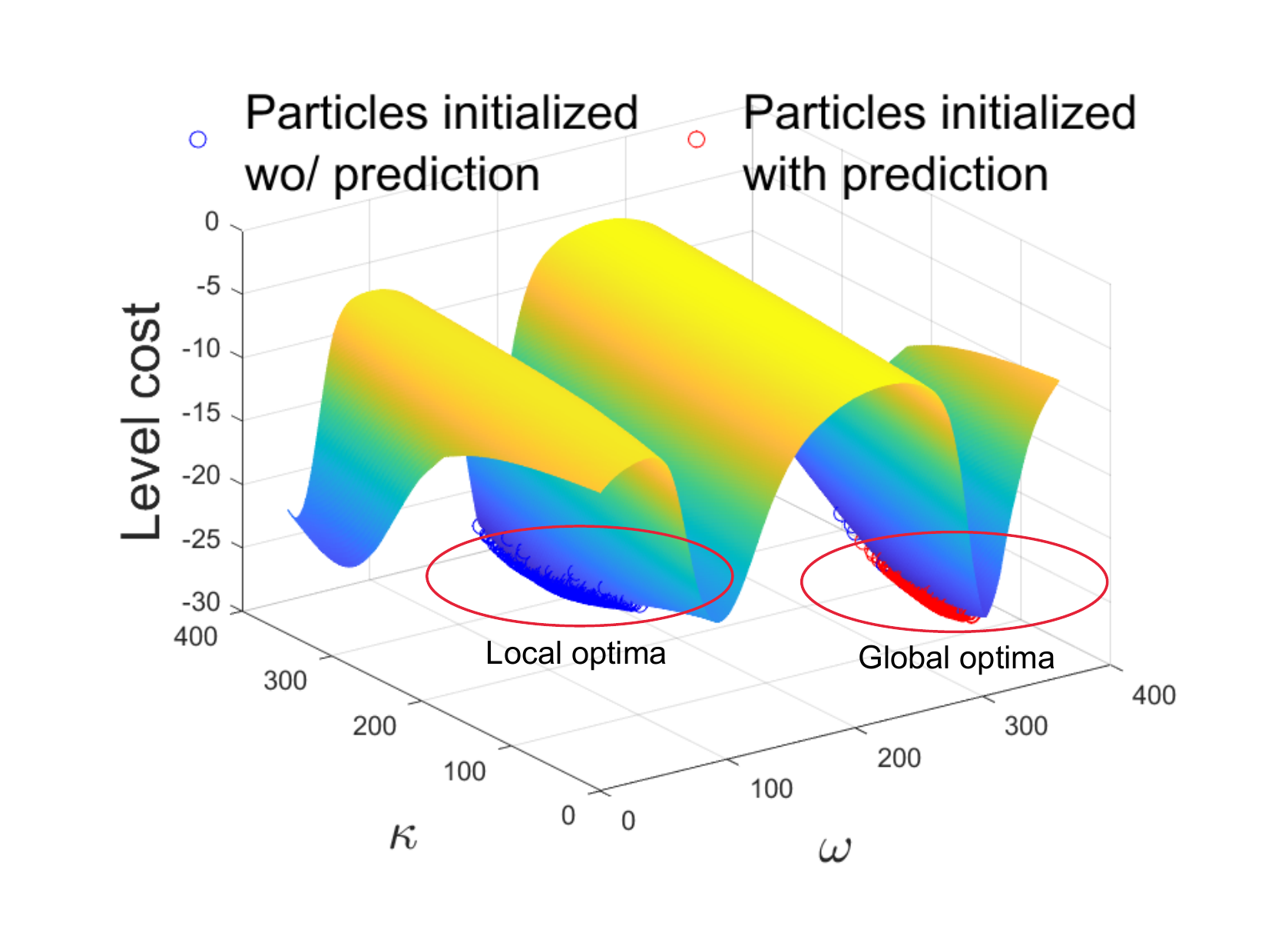} 
	\caption{Solutions of PSO in $E^2Coop$ and $E^2CoPre$ across 500 searches. } 
	\label{figPSO}
\end{figure}

\begin{figure}[bt]
	\centering
	\subfloat[]{\includegraphics[width=0.5\linewidth]{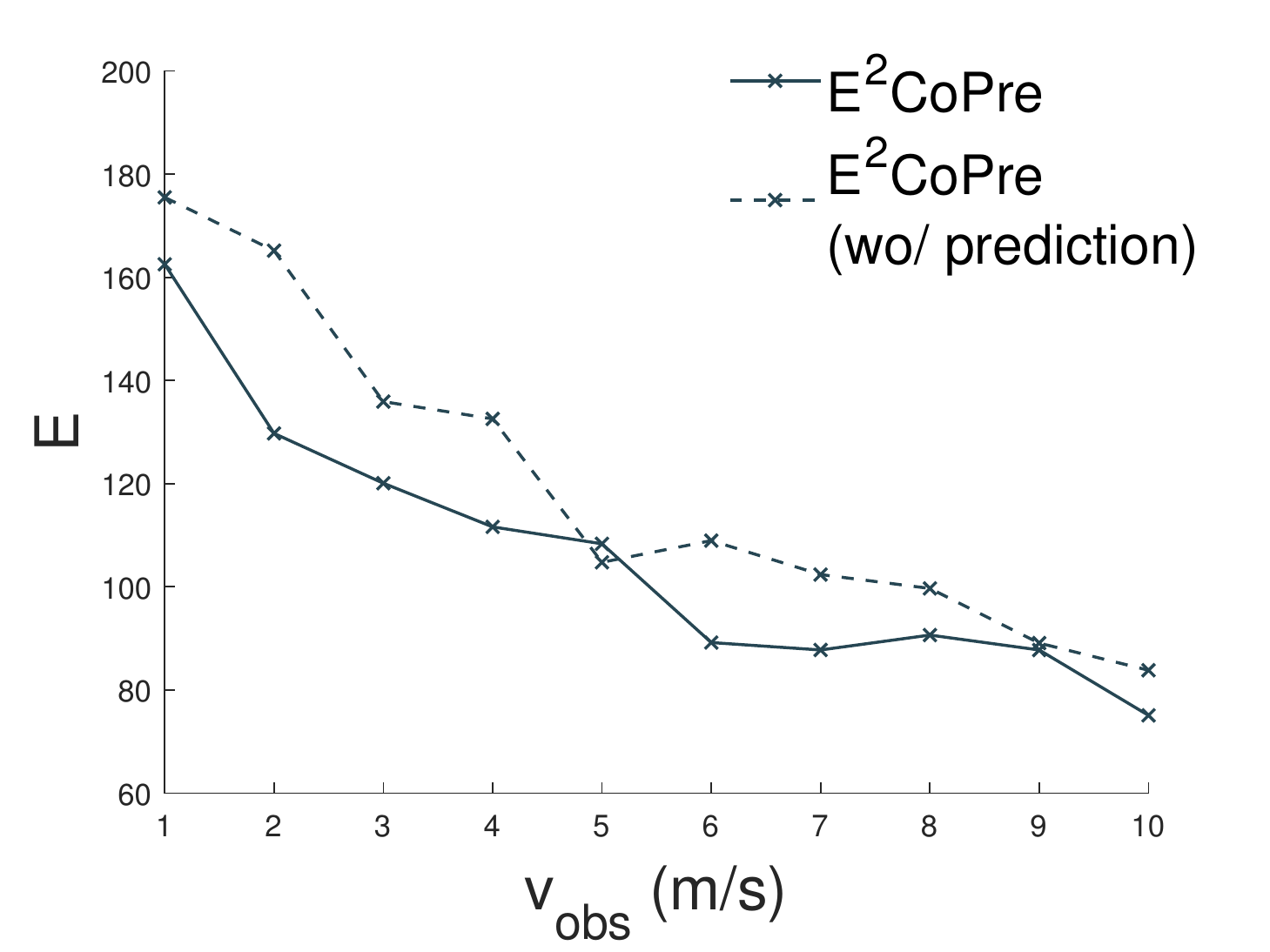}%
		\label{fig18a}}
	\hfil
	\subfloat[]{\includegraphics[width=0.5\linewidth]{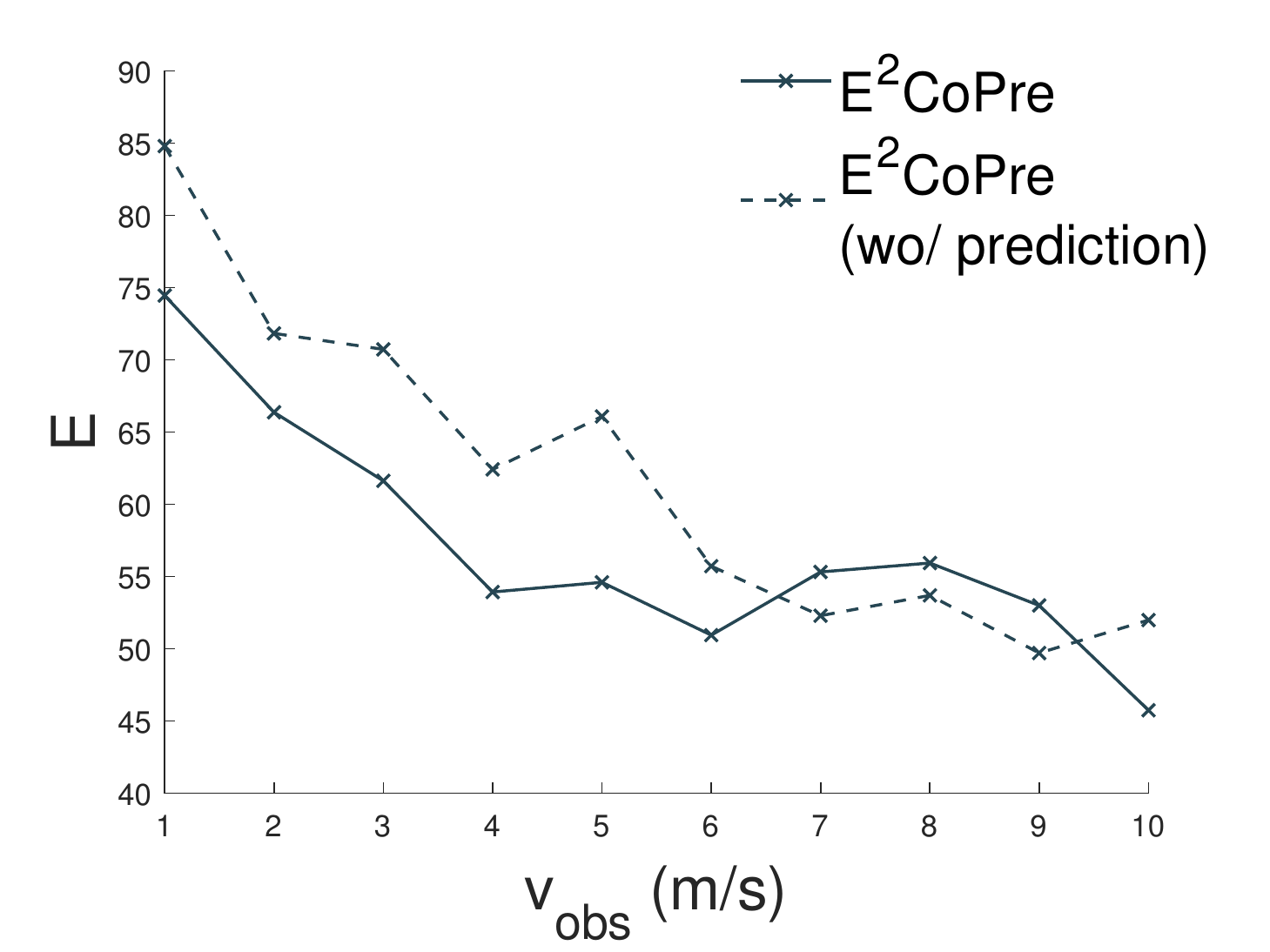}%
		\label{fig18b}}
	\hfil
	\subfloat[]{\includegraphics[width=0.5\linewidth]{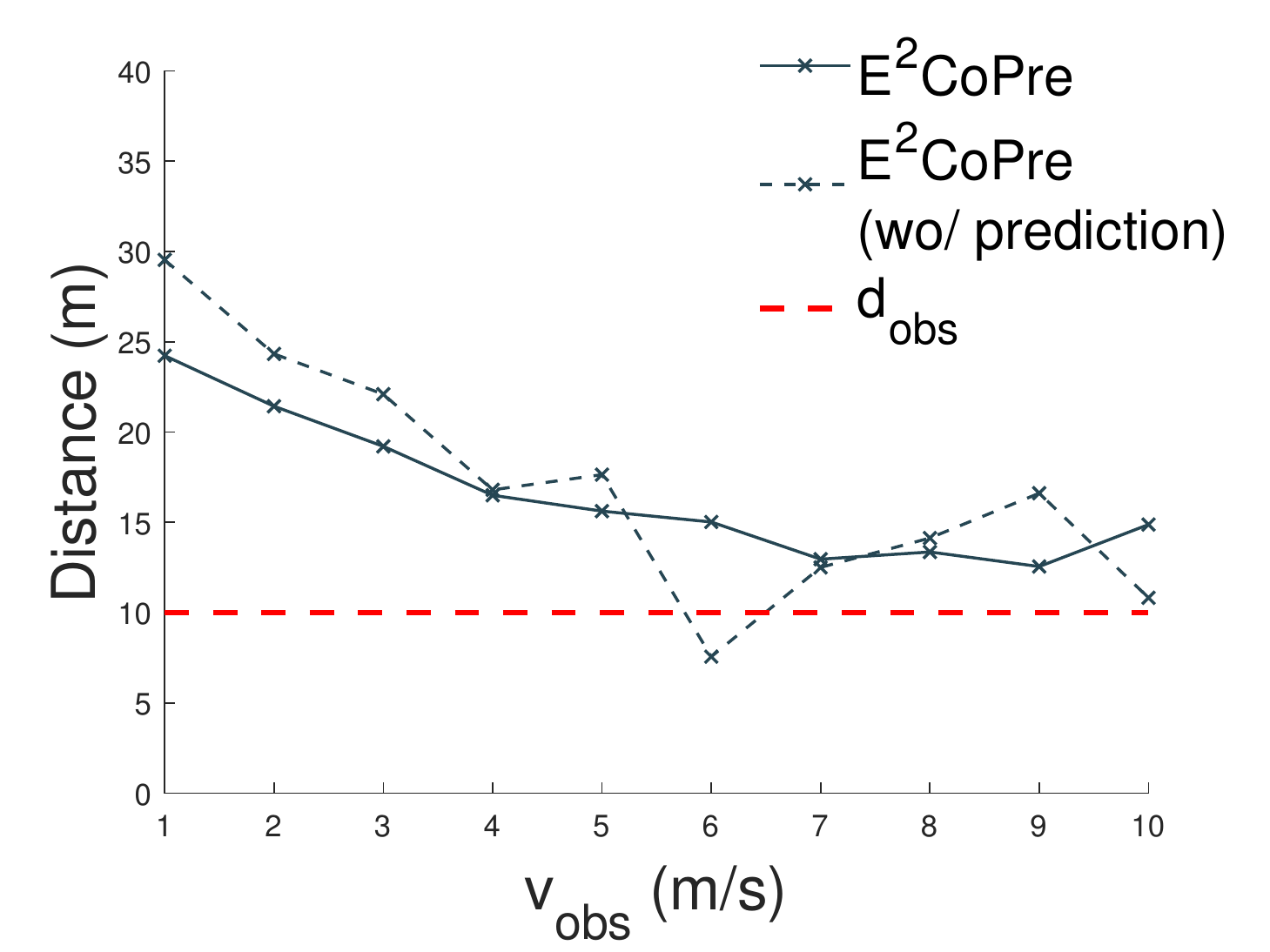}%
		\label{fig18c}}
	\hfil
	\subfloat[]{\includegraphics[width=0.5\linewidth]{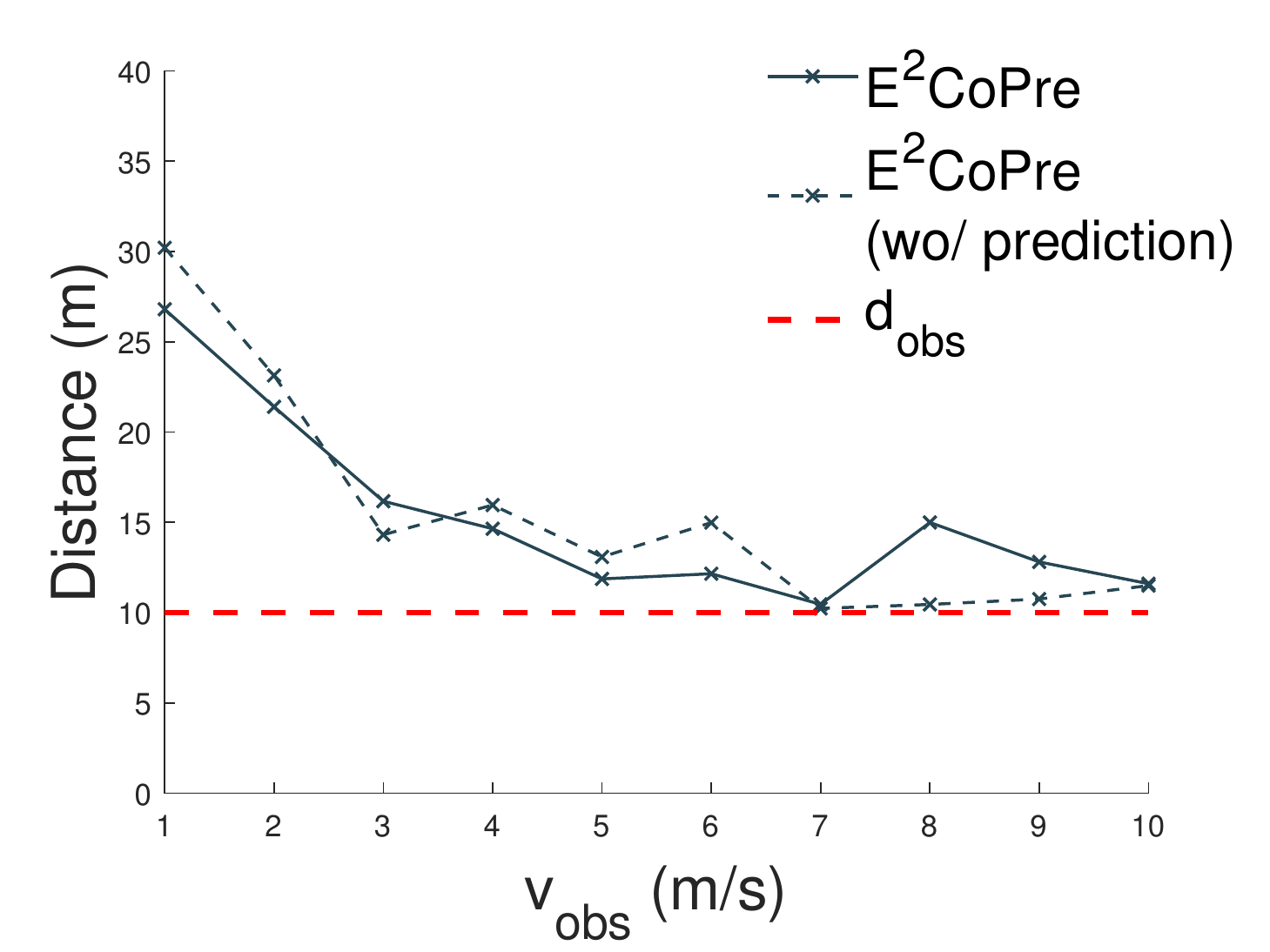}%
		\label{fig18d}}
	\hfil
	\subfloat[]{\includegraphics[width=0.5\linewidth]{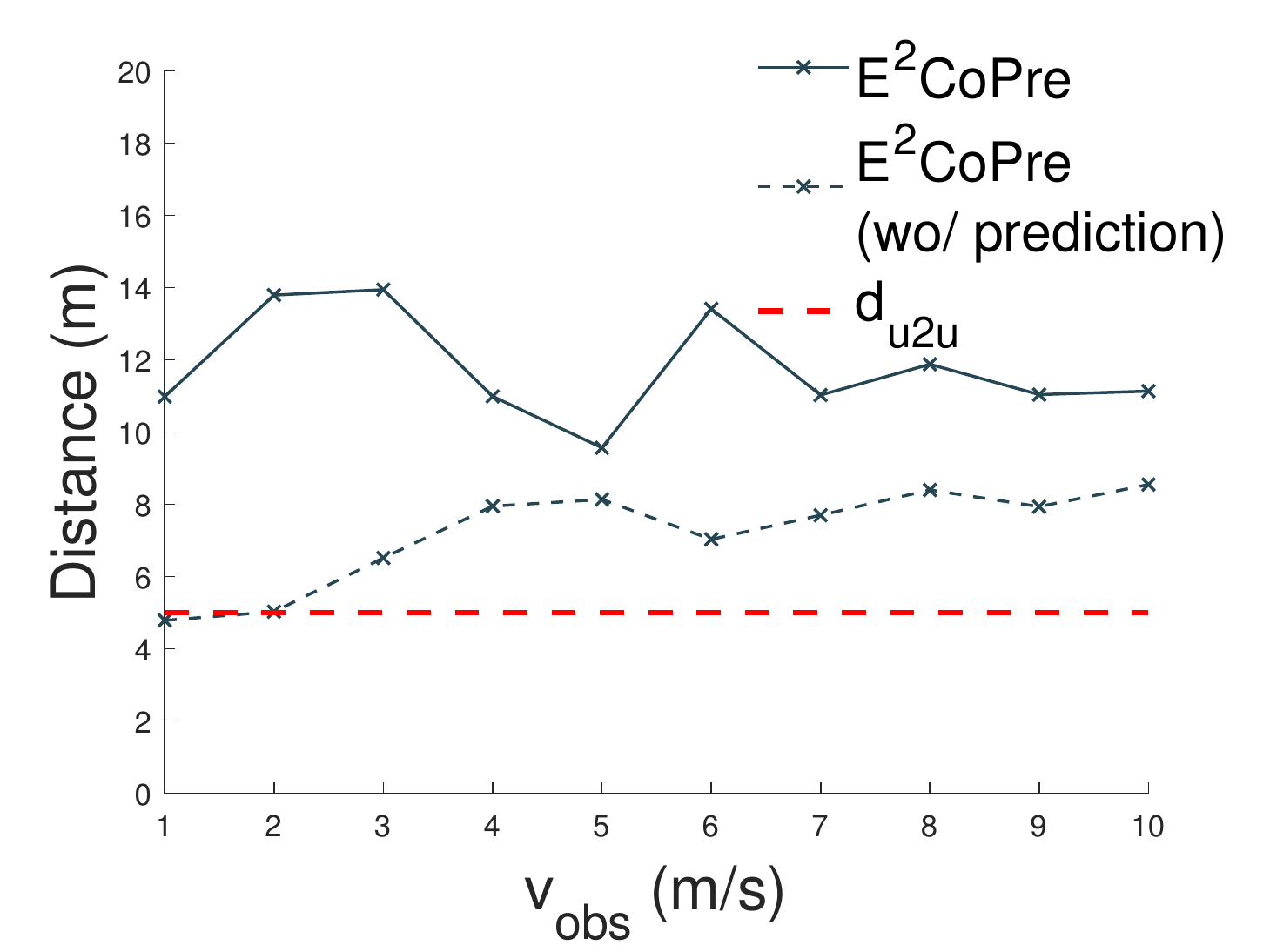}%
		\label{fig18e}}
	\hfil
	\subfloat[]{\includegraphics[width=0.5\linewidth]{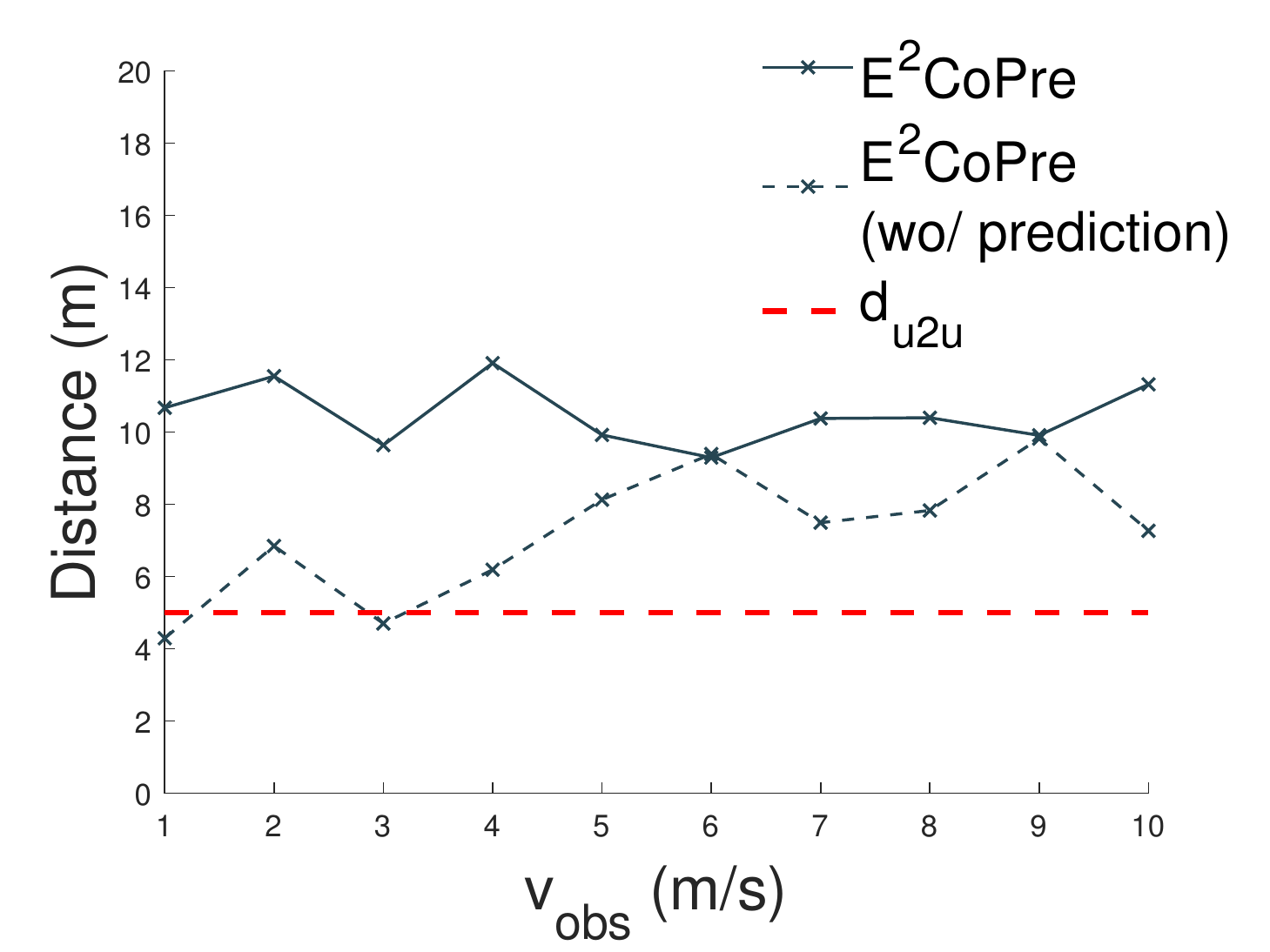}%
		\label{fig18f}}
	\caption{Ablation tests on trajectory prediction. (a) \textit{Obstacle in Front} Energy. (b) \textit{Obstacle on Side} Energy. (c) \textit{Obstacle in Front} U2O. (d) \textit{Obstacle on Side} U2O. (e) \textit{Obstacle in Front} U2U. (f) \textit{Obstacle on Side} U2U. }
	\label{fig18}
\end{figure}


{
	Ablation tests are necessary to validate the importance of trajectory prediction in $E^2CoPre$. 
	To illustrate the advantages of trajectory prediction in obtaining the optimal solutions in PSO search, we present the PSO-\textit{Level} search space in Fig. \ref{figPSO}. 
	The red and blue dots represent the solutions obtained with and without trajectory prediction, respectively. The experiment is repeated 500 times for each algorithm. Fig. \ref{figPSO} demonstrates that when initialized with trajectory prediction, the particles accurately identify the global optimum, while they frequently get trapped at local optima when initialized without trajectory prediction. Out of the 500 searches performed, the particles get trapped at local optima in 98.40\% of cases when initialized without trajectory prediction. In contrast, the particles experienced an occurrence of 0\% finding the local optima when initialized with trajectory prediction. Furthermore, the average Euclidean distances on $\kappa$-$\omega$ plane  between the solutions obtained by the particles and the global optimum is $227.22\pm28.33$ when initialized without trajectory prediction, whereas it is only $25.42\pm18.75$ when initialized with trajectory prediction. 
}

In addition, Fig.~\ref{fig18a} and \ref{fig18b} show that trajectory prediction helps reduce the average energy consumption across all obstacle velocities by 8.67\% in scenario \textit{Obstacle in Front} and 6.64\% in scenario \textit{Obstacle on Side}, respectively. On the other hand, Fig.~\ref{fig18c} to \ref{fig18f} show that trajectory prediction helps reduce redundant distances. 

\subsection{Trajectory Demonstration}
\begin{figure}[t]
	\centering
	\subfloat[]{\includegraphics[width=0.45\linewidth]{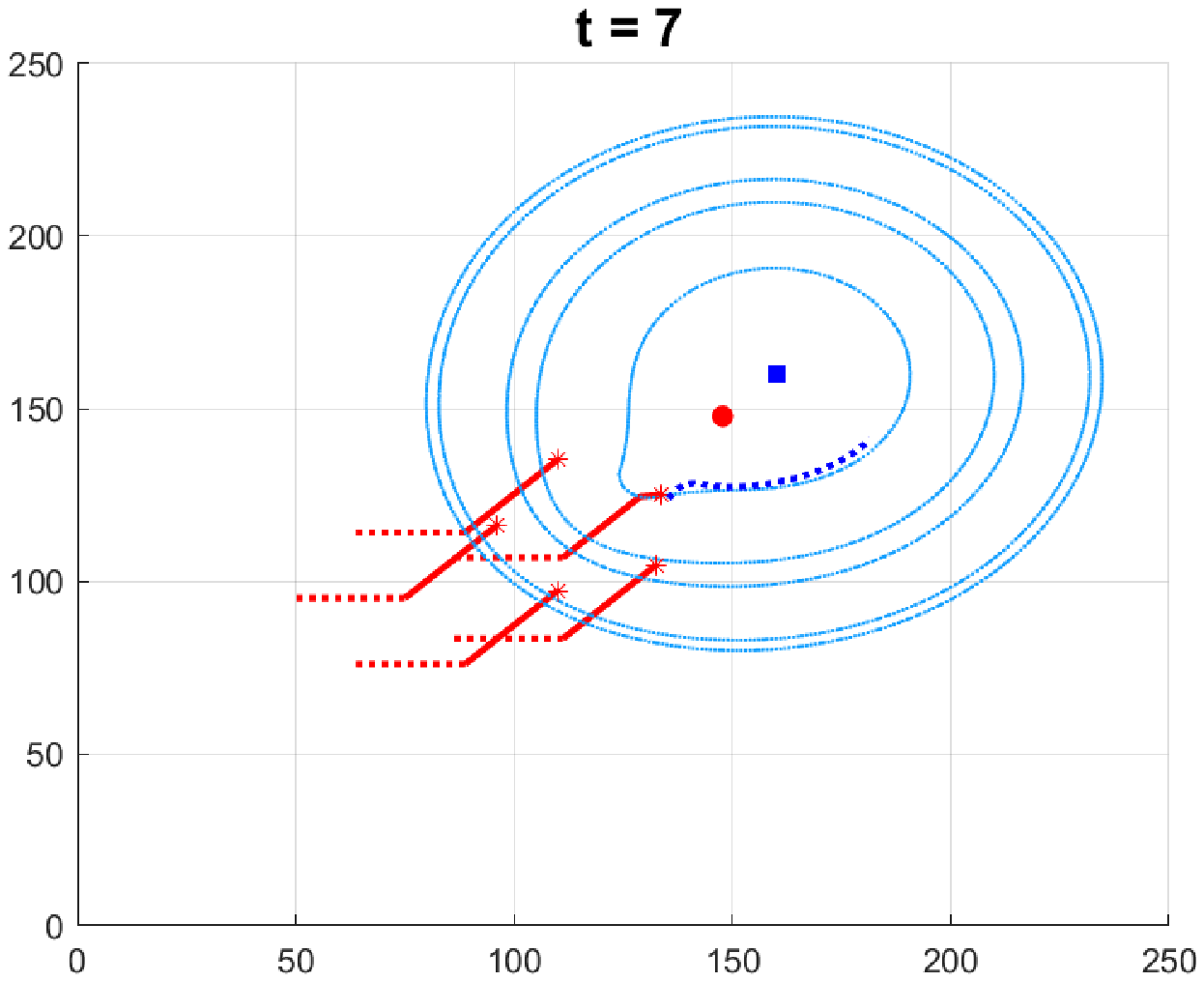}%
		\label{11b}}
	\hfil
	\subfloat[]{\includegraphics[width=0.45\linewidth]{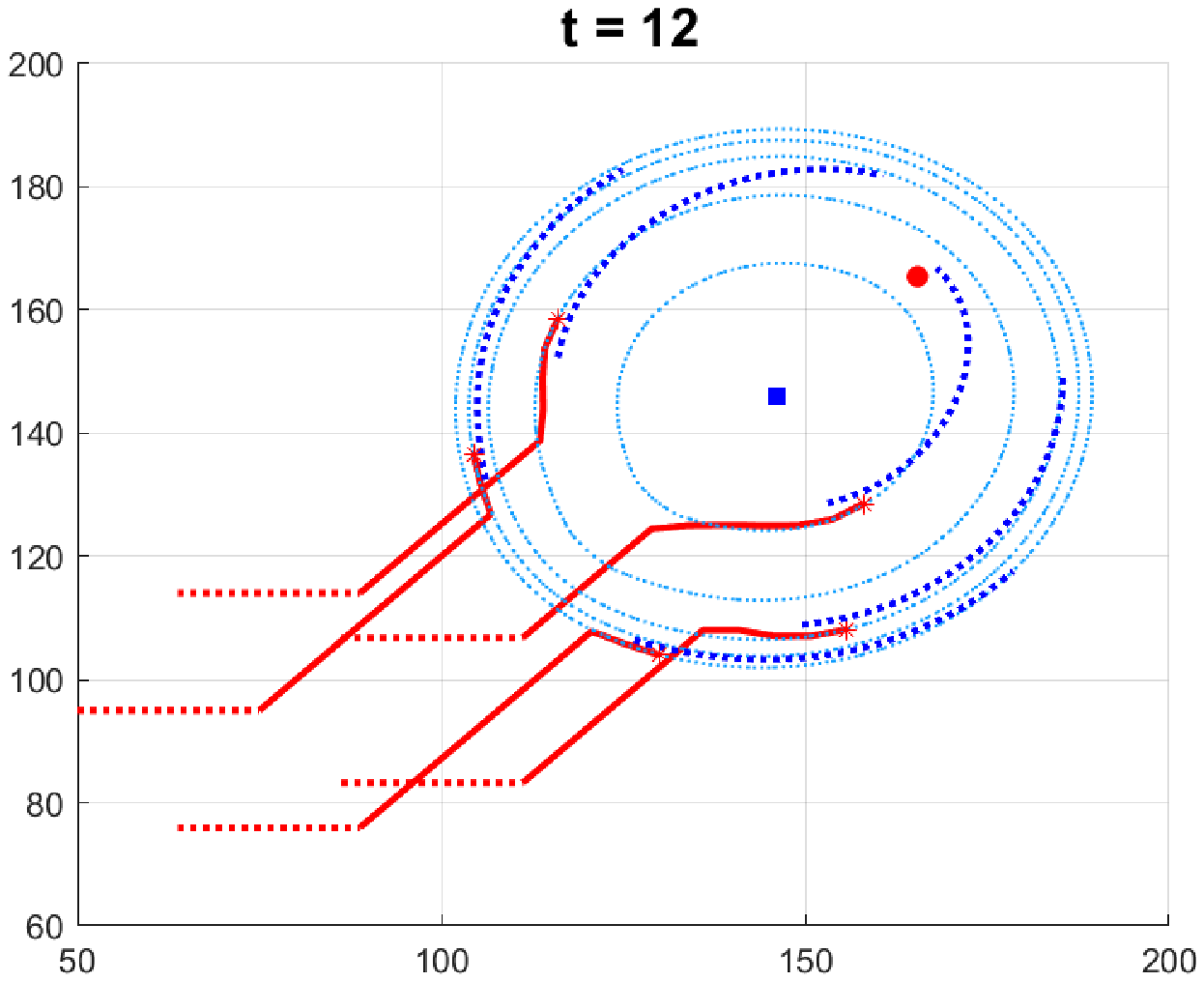}%
		\label{11d}}
	\hfil
	\subfloat[]{\includegraphics[width=0.45\linewidth]{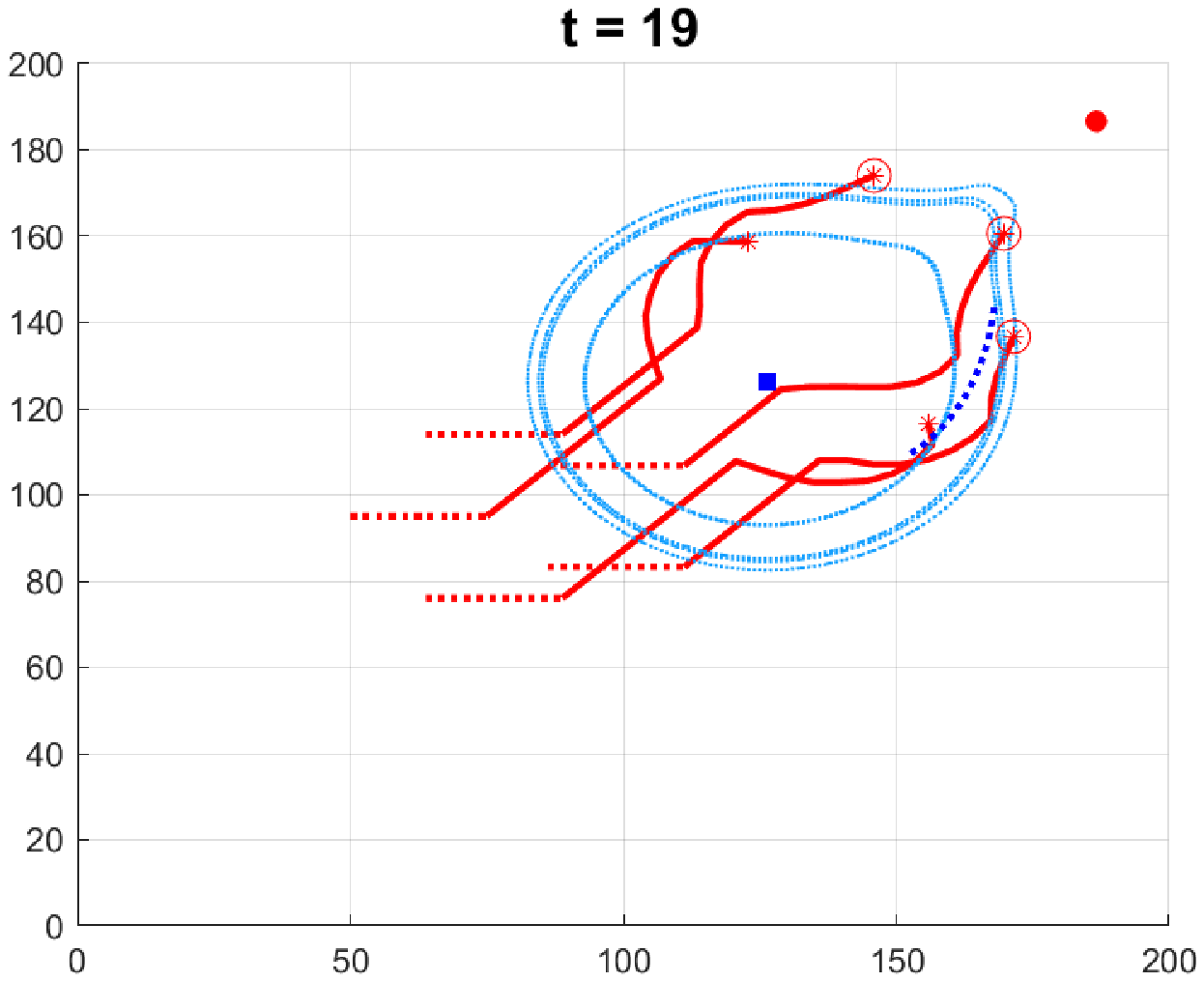}%
		\label{11f}}
	\hfil
	\subfloat[]{\includegraphics[width=0.45\linewidth]{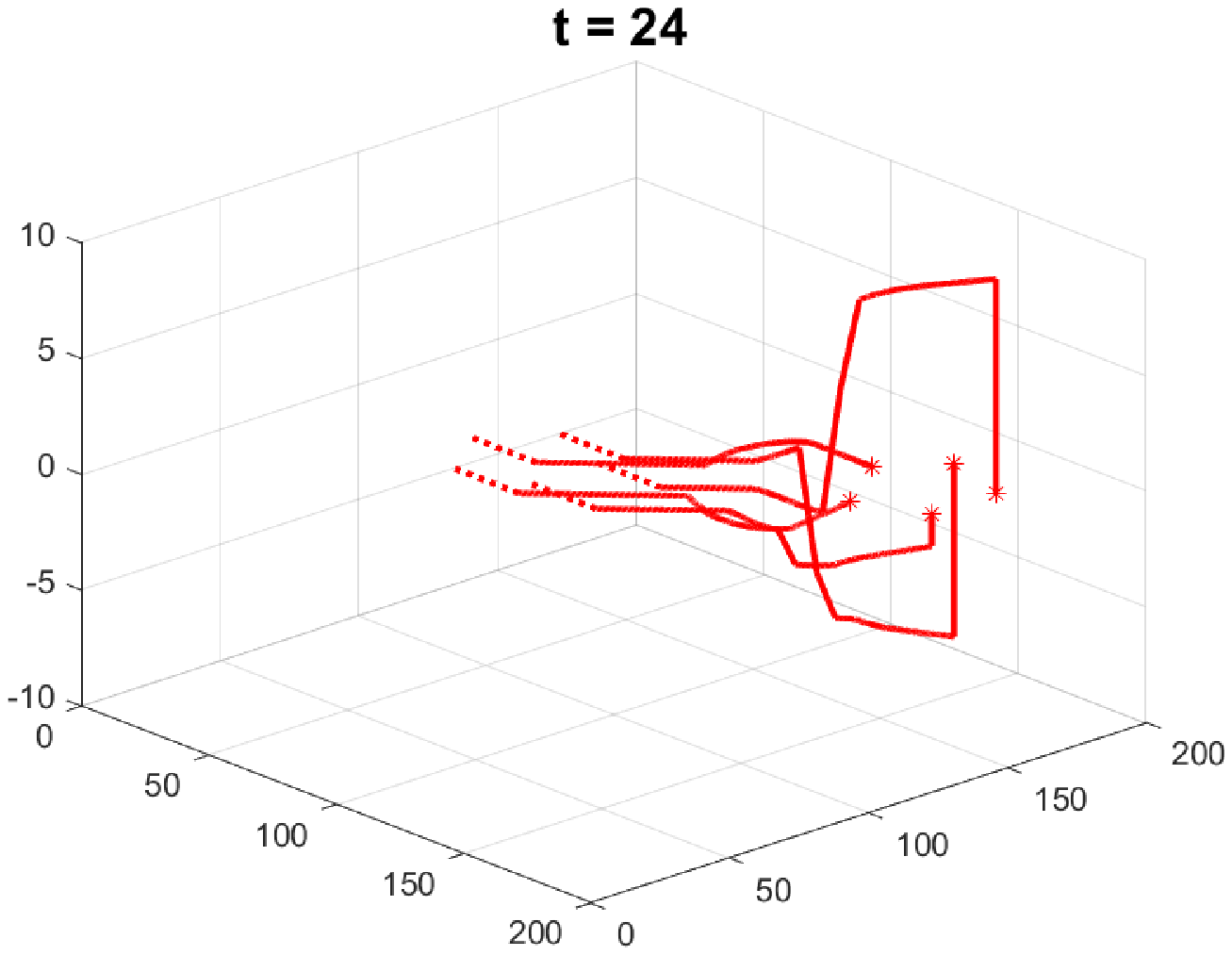}%
		\label{11h}}
	\caption{Trajectories of 5 UAVs with one obstacle. (a) Before avoidance. (b) Start of avoidance. (c) During avoidance. (d) After avoidance. }
	\label{fig11}
\end{figure}

\begin{figure}[t]
	\centering
	\subfloat[]{\includegraphics[width=0.45\linewidth]{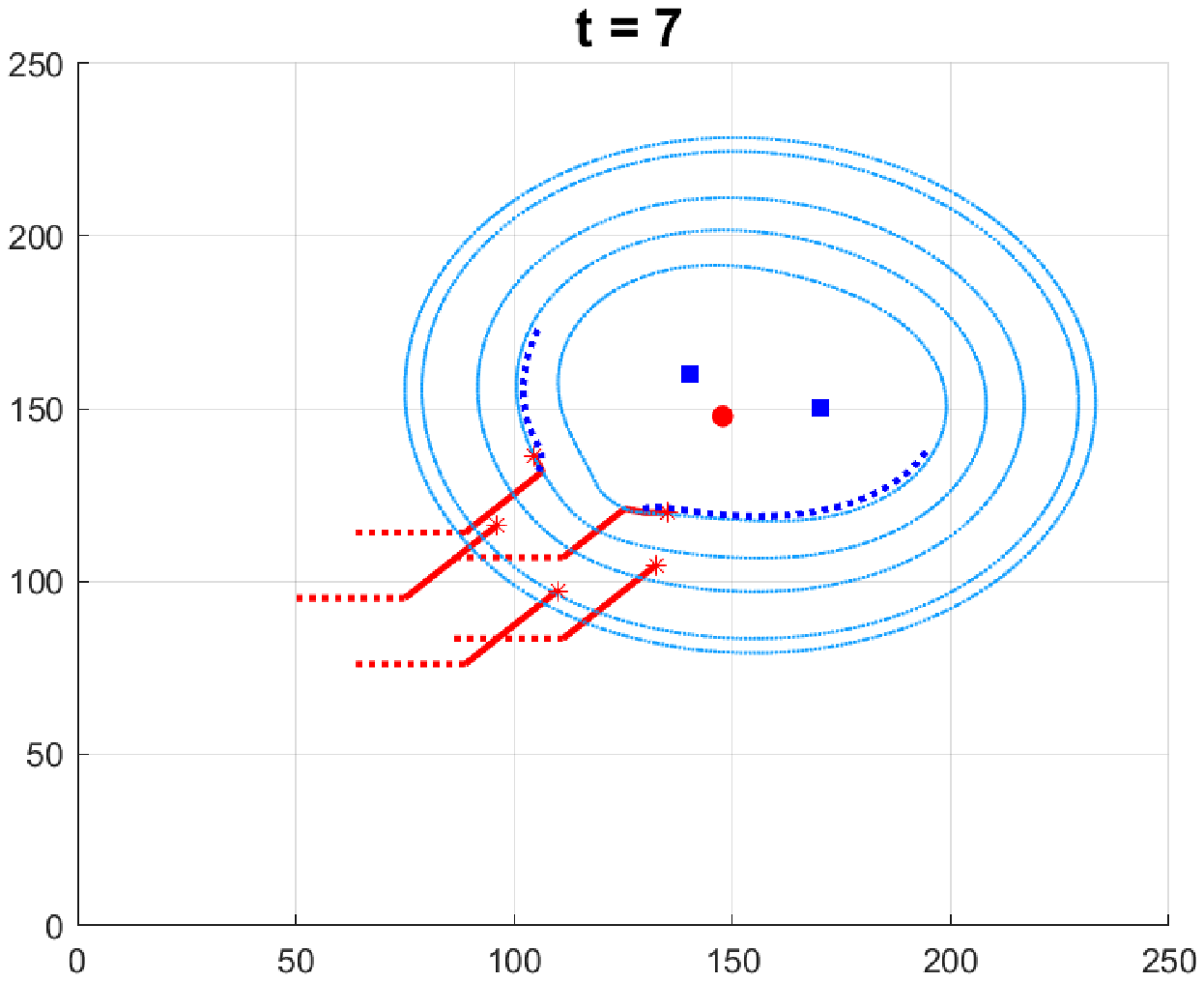}%
		\label{12b}}
	\hfil
	\subfloat[]{\includegraphics[width=0.45\linewidth]{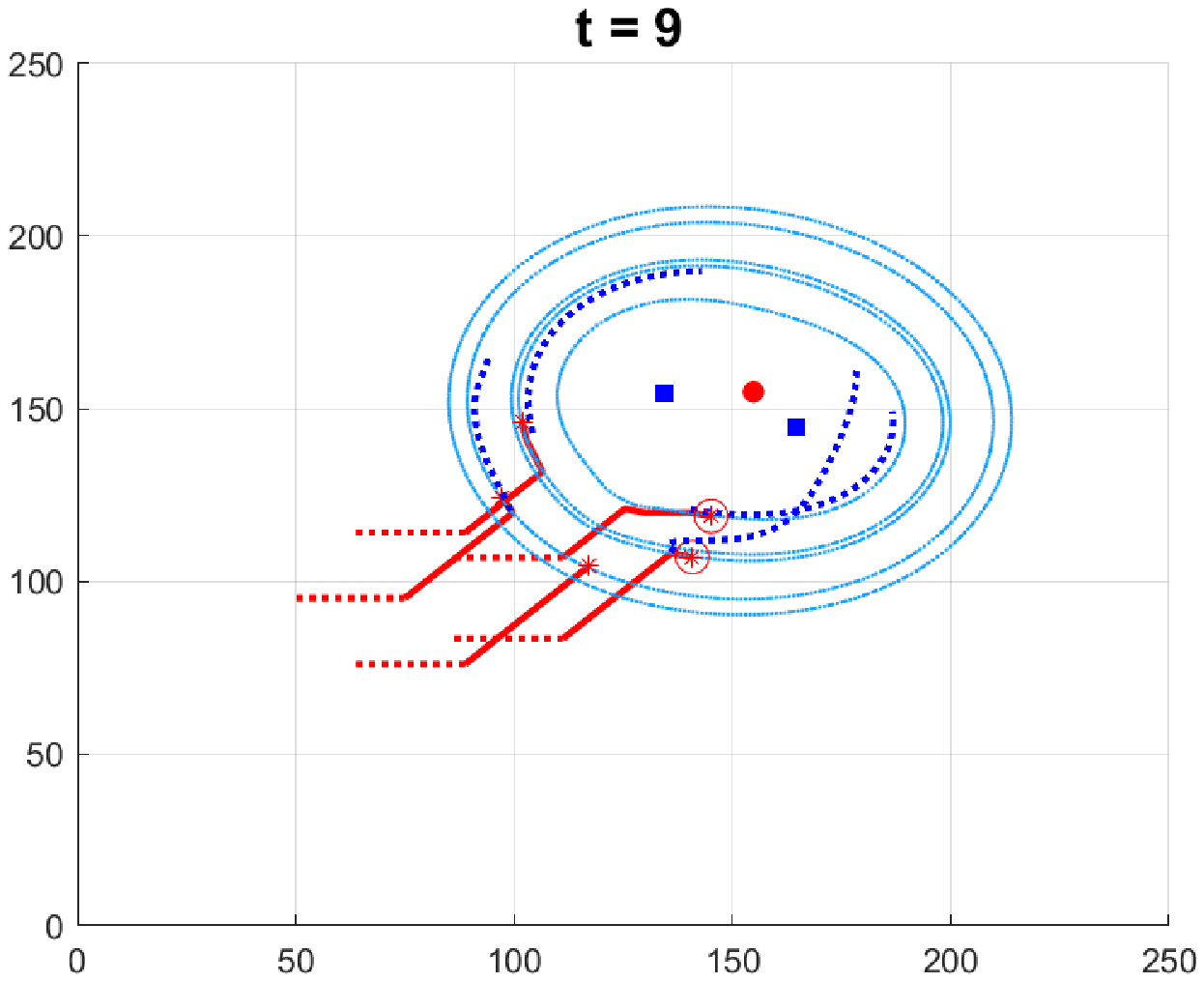}%
		\label{12d}}
	\hfil
	\subfloat[]{\includegraphics[width=0.45\linewidth]{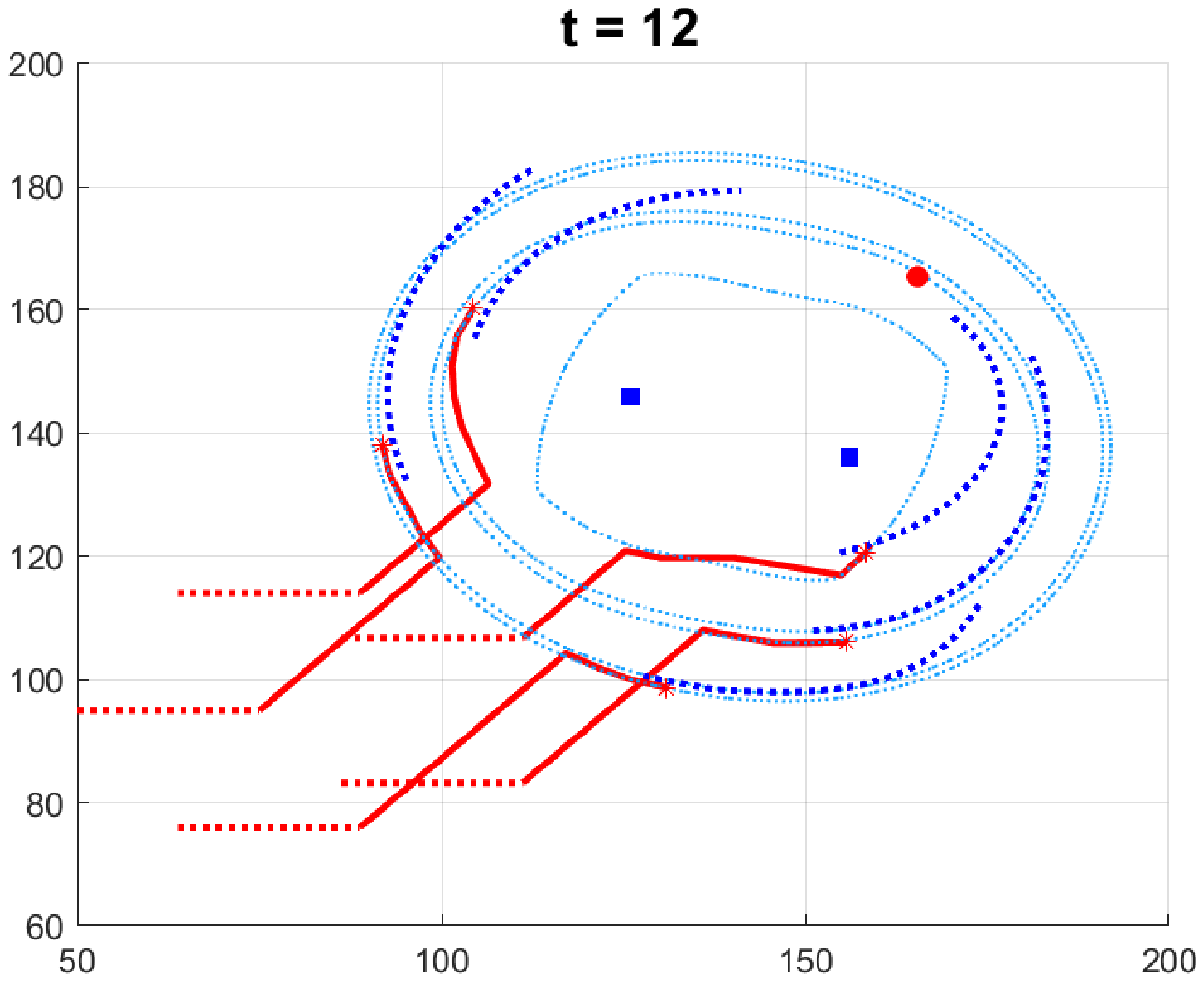}%
		\label{12f}}
	\hfil
	\subfloat[]{\includegraphics[width=0.45\linewidth]{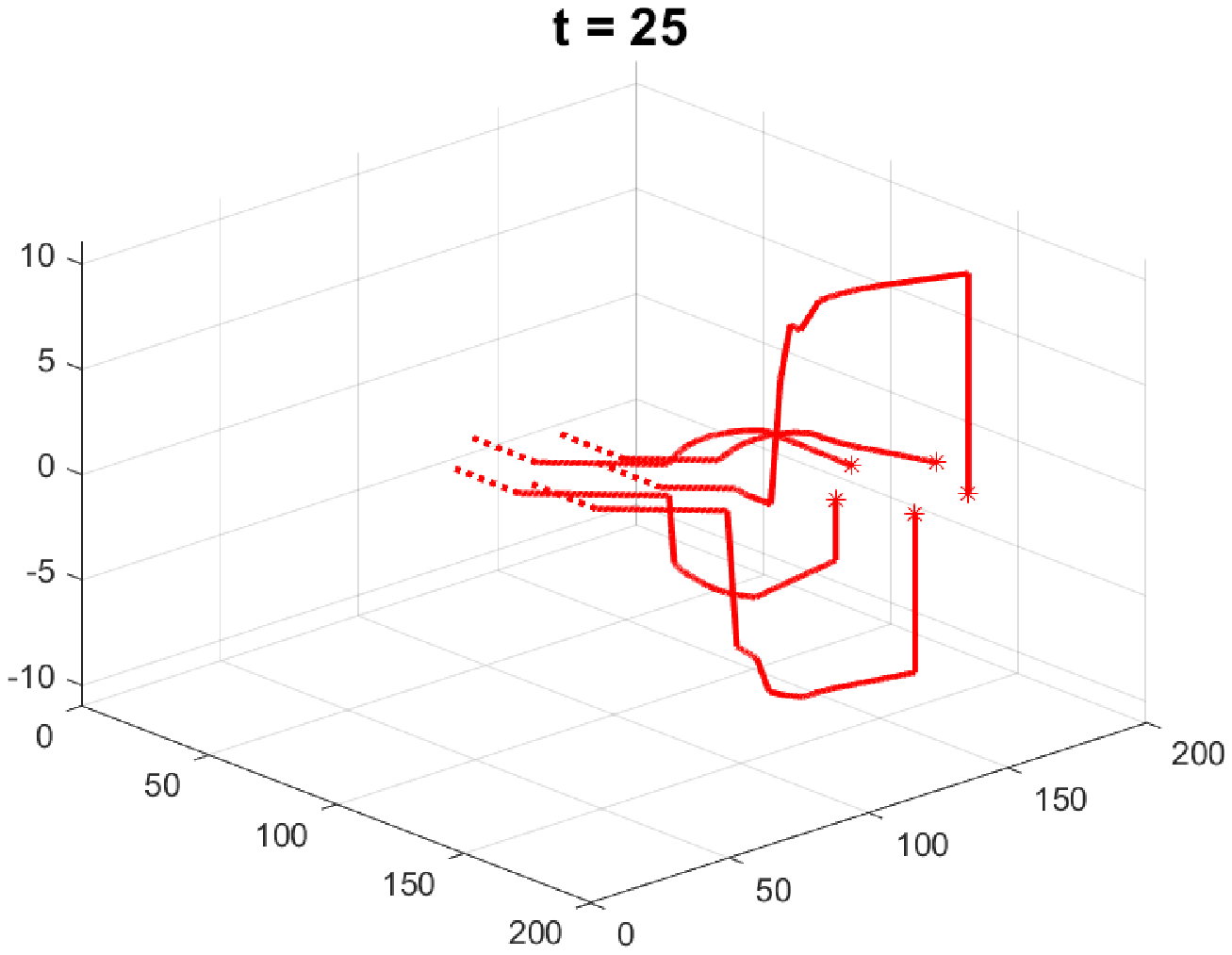}%
		\label{12h}}
	\caption{Trajectories of 5 UAVs with two obstacles. (a) Before avoidance. (b) Start of avoidance. (c) During avoidance. (d) After avoidance. }
	\label{fig12}
\end{figure}

Fig.~\ref{fig11} and \ref{fig12} demonstrate the trajectories generated by $E^2CoPre$. The swarm formation is a circle with $\tau=20\ m$. The swarm speed is $v_s=5\ m/s$. $\lambda_1=\lambda_2=0.5$. $d_{u2u}$ and $d_{obs}$ are set to 10 $m$. $d_{u2u}$ is deliberately set to as large as 10 $m$ to trigger PSO-\textit{Alt} to demonstrate how $E^2CoPre$ resolves U2U collisions in 3D space. 
The swarm's conceptual center and the obstacle's geometrical center are represented with a red circle and a blue square, respectively. The UAVs' original trajectories before avoidance are shown in red dashed lines. 
The planned trajectories are shown using red solid curves. 
The predicted trajectories are depicted with dark blue dotted curves. 
The contours of the environment field are shown in light blue dotted curves. 

Fig. \ref{fig11} shows the trajectories of 5 UAVs with one obstacle. Fig. \ref{11b}, \ref{11d} and \ref{11f} are in level planning. 
At the time $t=19$ (Fig. \ref{11f}), the distances between the predicted trajectories of three UAVs in red circles are smaller than $d_{u2u}$. Hence, altitude planning is triggered, and the collisions among the three UAVs are resolved in three-dimensional space. 
Fig. \ref{fig12} shows the trajectories of 5 UAVs with two obstacles. Fig. \ref{12b} and \ref{12d} are in level planning. 
At the time $t=12$ (Fig. \ref{12f}), the predicted trajectories of the two UAVs in red circles intersect. The collisions are resolved in three-dimensional space. 

\section{Conclusions} \label{Concl} 
In this paper, we address the challenge of trajectory planning for UAV swarms to acheive energy-efficient collision avoidance. To tackle this problem, we propose a novel approach called $E^2CoPre$, which combines the advantages of APF and PSO. $E^2CoPre$ introduces several key contributions, including implicit coordination, a novel cost function, trajectory-based PSO, and trajectory prediction. Our simulation results demonstrate the superior performance of $E^2CoPre$ compared to existing methods. Future work include evaluating the impact of packet loss in  wireless communication and integrating neural networks for intelligent decision making.

\bibliographystyle{IEEEtran}
\bibliography{IEEEfull}


\section{Biography Section}
\begin{IEEEbiography}[{\includegraphics[width=.9in,height=1.25in,clip,keepaspectratio]{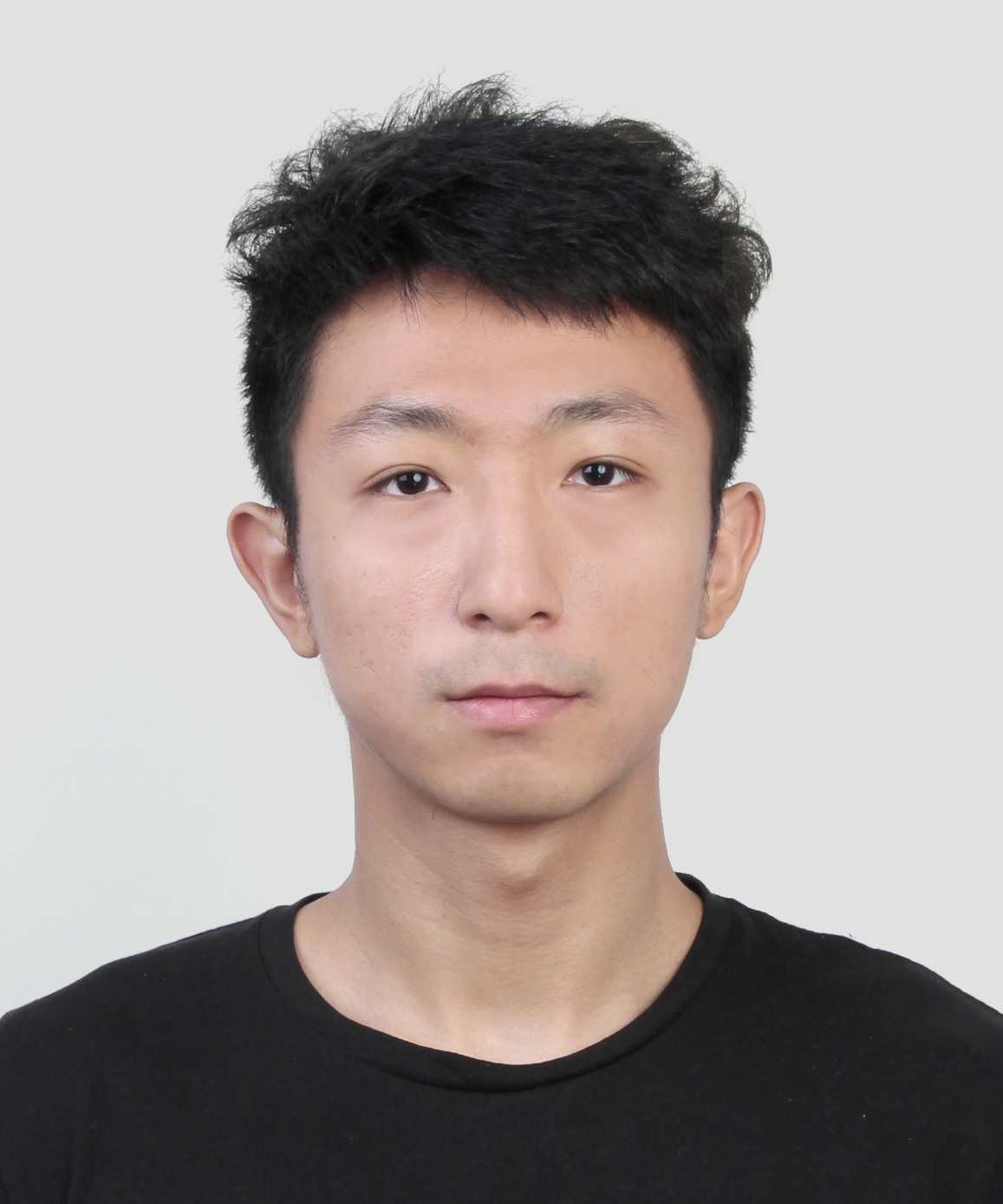}}]
	{Shuangyao Huang} received B.Eng. degree in communication engineering from the Northeast Forestry University, China in 2014 and M.Sc. degree in electrical engineering from the National University of Singapore in 2017. In 2023, he received his Ph.D. degree in computer science from the University of Otago, New Zealand. 
    His research interest is swarm intelligence. 
\end{IEEEbiography}
\begin{IEEEbiography}[{\includegraphics[width=1in,height=1.25in,clip,keepaspectratio]{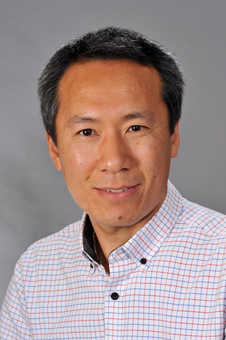}}]
	{Haibo Zhang} received M.Sc. degree in computer science from Shandong Normal University, China in 2005 and Ph.D. degree in computer science from the University of Adelaide, Australia in 2009. From 2009 to 2010, he was a Postdoctoral Research Associate with the School of Electrical Engineering, Royal Institute of Technology, Sweden. He is now an Associate Professor at the School of Computing, University of Otago, New Zealand. His current research interests include wireless networks, distributed computing and edge intelligence. 
\end{IEEEbiography}
\begin{IEEEbiography}[{\includegraphics[width=1in,height=1.25in,clip,keepaspectratio]{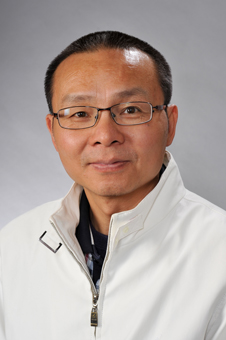}}] 
	{Zhiyi Huang} is a Professor in the School of Computing, University of Otago, New Zealand. He received his Ph.D. degree in Computer Science in 1992 from National University of Defense Technology, China. His research areas include machine learning and its applications, parallel/distributed computing, signal processing, multi-core/manycore architectures, cluster computing, operating systems, and computer networks.
\end{IEEEbiography}
 
\vspace{11pt}

%
%

\vfill

\end{document}